\documentclass[3p,times]{elsarticle}

\usepackage{amsmath,amssymb,amsfonts}
\usepackage{graphicx}
\usepackage{booktabs}
\usepackage{multirow}
\usepackage{hyperref}
\usepackage{bm}
\usepackage{siunitx}
\usepackage{xcolor}
\usepackage{array}
\usepackage{tabularx}
\usepackage{makecell}
\usepackage{adjustbox}
\usepackage{float}          
\usepackage{placeins}       
\usepackage{algorithm}
\usepackage{algpseudocode}
\usepackage{caption}
\usepackage{enumitem}

\journal{Preprint}

\begin{document}

\begin{frontmatter}

\title{MultiPUFFIN: A Multimodal Domain-Informed Foundation Model for Molecular Property Prediction of Small Molecules}

\author[ntnu]{Idelfonso B.R. Nogueira\corref{cor1}}
\cortext[cor1]{Corresponding author}
\ead{idelfonso.nogueira@ntnu.no}

\author[ntnu]{Carine M. Rebello}

\author[kuleuven]{Mumin Enis Leblebici}

\author[surrey]{Erick Giovani Sperandio Nascimento}

\affiliation[ntnu]{organization={Department of Chemical Engineering, Norwegian University of Science and Technology (NTNU)},
            city={Trondheim},
            postcode={7034},
            country={Norway}}

\affiliation[kuleuven]{organization={Faculty of Industrial Engineering, KU Leuven},
            city={Diepenbeek},
            country={Belgium}}

\affiliation[surrey]{organization={University of Surrey},
            city={Guildford},
            country={United Kingdom}}

\begin{highlights}
\item MultiPUFFIN fuses SMILES, molecular graphs, and 3D conformers with domain-informed prediction heads for nine thermophysical properties simultaneously.
\item Domain-informed inductive biases (Antoine, Andrade, van~'t Hoff, Born, Shomate equations) enforce physically consistent temperature dependence within individual properties by construction, and are competed against fragment and direct baselines in a per-property head tournament.
\item A self-supervised pretraining stage on 500{,}000 unlabeled PubChem molecules with three complementary objectives (graph--SMILES contrastive InfoNCE, masked atom feature reconstruction, masked SMILES token prediction) and two cross-property physical coupling terms (flash-point--vapor-pressure consistency and a stronger normal-boiling-point anchor on the vapor-pressure head) elevate the architecture to a thermodynamically coupled multimodal foundation model.
\item A four-stage training protocol (SSL pretrain, joint supervised multi-task training with the condition module stack, backbone-unfrozen targeted fine tune at very low learning rate, per-property G+ applicability-domain evaluation) yields a single deployable artifact whose accuracy and scope are reported jointly.
\item MultiPUFFIN achieves higher test $R^2$ than fine tuned ChemBERTa-2 on all nine properties despite using a single multi-task model vs.\ nine separately fine tuned models, with the most pronounced advantages on temperature-dependent properties (vapor pressure, viscosity) where ChemBERTa-2 cannot distinguish measurements of the same molecule at different temperatures.
\item Systematic ablation across a set of candidate thermodynamic representations per property reveals that the optimal domain-informed equation is property-specific, with group contribution heads (themselves rooted in thermodynamic additivity) providing the best inductive bias for several properties.
\item Inductive biases and multimodal encoding substantially reduce data and computational requirements compared to brute-force pretraining at scale.
\end{highlights}

\begin{abstract}
Predicting the physicochemical properties of small molecules across a diverse chemical space is an important step for chemical engineering, drug discovery and materials science. Existing molecular foundation models pretrain on millions of molecules to learn general-purpose representations, but their standard MLP output layers impose no physical constraints: vapour pressure predictions may violate monotonic temperature dependence, and viscosity curves may lack the functional form required by process simulators. Domain-informed approaches that do guarantee thermodynamic consistency have, by contrast, been limited to single properties and small datasets, and the existing multimodal molecular foundation models are typically aimed at biological-activity prediction rather than thermophysical properties. This work introduces MultiPUFFIN, a domain-informed multimodal foundation model that fills this gap. MultiPUFFIN fuses SMILES sequences, 2D molecular graphs and three-dimensional conformer geometries through bidirectional cross-modal attention and gated fusion, supplemented by auxiliary encoders for experimental conditions and molecular descriptors. The backbone is pretrained on 500{,}000 unlabelled PubChem molecules with three complementary self-supervised objectives, and a condition-aware refinement stack of five conditioners (T, pH, P, polymorph, method) routes per property to a four-head tournament that selects the best-performing thermodynamically-informed head per property. Two cross-property training-loss couplings (flash-point--vapor-pressure consistency and a normal-boiling-point anchor on the vapor-pressure head) softly enforce inter-property thermodynamic consistency. On a scaffold-split test set of \num{8877} molecules drawn from a multi-source dataset of \num{37968} unique molecules across nine thermophysical properties, MultiPUFFIN reaches an in-scope mean test $R^2 = 0.784$ and outperforms fine tuned ChemBERTa-2 on all nine properties despite training on roughly 2000$\times$ fewer labelled molecules; the largest margins are on the temperature-dependent properties (vapor pressure, viscosity) where ChemBERTa-2 has no mechanism to distinguish measurements at different temperatures. Per-property G+ applicability-domain filters explicitly bound the deployment scope. Systematic ablations quantify the contribution of each architectural component, of the choice of domain-informed equation per property, and of each coupling term.
\end{abstract}


\end{frontmatter}



\section{Introduction}
\label{sec:intro}

The prediction of physicochemical properties of small molecules underlies the design of separation processes, the formulation of pharmaceuticals, environmental fate modelling, and reaction engineering \cite{pang2023advanced}. Properties such as vapor pressure, viscosity, solubility, heat capacity and partition coefficients are conventionally estimated through empirical correlations whose parameters are component-specific: the Antoine equation for vapor pressure \cite{thomson1946}, the Andrade equation for viscosity \cite{andrade1930} and group contribution methods for boiling points \cite{nannoolal2008} all require dedicated fitted parameters per molecule, which restricts transferability and precludes large-scale screening.

Machine learning can help with this limitation by learning structure--property mappings directly from data \cite{das2024clm}. Graph neural networks treat atoms as nodes and bonds as edges, and have produced strong performance on quantum mechanical properties \cite{schutt2018schnet,gasteiger2020dimenet}, solvation free energies \cite{vermeire2021}, fuel ignition quality \cite{schweidtmann2020} and activity coefficients \cite{rittig2023gibbs}; chemical language models operating on SMILES strings offer competitive performance on property prediction tasks as well \cite{das2024clm}. The landscape of thermophysical property prediction in this body of work is governed by single-model, single-prediction architectures, occasionally extended with domain-informed prediction heads to recover physical consistency for one property at a time \cite{santana2024puffin,rebello2025expuffin}.

Foundation models extend the shallow learners discussed above by pretraining a single backbone on a large unlabelled corpus and adapting it through lightweight prediction heads \cite{jacsau2024perspective,natrevchemfm2025}. SMILES-BERT \cite{wang2019smilesbert}, ChemBERTa-2 \cite{ahmad2022chemberta2}, KPGT \cite{li2022kpgt}, MolCLR \cite{wang2022molclr}, MolE \cite{mendez2024mole} and the Uni-Mol family \cite{zhou2023unimol,lu2024unimol2,unimol3_2025} occupy this niche. The landscape of foundation models for molecules, despite this growth, is dominated by single-modal models, and the few multimodal members of the family target biological-activity prediction rather than thermophysical properties: MoleculeSTM \cite{liu2023moleculestm}, MoMu \cite{su2022momu} and GIT-Mol \cite{liu2023gitmol} align structural representations with natural language, while MolGT \cite{molgt2024}, FineMolTex \cite{li2024finemoltex} and MolPrompt \cite{molprompt2025} refine the alignment granularity. None of them combines multiple structural modalities with domain-informed prediction heads, and none of them does multi-task thermophysical property prediction. Table~\ref{tab:comparison_models} summarises this landscape: each row marks where the published architecture stands on the three orthogonal axes of multimodality, domain-informed heads and multi-task thermophysical coverage; the bottom row is the position MultiPUFFIN fills.

The single-modal limitation matters because each molecular representation carries a different slice of the underlying chemistry rather than the totality of it. The 2D molecular graph encodes topological connectivity and local functional-group environment, the SMILES string encodes long-range syntactic dependencies and substituent patterns \cite{weininger1988,baltruvsaitis2019multimodal}, and the 3D conformer encodes steric effects, solvation cavity geometry and through-space distances; no single representation contains all three. Any single-modal foundation model is therefore representationally bounded by the information present in the chosen modality, and the natural way to extend its representational reach is multimodal fusion.

Multimodality alone is not sufficient, however. The architectures cited above show that combining modalities does help on biological-activity tasks, but a property prediction trained against pure data, with no thermodynamic constraint baked into the output layer, may still violate elementary physics: vapor pressure can decrease with temperature, viscosity can increase with temperature, and Antoine coefficients can come out with unphysical signs. A multimodal architecture is one way to fuse domain expertise into the model inputs, and the gated cross-modal fusion of this work additionally exposes the implicit causality between modalities (Section~\ref{sec:fusion}). A complementary way is to fuse domain expertise into the model output or training objective: PUFFIN \cite{santana2024puffin} and ExPUFFIN \cite{rebello2025expuffin} introduced domain-informed \emph{inductive-bias neurons} that replace the standard output layer with a thermophysical equation whose coefficients are predicted by the network's penultimate layer; the network is never supervised on the coefficients, it learns them from property targets alone, and post-hoc inspection shows the recovered coefficients agree with the literature. The ExPUFFIN Andrade variant reduced viscosity RMSE by 37\% over the unconstrained baseline while delivering smooth monotonic viscosity--temperature curves. Both PUFFIN and ExPUFFIN are single-property, single-modality systems; the generalisation to a multi-property, multimodal foundation model with domain-informed heads has not been attempted.

A multimodal, multi-task and domain-informed model is exactly what molecular screening pipelines need at the pre-experimental stage. A medicinal chemist evaluating a drug candidate, a process engineer specifying an ester-based heat-transfer fluid, or a formulator selecting a solvent all need simultaneous, thermodynamically consistent estimates of multiple properties for the same molecule. The remainder of the paper presents \textbf{MultiPUFFIN} (Multimodal Path-Unifying Foundation Fusion Interfaced Network), a domain-informed multimodal foundation model that fills the empty bottom row of Table~\ref{tab:comparison_models} and supports this screening role.

\label{sec:gaps}

\begin{table*}[!htbp]
\centering
\caption{Comparison of molecular foundation models for property prediction. Modalities: S = SMILES (1D), G = 2D graph, 3D = 3D conformer, T = text, I = molecular image, Aux = auxiliary inputs (experimental conditions and molecular descriptors). Domain heads: whether the model employs thermodynamically-informed or domain-informed output layers (e.g., Antoine, Andrade, group contribution equations). Phys.\ consist.: whether physically consistent temperature dependence is enforced within individual properties by construction (note: this does not imply full cross-property thermodynamic consistency). Multi-task thermo.: whether the model simultaneously predicts multiple thermophysical properties in a single architecture.}
\label{tab:comparison_models}
\begin{adjustbox}{max width=\textwidth}
\begin{tabular}{lccccccc}
\toprule
\textbf{Model} & \textbf{Year} & \textbf{Modalities} & \textbf{Pretraining} & \makecell{\textbf{Domain}\\\textbf{heads}} & \makecell{\textbf{Multi-task}\\\textbf{thermo.}} & \makecell{\textbf{Phys.}\\\textbf{consist.}} & \makecell{\textbf{Pretrain}\\\textbf{data size}} \\
\midrule
SMILES-BERT \cite{wang2019smilesbert} & 2019 & S & MLM & \texttimes & \texttimes & \texttimes & ZINC subset \\
ChemBERTa-2 \cite{ahmad2022chemberta2} & 2022 & S & MLM + MTR & \texttimes & \texttimes & \texttimes & 77M mol. \\
KPGT \cite{li2022kpgt} & 2022 & G & SSL + knowledge & \texttimes & \texttimes & \texttimes & 2M mol. \\
MoMu \cite{su2022momu} & 2022 & G, T & Contrastive & \texttimes & \texttimes & \texttimes & 15K pairs \\
Uni-Mol \cite{zhou2023unimol} & 2023 & 3D & SSL (3D denoising) & \texttimes & \texttimes & \texttimes & 209M conf. \\
MoleculeSTM \cite{liu2023moleculestm} & 2023 & S, G, T & Contrastive & \texttimes & \texttimes & \texttimes & 281K pairs \\
MolE \cite{mendez2024mole} & 2024 & G & SSL + multi-task & \texttimes & \texttimes & \texttimes & 842M graphs \\
GIT-Mol \cite{liu2023gitmol} & 2024 & G, I, T & Multimodal LLM & \texttimes & \texttimes & \texttimes & 304K mol. \\
MolGT \cite{molgt2024} & 2024 & G, T & SSL multi-view & \texttimes & \texttimes & \texttimes & 250K mol. \\
Uni-Mol2 \cite{lu2024unimol2} & 2024 & 3D & SSL (scaled) & \texttimes & \texttimes & \texttimes & 800M conf. \\
MoleculeFormer \cite{moleculeformer2025} & 2025 & S, G & Supervised MT & \texttimes & \texttimes & \texttimes & 28 datasets \\
PUFFIN \cite{santana2024puffin} & 2024 & G & Transfer learn. & \checkmark & \texttimes & \checkmark & 6K mol. \\
ExPUFFIN \cite{rebello2025expuffin} & 2025 & G & Supervised & \checkmark & \texttimes & \checkmark & 3K mol. \\
\midrule
\textbf{MultiPUFFIN} (this work) & 2025 & \textbf{S, G, 3D, Aux} & \textbf{Multi-task} & \checkmark & \checkmark & \checkmark & \textbf{38K mol.} \\
\bottomrule
\end{tabular}
\end{adjustbox}
\end{table*}

The contributions of this work are: (i)~a multimodal backbone that fuses a Transformer over SMILES, a GCN over the 2D graph, and SchNet over the 3D conformer through bidirectional cross-modal attention and gated fusion, with auxiliary encoders for experimental conditions and molecular descriptors, so that each modality contributes its own slice of information to the unified embedding; (ii)~a generalised domain-informed inductive-bias mechanism that extends the PUFFIN / ExPUFFIN single-property paradigm to nine simultaneous thermophysical targets through a per-property four-head tournament (primary thermophysical equation, Joback group contribution, RDKit fragment counts, and an alternative thermodynamically-informed or direct FFNN baseline), enforcing intra-property physical consistency by construction; (iii)~a four-stage training strategy combining SSL pretraining on 500{,}000 unlabelled PubChem molecules, joint supervised multi-task training with a condition-aware refinement stack of five conditioners (T, pH, P, polymorph, method), backbone-unfrozen targeted fine tuning at very low learning rate, and per-property G+ applicability-domain evaluation that bounds the deployment scope. On a scaffold-split test set of 8{,}877 molecules drawn from a 37{,}968-molecule multi-source dataset (eleven public sources: OPERA, NIST ThermoML, ECHA REACH, ChEMBL, AqSolDB, FreeSolv, Bradley melting points, Sun et~al.\ flash points, ABB-ADD heat capacities, the Chew et~al.\ viscosity compilation, and PubChem), MultiPUFFIN reaches an in-scope mean test $R^2 = 0.784$ and beats fine tuned ChemBERTa-2 on all nine properties despite training on roughly 2000$\times$ fewer labelled molecules.

The remainder of this paper is organised as follows. Section~\ref{sec:methodology} presents the methodology, including data curation, model architecture and training. Section~\ref{sec:results} presents the experimental results and discussion. Section~\ref{sec:conclusions} concludes.


\section{Methodology}
\label{sec:methodology}

The overall MultiPUFFIN framework encompasses five components: a data curation and preprocessing pipeline, a self-supervised pretraining pipeline that produces a chemistry-aware backbone, the multimodal model architecture with a condition-aware embedding refinement stack (TConditioner module among five) and domain-informed prediction heads with cross-property physical coupling, a four-stage training strategy that culminates in a backbone-unfrozen targeted fine tune at very low learning rate, and a per-property applicability-domain (G+) evaluation that explicitly bounds the deployment regime. This section presents each in detail.

MultiPUFFIN is designed as a \textit{multimodal foundation model for thermophysical property prediction}. ``Foundation model'' is used in the now-standard sense established in vision--language and chemistry pretraining literature: a single model whose backbone is first pretrained on a large unlabeled molecular corpus through self-supervised objectives that do not require property labels, and whose representations are then fine tuned for downstream property regression through a small set of supervised signals. In MultiPUFFIN the backbone is multimodal: it jointly processes three fundamentally different structural data modalities (text, graph, and spatial structure) through modality-specific encoders, supplemented by two auxiliary encoders for experimental conditions and molecular descriptors, before fusing all representations into a shared molecular embedding. This design mirrors the broader paradigm of multimodal foundation models in machine learning (e.g., CLIP for vision--language, Gato for vision--language--action), where the central premise is that jointly learning from complementary data modalities produces richer representations than any single modality alone. In the molecular context, the three structural modalities correspond to: (i)~the \textit{text modality}, where a SMILES string is processed by a Transformer encoder as a chemical language sequence, capturing long-range syntactic dependencies and implicit chemical grammar; (ii)~the \textit{graph modality}, where the 2D molecular graph is processed by a GCN encoder through message-passing operations, capturing topological connectivity, ring systems, and local functional group patterns; and (iii)~the \textit{spatial modality}, where a 3D conformer is processed by a SchNet encoder through continuous-filter convolutions on interatomic distances, capturing through-space geometry, molecular shape, and steric effects. In addition, two auxiliary encoders process non-structural information: an \textit{experimental encoder} that embeds thermodynamic conditions (temperature, pressure) into the representation space, and a \textit{descriptor encoder} that incorporates precomputed molecular descriptors (molecular weight, topological polar surface area, hydrogen bond donors/acceptors, and other constitutional features).

Three additional components distinguish MultiPUFFIN from prior PUFFIN-family models and from generic multimodal foundation models for molecules. First, the backbone is pretrained with three complementary self-supervised objectives on an unlabeled molecular corpus before any property labels are used, so that the fused 512-dimensional embedding already encodes chemistry-aware features by the time the supervised multi-task training begins. Second, between the fused embedding and the prediction heads sits a stack of five identity-initialized condition-aware refinement modules (Section~\ref{sec:t_conditioner}): a TConditioner for measurement temperature, a PHConditioner for solution pH, a PressureConditioner for ambient pressure, a PolymorphEmbedding for crystal-form identifier, and a MethodFlag for test-protocol identifier. The modules are uniformly architectured as 2-token cross-attention blocks with zero-init residuals, route per property to the heads where the corresponding condition is physically relevant, and compose by chaining when multiple conditions matter (e.g., $\mathbf{u} \to \mathbf{u}_T \to \mathbf{u}_{T \to P}$ for viscosity and heat capacity). The stack lets the network learn arbitrary condition-dependent embedding modulations beyond what a fixed Antoine, Andrade, or Shomate functional form can express, and the identity-init property guarantees that any module that sees no condition variance in its training subset gracefully degrades to the unconditioned embedding. Third, the prediction heads are organized as a \textit{tournament} of candidate inductive biases per property, and the training loss carries explicit \textit{cross-property physical coupling terms} (flash-point--vapor-pressure consistency and a stronger normal-boiling-point anchor on the vapor-pressure head) that enforce thermodynamic relationships no purely data-driven foundation model can provide. The deployed model is a single forward pass of the four-component pipeline (SSL-pretrained backbone $\to$ condition-aware refinement stack with per-property routing $\to$ tournament-winning head per property $\to$ G+ scope check); no model-level ensembling is used at inference.

\begin{figure}[!htbp]
    \centering
    \includegraphics[width=0.85\textwidth]{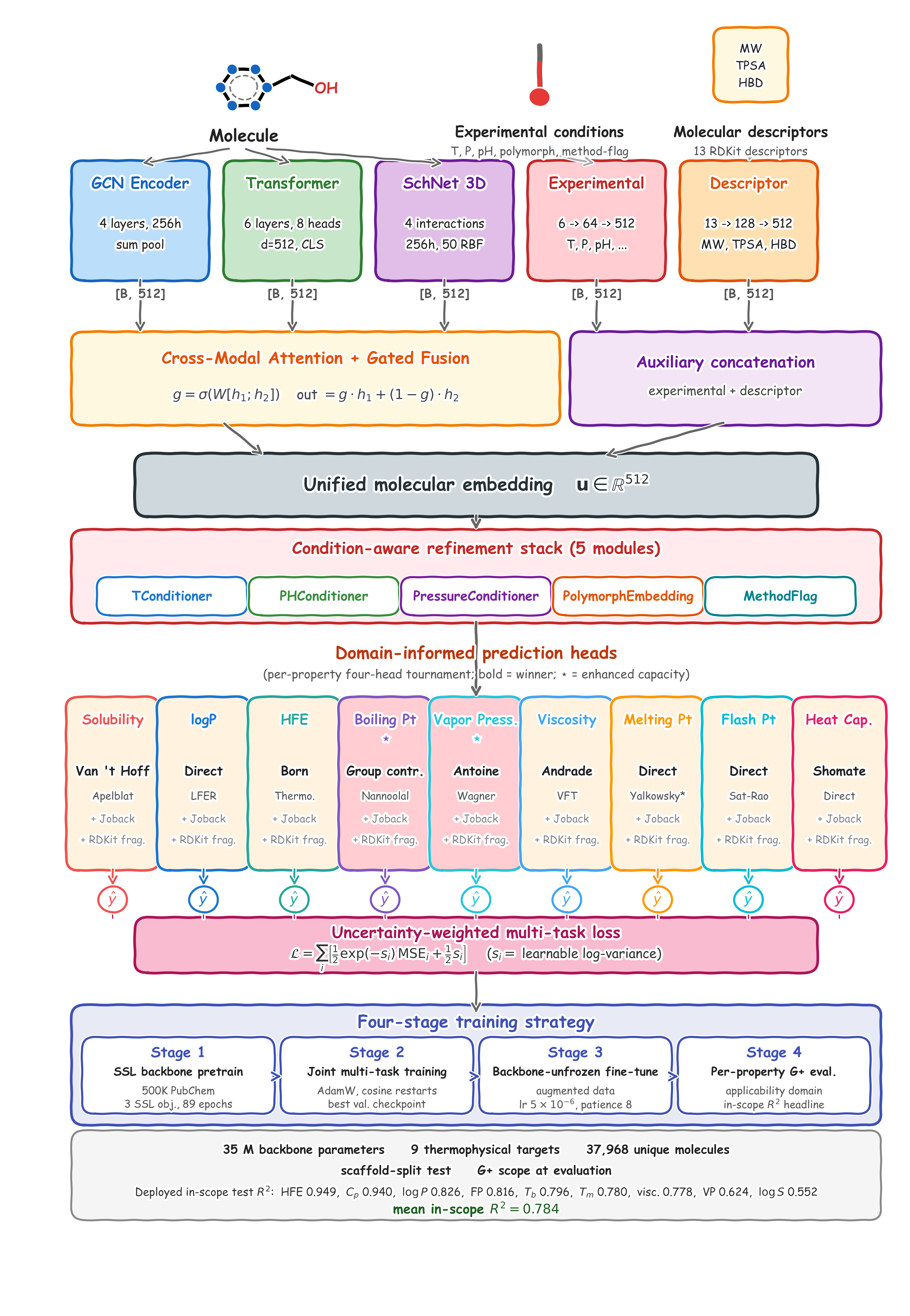}
    \caption{Architecture overview of MultiPUFFIN. Three structural encoders (GCN over the 2D graph, Transformer over the SMILES string, SchNet over the 3D conformer) and two auxiliary encoders (experimental conditions; molecular descriptors) form the multimodal backbone, pretrained on 500{,}000 unlabeled PubChem molecules with three SSL objectives. Cross-modal attention plus gated fusion produces a 512-dimensional unified embedding, which feeds a stack of five identity-initialized condition-aware refinement modules (T, pH, P, polymorph, method); each module is routed only to heads where the corresponding condition is physically relevant. Each property is then predicted by the winner of a four-head tournament (primary thermophysical equation, Joback, RDKit, alternative thermodynamically-informed/direct FFNN). Stars ($\star$) mark enhanced-capacity heads. Cross-property coupling (flash-point--vapor-pressure consistency; normal-boiling-point anchor) links the heads thermodynamically. The deployed model is a single forward pass; no model-level ensembling is used at inference.}
    \label{fig:architecture}
\end{figure}

Figure~\ref{fig:architecture} provides a schematic overview of the architecture. Bidirectional cross-modal attention between the GCN and Transformer branches enables each modality to attend to the other, and a learned sigmoid gate fuses the cross-attended representations. The SchNet embedding is incorporated through a separate geometry gate that can suppress the 3D contribution when conformer data is unavailable or unreliable. The outputs of the two auxiliary encoders (experimental conditions and molecular descriptors) are then concatenated with the fused structural embedding and projected to yield the final 512-dimensional unified embedding $\mathbf{u}$, which serves as the shared molecular representation from which all nine property-specific \textit{tournament heads} operate. Each tournament comprises four candidate inductive biases (Section~\ref{sec:prediction_heads}): a primary thermophysical equation, a Joback group-contribution head, an RDKit fragment-count head, and an alternative thermodynamically-informed or direct feed-forward head; the winning head per property is selected on the validation set. An uncertainty-weighted multi-task loss automatically balances the nine prediction objectives during training, and two cross-property physical coupling terms (Section~\ref{sec:coupling}) enforce thermodynamic consistency between flash point and vapor pressure, and between boiling point and vapor pressure.

\subsection{Data curation and dataset construction}
\label{sec:data}

A key component of this work is the assembly of a multi-property molecular dataset from diverse public sources. The final dataset comprises \num{37968} unique molecules with measurements across nine physicochemical properties, totaling \num{40904} data rows, since molecules with temperature-dependent properties contribute multiple rows at different temperatures. This dataset was assembled through a systematic multi-stage curation pipeline that integrates eleven public sources (OPERA, NIST ThermoML, ECHA REACH, ChEMBL, AqSolDB, FreeSolv, Bradley melting points, Sun et~al.\ flash points, ABB-ADD heat capacities, the Chew et~al.\ viscosity compilation, and PubChem), as summarized in Table~\ref{tab:data_sources}. Figure~\ref{fig:data_coverage} illustrates the highly heterogeneous data availability across properties and splits.

The dataset was constructed by merging an initial curated baseline with multiple external databases. The primary sources and their contributions are as follows.

AqSolDB and ESOL \cite{sorkun2019aqsoldb,wu2018moleculenet}. Aqueous solubility ($\log S$) data were aggregated from the AqSolDB database and the ESOL dataset from MoleculeNet, providing solubility values in $\log(\text{mol/L})$ at \SI{298.15}{\kelvin}, with temperature-dependent solubility data from PubChem included where available. After deduplication by canonical SMILES (retaining the median value for duplicate entries), the solubility dataset comprises \num{14085} unique molecules.

MoleculeNet Lipophilicity and ChEMBL \cite{wu2018moleculenet,gaulton2017chembl}. Octanol-water partition coefficients ($\log P$) from the Lipophilicity dataset in MoleculeNet and PubChem were merged with ChEMBL lipophilicity measurements into \num{11050} unique molecules. It should be noted that the ChEMBL data reports $\log D$ at pH~7.4, which is used here as a $\log P$ proxy. For non-ionizable molecules, $\log D \approx \log P$; for ionizable species, $\log D$ at physiological pH can differ from $\log P$ due to the fraction of the molecule in its ionized form. This approximation is standard in the cheminformatics community \cite{wu2018moleculenet} and is justified by the fact that the majority of drug-like molecules in these datasets are predominantly un-ionized at pH~7.4. Future work could improve the treatment of ionizable molecules by explicitly modeling the ionization equilibrium or by training on curated $\log P$-only datasets.

FreeSolv \cite{mobley2014freesolv}. Hydration free energies ($\Delta G_{\text{hyd}}$) were taken from the FreeSolv database, providing measurements in \si{\kilo\cal\per\mol} at \SI{298.15}{\kelvin} for \num{642} unique molecules.

OPERA 2.9 \cite{mansouri2018opera}. The OPERA (OPEn structure-activity/property Relationship App) models database contributed computed and validated measurements across multiple properties: $\log P$ (\num{4191} compounds), melting point (\num{13856} compounds, converted from \si{\celsius} to \si{\kelvin}), boiling point (\num{2426} compounds), vapor pressure (\num{3945} compounds, converted from $\log_{10}(\text{mmHg})$ to \si{\pascal}), and water solubility (\num{4674} compounds in $\log_{10}(\text{mol/L})$).

NIST ThermoML Archive \cite{frenkel2005thermoml}. The ThermoML archive, comprising \num{11923} structured data files from the NIST/TRC source data system, was systematically parsed to extract experimental thermophysical measurements. This yielded boiling point data for 220 molecules, melting point for \num{1363} molecules, vapor pressure for \num{1797} molecules (converted from \si{\kilo\pascal} to \si{\pascal}), viscosity for 527 molecules (converted from \si{\pascal\second} to \si{\milli\pascal\second}), and heat capacity for 744 molecules (in \si{\joule\per\mol\per\kelvin}). A dedicated parsing script was developed to extract property values, temperatures, and compound identifiers from the ThermoML XML schema, with InChI-to-SMILES conversion via RDKit for compound identification.

ECHA REACH \cite{echa2024reach}. The European Chemicals Agency (ECHA) Registration, Evaluation, Authorisation and Restriction of Chemicals (REACH) database was parsed for experimentally registered thermophysical data, contributing measurements of melting point, boiling point, vapor pressure, solubility, $\log P$, flash point, and viscosity for industrial chemicals.

Bradley Open Melting Point Dataset \cite{bradley2014melting}. Melting point data ($T_m$) for approximately \num{3041} compounds were obtained, with values converted from \si{\celsius} to \si{\kelvin} and filtered to the physically reasonable range of \SIrange{50}{1000}{\kelvin}.

Sun et al.\ Flash Point Dataset \cite{sun2019flashpoint}. Flash point data ($T_f$) for approximately \num{14696} entries with SMILES and flash points in Kelvin, filtered to \SIrange{100}{1000}{\kelvin}.

ABB-ADD Heat Capacity Dataset \cite{abb_add_cp}. Liquid heat capacity ($C_p$) data with multi-temperature measurements for approximately 968 compounds identified by InChI strings; InChI identifiers were converted to canonical SMILES via RDKit, and for each molecule the measurement closest to \SI{298.15}{\kelvin} was selected.

Chew et al.\ Viscosity Dataset \cite{chew2024viscosity}. An additional viscosity compilation of approximately \num{1005} compounds from the supplementary material of a recent Journal of Cheminformatics publication was integrated to enrich the viscosity training data.

PubChem \cite{kim2023pubchem}. The PubChem database served as the primary aggregation source for boiling point, vapor pressure, and viscosity data, with experimental measurements spanning different temperatures. Boiling point data were converted to Kelvin and filtered to \SIrange{100}{1000}{\kelvin}; vapor pressure data are stored as $\log_{10}(P/\text{Pa})$; viscosity data are stored as $\log_{10}(\eta/\text{mPa}{\cdot}\text{s})$, preserving multi-temperature measurements per molecule.

\begin{table*}[!htbp]
\centering
\caption{Summary of data sources and per-property coverage in the final curated dataset. Unique molecules indicates the total number of distinct chemical compounds with measurements for each property across all sources. The rightmost columns show the train/validation/test split sizes (unique molecules per property per split). The Stage-3 augmentation step (Section~\ref{sec:two_stage}) extends the per-property training rows but does not change the unique-molecule totals or test-split assignment reported here; the per-property test counts used in the deployment evaluation tables (Tables~\ref{tab:test_results} and \ref{tab:chemberta_comparison}) are the post-Stage-3 data-row counts and therefore differ from the unique-molecule counts in this table for the multi-temperature properties.}
\label{tab:data_sources}
\small
\begin{adjustbox}{max width=\textwidth}
\begin{tabular}{llccccc}
\toprule
\textbf{Property} & \textbf{Primary sources} & \textbf{Total mol.} & \textbf{Train} & \textbf{Val} & \textbf{Test} & \textbf{Units} \\
\midrule
Solubility ($\log S$) & AqSolDB, ESOL, OPERA, PubChem, ECHA & \num{14085} & \num{8606} & \num{2823} & \num{2656} & $\log(\text{mol/L})$ \\
$\log P$ & MoleculeNet, ChEMBL, OPERA, ECHA & \num{11050} & \num{8329} & \num{1360} & \num{1361} & -- \\
Hydration free energy & FreeSolv & 642 & 542 & 50 & 50 & \si{\kilo\cal\per\mol} \\
Boiling point ($T_b$) & PubChem, NIST, OPERA, ThermoML, ECHA & \num{8549} & \num{3776} & \num{2925} & \num{1848} & \si{\kelvin} \\
Vapor pressure & PubChem, NIST, OPERA, ThermoML, ECHA & \num{8956} & \num{5090} & \num{2244} & \num{1622} & $\log_{10}(\text{Pa})$ \\
Viscosity & PubChem, Chew et al., ThermoML, ECHA & \num{1901} & \num{1752} & 78 & 71 & $\log_{10}(\text{mPa}{\cdot}\text{s})$ \\
Melting point ($T_m$) & Bradley, OPERA, ThermoML, ECHA & \num{18915} & \num{10412} & \num{2491} & \num{6012} & \si{\kelvin} \\
Flash point ($T_f$) & Sun et al., ECHA & \num{10260} & \num{4508} & \num{3171} & \num{2581} & \si{\kelvin} \\
Heat capacity ($C_p$) & ABB-ADD, ThermoML & \num{1527} & \num{1380} & 78 & 69 & \si{\joule\per\mol\per\kelvin} \\
\midrule
\textbf{Total (unique molecules)} & & \textbf{\num{37968}} & \textbf{\num{22015}} & \textbf{\num{7076}} & \textbf{\num{8877}} & \\
\textbf{Total (data rows)} & & \textbf{\num{40904}} & \textbf{\num{24513}} & \textbf{\num{7254}} & \textbf{\num{9137}} & \\
\bottomrule
\end{tabular}
\end{adjustbox}
\end{table*}

All molecular structures are represented as canonical SMILES strings generated by RDKit \cite{rdkit}. A uniform quality control protocol was applied across all sources: each SMILES string is parsed and re-canonicalized, with entries that fail RDKit parsing discarded; multi-component molecules (salts, mixtures, identified by the presence of ``.'' in the SMILES) are removed; molecules are filtered to contain between 2 and 100 heavy atoms; for entries sharing the same canonical SMILES and temperature, the median property value is retained; and property values outside physically reasonable ranges (e.g., boiling points below \SI{100}{\kelvin} or above \SI{1000}{\kelvin}, negative viscosities, hydration free energies outside $[-30, +10]$~\si{\kilo\cal\per\mol}) are removed. For temperature-independent properties ($\log P$, $T_b$, $\Delta G_{\text{hyd}}$, $T_m$, $T_f$, $C_p$), values are associated with the measurement temperature or a default of \SI{298.15}{\kelvin}, while for temperature-dependent properties (solubility, vapor pressure, viscosity), each temperature--property pair constitutes an independent training sample.

Several temperature-independent properties are propagated across all rows of a given molecule. For example, if a molecule has vapor pressure measurements at ten different temperatures and a single boiling point value, the boiling point label is copied to all ten rows. This maximizes the multi-task learning signal per training example. An evaluation mask is maintained for each temperature-independent property to prevent artificial inflation of evaluation metrics: only one row per molecule (the one closest to \SI{298.15}{\kelvin}) is flagged for evaluation.

\label{sec:scarcity_audit}

A non-trivial design question for any multi-source dataset is whether the test distribution is genuinely covered by the training distribution within each property. The headline split-level statistics ($n_{\mathrm{train}} = 22{,}015$, $n_{\mathrm{test}} = 8{,}877$) hide structural imbalances at the per-(property $\times$ molecule-class) level: a property may be well-supported overall but have a single chemical class (e.g., salts, polyols, sulfonic acids) that is heavily represented in the test set without comparable training coverage. To make these imbalances visible, we built a per-(property, class) coverage matrix (Table~\ref{tab:scarcity_matrix}). The classes correspond to the seven chemistry exclusions encoded in the per-property G+ applicability-domain filters of Section~\ref{sec:applicability_domain} (salt, charged ion, polyol, sulfonic/phosphonic acid, very-large molecule, very-small molecule, zwitterion); these are the same classes for which a generic neutral-organic backbone is most likely to fail. Each cell reports the train and test counts. The audit informs two complementary actions taken in the deployed model: (i)~where targeted external data acquisition can close a gap (e.g., AqSolDB salts for solubility, brought in during the Stage~3 augmented fine tune), the gap is filled; (ii)~where no public condition-matched data was obtainable within the scope of this work (e.g., MNSol/DISSOLVE polyols for hydration free energy require institutional license access), the affected (property, class) cell is declared out-of-scope under the per-property G+ filter.

\begin{table*}[!htbp]
\centering
\caption{Per-(property $\times$ scarcity-class) coverage matrix on the training set, with the corresponding test counts in parentheses for transparency. Each cell is $n_{\mathrm{train}}~(n_{\mathrm{test}})$ and counts molecules of the indicated structural class for which the property is labeled. The classes correspond to the chemistry exclusions encoded in the per-property G+ applicability-domain filters of Section~\ref{sec:applicability_domain}. Where the training count is small relative to the test count for a given (property, class) cell, the class is a candidate for either targeted data augmentation (e.g., AqSolDB salts for solubility, which closes the salt cell) or for declaration as out-of-scope under the per-property G+ filter (e.g., HFE polyols, where MNSol/DISSOLVE coverage is access-restricted).}
\label{tab:scarcity_matrix}
\small
\begin{adjustbox}{max width=\textwidth}
\begin{tabular}{lrrrrrrrr}
\toprule
\textbf{Property} & \textbf{salt} & \textbf{ion} & \textbf{polyol} & \textbf{sulf./phos.} & \textbf{very-large} & \textbf{very-small} & \textbf{zwitt.} & \textbf{total} \\
\midrule
$\log S$ & 962~(0) & 376~(76) & 109~(29) & 752~(102) & 405~(3) & 246~(33) & 1287~(306) & 9978~(2700) \\
$\log P$ & -- & 86~(19) & 31~(13) & 254~(70) & 192~(2) & 377~(136) & 349~(120) & 9975~(1570) \\
$\Delta G_{\rm hyd}$ & -- & -- & 2~(2) & -- & -- & 292~(140) & 52~(4) & 1862~(269) \\
$T_b$ & 199~(4) & 28~(7) & 36~(5) & 129~(20) & 66~(1) & 465~(142) & 318~(122) & 7443~(2106) \\
VP & 420~(0) & 51~(9) & 68~(5) & 194~(27) & 174~(2) & 235~(30) & 589~(144) & 8808~(1651) \\
$\eta$ & 7~(0) & 3~(0) & 34~(3) & 17~(0) & 10~(0) & 868~(141) & 146~(2) & 7821~(331) \\
$T_m$ & -- & 77~(16) & 91~(21) & 185~(66) & 220~(4) & 402~(105) & 682~(733) & 14938~(6206) \\
FP & 63~(8) & 15~(14) & 48~(7) & 27~(19) & 29~(2) & 469~(140) & 441~(287) & 10700~(2840) \\
$C_p$ & -- & 1~(0) & 65~(3) & 6~(0) & 15~(0) & 339~(139) & 86~(2) & 3900~(297) \\
\bottomrule
\end{tabular}
\end{adjustbox}
\end{table*}

The dataset is split at the molecule level into training (\SI{80}{\percent}), validation (\SI{10}{\percent}), and test (\SI{10}{\percent}) sets using a hybrid strategy that balances two competing objectives: (i)~evaluating the model's ability to generalize to structurally novel molecules, and (ii)~ensuring sufficient representation of rare properties in evaluation sets. The rationale behind this split design is that a simple random partition, while maximizing statistical power, would underestimate the generalization challenge because structurally similar molecules (e.g., members of the same homologous series) would appear in both training and test sets. Conversely, a purely scaffold-based split would provide the most rigorous generalization test but could leave data-scarce properties (hydration free energy, viscosity, heat capacity) with too few test samples for reliable evaluation. The hybrid strategy resolves this tension as follows. For molecules possessing only common properties (solubility, $\log P$, boiling point, vapor pressure, melting point, flash point), a scaffold-based split \cite{wu2018moleculenet} is employed: Murcko scaffolds (the core ring systems of each molecule) are computed via RDKit, and entire scaffold groups are assigned to the same partition to ensure that the test set evaluates generalization to structurally novel chemical series rather than interpolation within familiar scaffolds. For molecules possessing rare properties (hydration free energy, viscosity, heat capacity), a greedy assignment strategy ensures that each property has at least 50 unique molecules in both the validation and test sets, preventing the evaluation from being dominated by statistical noise. All rows corresponding to a given molecule (i.e., measurements at different temperatures) are assigned to the same split, preventing data leakage between partitions. This hybrid approach yields the final split sizes of \num{22015} training, \num{7076} validation, and \num{8877} test molecules (Table~\ref{tab:data_sources}).

The two-pass procedure is summarized in Algorithm~\ref{alg:hybrid_split}. Pass~1 reserves data-scarce property samples first, so that subsequent scaffold-based assignment cannot accidentally drain those properties below the 50-molecule per-split floor. Pass~2 then assigns the remaining (common-property) molecules by their Murcko scaffold, with whole scaffold groups routed to a single split.

\begin{algorithm}[!htbp]
\caption{Hybrid scaffold + property-balanced splitting.}
\label{alg:hybrid_split}
\begin{algorithmic}[1]
\Require Molecule set $\mathcal{M}$ with property-availability mask $\mathbf{p}_m \in \{0,1\}^9$ for each $m \in \mathcal{M}$
\Require Scarce property set $\mathcal{P}_{\text{scarce}} = \{\Delta G_{\text{hyd}}, \eta, C_p\}$
\Require Per-split minimum count $n_{\min} = 50$ per scarce property; target ratios $(0.80, 0.10, 0.10)$ for (train, val, test)
\Ensure Disjoint partitions $\mathcal{M}_{\text{train}}, \mathcal{M}_{\text{val}}, \mathcal{M}_{\text{test}}$
\Statex
\State \emph{Pass~1: greedy scarce-property assignment}
\State $\mathcal{M}_{\text{val}} \gets \emptyset$;~~$\mathcal{M}_{\text{test}} \gets \emptyset$
\For{each property $p \in \mathcal{P}_{\text{scarce}}$}
  \State $\mathcal{M}_p \gets \{m \in \mathcal{M} : \mathbf{p}_m[p] = 1\}$ \Comment{molecules with property $p$}
  \State Shuffle $\mathcal{M}_p$ with fixed seed
  \While{$|\mathcal{M}_{\text{val}} \cap \mathcal{M}_p| < n_{\min}$}
    \State move next $m$ from $\mathcal{M}_p$ to $\mathcal{M}_{\text{val}}$
  \EndWhile
  \While{$|\mathcal{M}_{\text{test}} \cap \mathcal{M}_p| < n_{\min}$}
    \State move next $m$ from $\mathcal{M}_p$ to $\mathcal{M}_{\text{test}}$
  \EndWhile
\EndFor
\Statex
\State \emph{Pass~2: scaffold-grouped assignment for the remainder}
\State $\mathcal{M}_{\text{rest}} \gets \mathcal{M} \setminus (\mathcal{M}_{\text{val}} \cup \mathcal{M}_{\text{test}})$
\State Compute Murcko scaffold $s(m)$ for each $m \in \mathcal{M}_{\text{rest}}$ \Comment{RDKit}
\State Group $\mathcal{M}_{\text{rest}}$ by scaffold: $\{\mathcal{G}_s\}_{s \in \mathcal{S}}$
\State Order scaffolds $\mathcal{S}$ by descending $|\mathcal{G}_s|$ \Comment{largest scaffolds placed first}
\For{each scaffold group $\mathcal{G}_s$}
  \State Assign \emph{entire} $\mathcal{G}_s$ to the split currently furthest below its target ratio
\EndFor
\Statex
\State \emph{Pass~3: row-level consolidation}
\For{each molecule $m$ assigned to split $\sigma \in \{\text{train}, \text{val}, \text{test}\}$}
  \State Route \emph{all} measurement rows of $m$ (across temperatures and conditions) to $\sigma$
\EndFor
\State \Return $\mathcal{M}_{\text{train}}, \mathcal{M}_{\text{val}}, \mathcal{M}_{\text{test}}$
\end{algorithmic}
\end{algorithm}

\label{sec:conformers}

Three-dimensional molecular conformers are required as input to the SchNet geometry encoder. The inclusion of explicit three-dimensional geometry is motivated by the fact that many thermophysical properties are governed by spatial effects that cannot be inferred from two-dimensional topology alone: vapor pressure depends on molecular surface area and shape, viscosity on molecular packing and flow resistance, and hydration free energy on the three-dimensional cavity that a solute carves in the solvent. The two-dimensional graph encodes \textit{which} atoms are bonded but not \textit{how far apart} non-bonded atoms are in space, nor the dihedral angles that define rotational conformations. By providing the SchNet encoder with explicit Cartesian coordinates, the model gains access to through-space distances, molecular shape descriptors, and steric information that complement the topological and sequential features captured by the GCN and Transformer encoders, respectively.

Conformers are precomputed for all \num{37968} unique molecules and cached for efficient training. Each canonical SMILES is converted to an RDKit molecule object with explicit hydrogens, and a three-dimensional embedding is generated using the ETKDGv3 algorithm \cite{riniker2015etkdg} with a fixed random seed for reproducibility. If ETKDGv3 fails (approximately \SI{1}{\percent} of molecules), a fallback with random initial coordinates is attempted. The embedded conformer is subsequently optimized using the MMFF94 force field \cite{halgren1996merck} for up to 200 iterations, after which explicit hydrogens are removed to yield atomic numbers and Cartesian coordinates consistent with the heavy-atom graph representation. As a final fallback for molecules where 3D embedding fails entirely, two-dimensional coordinates are computed and extended with $z = 0$. Full coverage (\SI{100}{\percent}) was achieved across the dataset.

A notable limitation of this approach is that a single static conformer from force-field optimization cannot represent the Boltzmann ensemble of conformations that a flexible molecule populates in solution. For rigid aromatic compounds, the single-conformer approximation is reasonable, but for molecules with many rotatable bonds (e.g., long-chain alkanes, polyethers), the properties of interest (particularly viscosity and solvation free energy) may depend on the conformational distribution rather than any single geometry. The geometry gate mechanism (Section~\ref{sec:fusion}) partially mitigates this limitation by allowing the model to suppress the 3D contribution for molecules where the single conformer is unrepresentative, but this is an imperfect solution. Future work could address this by either (i)~generating multiple conformers per molecule and aggregating their SchNet representations, following the multi-conformer ensemble approach of Uni-Mol \cite{zhou2023unimol}, or (ii)~incorporating conformer generation uncertainty as an additional input feature. 

\subsection{Self-supervised pretraining of the molecular backbone}
\label{sec:ssl_pretrain}

The multimodal backbone is pretrained through a self-supervised objective on a large unlabeled molecular corpus before any property labels are introduced. The rationale is the same as for foundation models in natural language processing and vision: labeled thermophysical data are scarce (tens of thousands of molecules per property at best), while unlabeled molecular data; drug-like compounds, commercially available chemicals, and structures catalogued in public repositories; are available in the tens to hundreds of millions. A backbone that has been exposed to a broad chemical distribution before it sees any property labels starts supervised training from a chemistry-aware initialization rather than a random one, and this initialization is carried through every downstream prediction head.

A corpus of \num{500000} drug-like and industrial-chemical SMILES strings was assembled from PubChem \cite{kim2023pubchem} by filtering to molecules with 5--60 heavy atoms, passing RDKit canonicalization, and excluding charged or multi-component entries. The corpus is intentionally disjoint from the property-labeled training data: molecules that appear in any of the nine property datasets are excluded from the pretraining corpus to prevent information leakage from property labels into the pretraining objective.

The backbone is pretrained with three complementary self-supervised signals that together expose the GCN, Transformer and SchNet encoders to invariance, structural-recovery and language-recovery pressure before any thermophysical label is seen.

For each molecule $m$ in a mini-batch of size $B$, two chemically equivalent views are constructed: the canonical SMILES $s^c_m$ and a randomly enumerated SMILES $s^r_m$ produced by RDKit's randomised traversal. Each view is independently routed through the multimodal backbone, and the resulting unified embeddings are $\ell_2$-normalised to unit length, $\mathbf{z}^c_m, \mathbf{z}^r_m \in \mathbb{S}^{d-1}$ with $d = 512$. The contrastive InfoNCE \cite{oord2018cpc} loss for the batch is
\begin{equation}
\mathcal{L}_{\mathrm{NCE}} \;=\; -\frac{1}{B}\sum_{m=1}^{B} \log
\frac{\exp\!\big(\mathbf{z}^c_m \cdot \mathbf{z}^r_m / \tau\big)}
{\sum_{j=1}^{B} \exp\!\big(\mathbf{z}^c_m \cdot \mathbf{z}^r_j / \tau\big)},
\label{eq:infonce}
\end{equation}
where $\tau = 0.07$ is the temperature parameter and the denominator runs over the $B-1$ in-batch negatives plus the positive at $j = m$. The objective drives the fused embedding to be insensitive to the SMILES traversal order while remaining discriminative across distinct molecules.

Let $\mathbf{X}_m \in \mathbb{R}^{N_m \times 39}$ denote the atom-feature matrix of molecule $m$ (Section~\ref{sec:featurization}), where the first 11 dimensions of each row encode the atomic-number one-hot. A masking set $\mathcal{M}_m \subset \{1, \ldots, N_m\}$ is sampled uniformly at random with $|\mathcal{M}_m| / N_m \approx 0.15$, and the atomic-number block of each masked row is replaced by a learnable $[\mathrm{MASK}]$ vector while every other feature dimension is preserved. The masked features are propagated through the GCN encoder, and a small two-layer auxiliary head $g_\theta : \mathbb{R}^{256} \to \mathbb{R}^{11}$ is trained on the penultimate GCN layer to recover the original atomic-number distribution. The mean-squared-error reconstruction loss is
\begin{equation}
\mathcal{L}_{\mathrm{MAR}} \;=\; \frac{1}{B}\sum_{m=1}^{B}\frac{1}{|\mathcal{M}_m|}
\sum_{i \in \mathcal{M}_m} \big\| g_\theta\!\big(\mathbf{h}_{m,i}\big) - \mathbf{x}_{m,i}^{(\mathrm{atom})}\big\|_2^2,
\label{eq:mar}
\end{equation}
where $\mathbf{h}_{m,i}$ is the masked-input GCN representation of atom $i$ and $\mathbf{x}_{m,i}^{(\mathrm{atom})}$ is the original 11-dimensional atomic-number target. The objective forces the GCN to use bond context (neighbour atomic numbers, hybridisation, ring membership) to infer the identity of the masked atom, which is the graph analogue of BERT's masked-language modelling \cite{devlin2019bert} adapted to chemistry.

Independently from the graph masking above, a masking set $\mathcal{S}_m$ of token positions in the SMILES sequence $s^c_m$ is sampled uniformly with $|\mathcal{S}_m| / |s^c_m| \approx 0.15$ and replaced by a learnable $[\mathrm{MASK}]$ token. The full sequence is processed by the SMILES Transformer (Section~\ref{sec:transformer_encoder}) and a tied output projection $W_{\mathrm{out}} \in \mathbb{R}^{|V| \times 512}$ produces a logit vector at every position, where $|V| = 50$ is the vocabulary size. The cross-entropy loss is
\begin{equation}
\mathcal{L}_{\mathrm{MTP}} \;=\; -\frac{1}{B}\sum_{m=1}^{B}\frac{1}{|\mathcal{S}_m|}
\sum_{t \in \mathcal{S}_m} \log P\!\big(s^c_{m,t} \,\big|\, \tilde{s}^c_m;\, \theta\big),
\label{eq:mtp}
\end{equation}
where $\tilde{s}^c_m$ is the masked sequence and $P(\cdot \mid \tilde{s}^c_m; \theta)$ is the softmax over the Transformer's output logits at the masked position.

The three terms are combined with unit weights
\begin{equation}
\mathcal{L}_{\mathrm{SSL}} \;=\; \mathcal{L}_{\mathrm{NCE}} \;+\; \mathcal{L}_{\mathrm{MAR}} \;+\; \mathcal{L}_{\mathrm{MTP}}.
\label{eq:ssl_total}
\end{equation}
Equal weighting is justified empirically: (i)~the three losses live on comparable numerical scales after the first warm-up epoch (each in the range $0.5$--$3$); (ii)~weight searches over the simplex $\{(\alpha_{\mathrm{NCE}}, \alpha_{\mathrm{MAR}}, \alpha_{\mathrm{MTP}}) : \alpha_i \geq 0, \sum_i \alpha_i = 3\}$ within the $\{0.5, 1.0, 1.5, 2.0\}^3$ grid did not yield a downstream improvement larger than $\Delta R^2 \approx 0.005$ on the validation set, well within the seed-to-seed noise of multi-task training; and (iii)~unit weights preserve the interpretability that each modality is exposed to one supervisory signal of comparable strength. Temperature-dependent auxiliary inputs are clamped to a reference state ($T = \SI{298.15}{\kelvin}$, $P = \SI{101325}{\pascal}$) throughout pretraining so that the backbone learns chemistry-aware representations first; the experimental encoder is fine tuned only during the supervised stage. Gradients propagate through the entire backbone (GCN + Transformer + SchNet + cross-modal fusion) for all three terms, ensuring that the fused embedding is jointly shaped by invariance, atom-recovery and language-recovery pressure rather than dominated by any single modality.

Pretraining runs for 89 epochs (paused before the 100-epoch budget after validation-loss plateau) with AdamW ($\alpha = 5 \times 10^{-4}$, weight decay $1 \times 10^{-5}$), batch size 256, linear warmup over the first 2 epochs, and cosine decay thereafter. The pretrained backbone weights (GCN encoder, Transformer encoder, SchNet encoder, cross-modal fusion module) are checkpointed at the end of pretraining and then used as the initialization for the supervised multi-task training described in Section~\ref{sec:training}. The auxiliary encoders (experimental conditions and molecular descriptors), the condition module stack (Section~\ref{sec:t_conditioner}), and the nine property-specific prediction heads are initialized randomly at the start of supervised training and learned from scratch.

Ablation against a randomly initialized backbone (Section~\ref{sec:ablation_architecture}) confirms that self-supervised pretraining is the single largest non-architectural contributor to downstream performance, on the order of $+0.03$ in mean test $R^2$. The improvement is largest on data-scarce properties (hydration free energy, viscosity, heat capacity, solubility), where the chemistry-aware initialization compensates for the limited supervised training signal available per task.

\subsection{Molecular featurization}
\label{sec:featurization}

Molecular featurization translates the raw chemical structure into numerical representations suitable for neural network processing. The featurization scheme is designed to encode the chemical identity, local bonding environment, and electronic properties of each atom and bond in a compact, fixed-dimensional vector. These features serve as the input to the GCN encoder (atom and bond features) and the SMILES Transformer encoder (tokenized sequences), while the SchNet encoder operates directly on atomic numbers and Cartesian coordinates from the conformer generation step. The choice and granularity of features directly affect the information available to the model: too few features limit the model's ability to distinguish chemically distinct environments, while excessive dimensionality can dilute the learning signal and increase computational cost.

Each atom in the molecular graph is represented by a 39-dimensional feature vector. This vector encodes the atomic number as an 11-dimensional one-hot vector over the ten most common elements (C, N, O, F, P, S, Cl, Br, I, Si) plus an ``other'' category; the degree (number of bonded neighbors) as a 7-dimensional one-hot encoding ($d \in \{0, \ldots, 6\}$); formal charge as a 5-dimensional one-hot encoding ($q \in \{-2, \ldots, +2\}$); the number of attached hydrogens (5 dimensions, $n_H \in \{0, \ldots, 4\}$); hybridization state as a 5-dimensional one-hot encoding over sp, sp$^2$, sp$^3$, sp$^3$d, and sp$^3$d$^2$; a binary aromaticity indicator; a binary ring membership indicator; and chirality as a 4-dimensional one-hot encoding over unspecified, $R$, $S$, and other.

Each bond is represented by a 12-dimensional feature vector comprising a 4-dimensional one-hot encoding of bond type (single, double, triple, aromatic), a binary conjugation indicator, a binary ring membership indicator, and a 6-dimensional one-hot encoding of stereo configuration (none, $Z$, $E$, cis, trans, other).

SMILES strings are tokenized at the character level using a vocabulary of 50 tokens, including special tokens for padding, start-of-sequence, end-of-sequence, and unknown characters, alongside 46 chemistry tokens corresponding to atoms, bonds, brackets, and ring-closure digits. Two-character tokens (Cl, Br, Si, Se, Te) are handled with priority matching. Sequences are padded or truncated to a maximum length of 256 tokens.

\subsection{Dual-encoder rationale}
\label{sec:dual_encoder_rationale}

The MultiPUFFIN encoder architecture deliberately processes each molecular modality through a separate, specialized encoder before fusing the resulting embeddings, rather than pre-aligning representations (as in contrastive approaches such as MoleculeSTM \cite{liu2023moleculestm}) or encoding a single modality (3D conformers) through a unified backbone (as in Uni-Mol \cite{zhou2023unimol}, which processes only 3D molecular geometry). This design is motivated by two key observations.

First, the three molecular representations (2D graphs, 1D SMILES strings, and 3D conformers) encode fundamentally different and complementary aspects of molecular identity. The molecular graph captures topological connectivity, ring systems, and local functional group patterns through its adjacency structure; the SMILES string encodes the same molecule as a linear sequence in which long-range syntactic dependencies (e.g., ring-closure digits linking distant atoms) provide a complementary inductive bias that differs from the spatial locality inherent to message passing; and the 3D conformer captures through-space distances, dihedral angles, and steric effects that are entirely absent from the 2D representation. A molecule's thermophysical properties are governed by the combination of all three: topology determines the presence of hydrogen-bonding groups, sequential patterns capture substituent effects on electron density, and three-dimensional shape governs molecular packing and solvation. By processing each modality through a dedicated encoder optimized for that representation's structure, the model preserves modality-specific features that would be lost or diluted in a shared-encoder architecture.

Second, modality-specific encoders enable graceful degradation when one modality is unavailable. In practice, 3D conformers may fail to generate for certain molecules or may be unreliable for flexible macrocycles; by keeping the 3D branch separate, the model can fall back to the 2D/1D fusion without architectural modification, using the geometry gate (Section~\ref{sec:fusion}) to automatically suppress the missing modality's contribution.

\subsection{Graph convolutional network encoder}
\label{sec:gcn_encoder}

The GCN encoder processes the two-dimensional molecular graph and contributes \emph{topological} information to the unified embedding: which atoms are bonded to which, the local neighbourhood that each atom sits in, and the additive functional-group structure that drives bulk thermophysical properties. Formally, a molecule is represented as an undirected graph $G = (V, E)$ where the nodes $V$ correspond to atoms (the node feature matrix is $\mathbf{X} \in \mathbb{R}^{N \times 39}$, encoding atomic number, hybridisation, formal charge and stereochemistry) and the edges $E$ correspond to covalent bonds (with bond-type and ring-membership features). The graph is processed through $L_g = 4$ Kipf--Welling graph convolutional layers \cite{kipf2017},
\begin{equation}
    \mathbf{H}^{(l+1)} = \sigma\!\left(\tilde{\mathbf{D}}^{-1/2}\,\tilde{\mathbf{A}}\,\tilde{\mathbf{D}}^{-1/2}\,\mathbf{H}^{(l)}\,\mathbf{W}^{(l)}\right),
    \label{eq:gcn_layer}
\end{equation}
where $\tilde{\mathbf{A}} = \mathbf{A} + \mathbf{I}_N$ is the adjacency matrix with self-loops, $\tilde{\mathbf{D}}$ is the corresponding degree matrix, $\mathbf{W}^{(l)} \in \mathbb{R}^{d_l \times d_{l+1}}$ are learnable weights and $\sigma$ is the ReLU non-linearity; the hidden dimension is $d_g = 256$ at every layer. The information that this operator extracts is best read off the unrolled message-passing equation: at layer $l+1$, the embedding of atom $i$ becomes
\begin{equation}
    \mathbf{h}_i^{(l+1)} \;=\; \sigma\!\left(\sum_{j \in \mathcal{N}(i) \cup \{i\}} \frac{1}{\sqrt{\tilde{d}_i\,\tilde{d}_j}}\;\mathbf{W}^{(l)}\,\mathbf{h}_j^{(l)}\right),
    \label{eq:gcn_msg}
\end{equation}
so that after $L_g$ layers each atom embedding aggregates information from a neighbourhood of radius up to $L_g$ bonds. The fixed receptive field of $L_g = 4$ bonds matches the topological scale at which functional-group additivity (Joback-style group contribution) operates, which is precisely the regime where conventional thermodynamic correlations are most predictive. Long-range substituent effects beyond four bonds, by construction, must be supplied by the SMILES Transformer or the 3D SchNet branches; this is the principled basis for the multimodal split.

Each GCN layer is followed by GraphNorm \cite{cai2021graphnorm} and BatchNorm \cite{ioffe2015batchnorm} for training stability, with dropout $p = 0.15$. Residual connections \cite{he2016deep} are used at every layer (the first layer applies a linear projection from $39$ to $256$, subsequent layers add directly). The graph-level molecular embedding is obtained by sum pooling over all node representations, $\mathbf{h}_{\text{graph}} = \sum_{i=1}^{N} \mathbf{h}_i^{(L_g)}$, followed by a two-layer projection
\begin{equation}
    \mathbf{z}_{\text{graph}} = \mathbf{W}_2\,\text{ReLU}\!\left(\text{Dropout}\!\left(\mathbf{W}_1\,\mathbf{h}_{\text{graph}} + \mathbf{b}_1\right)\right) + \mathbf{b}_2 \;\in\; \mathbb{R}^{512},
\end{equation}
into the shared 512-dimensional embedding space. Sum pooling rather than mean pooling is chosen so that the graph-level embedding is sensitive to molecular size, which is itself a primary driver of boiling point, vapour pressure and viscosity.

\subsection{SMILES Transformer encoder}
\label{sec:transformer_encoder}

The SMILES Transformer encoder contributes \emph{long-range syntactic} information to the unified embedding: substituent patterns that span tens of tokens, ring-closure couplings that define macrocyclic topology, and stereochemistry / valence / formal-charge cues that the SMILES grammar carries explicitly but that the 2D graph carries only through node features. The Transformer is the natural complement to the GCN: where the GCN sums local neighbourhoods of bounded radius, self-attention can directly relate any pair of tokens at arbitrary sequence distance, with attention weights $\alpha_{ij}^{(h)} = \mathrm{softmax}_j\!\left(\mathbf{q}_i^{(h)} \cdot \mathbf{k}_j^{(h)} / \sqrt{d_k}\right)$ at head $h$. The information added to the unified embedding by this branch is the part of the structure--property mapping that depends on long-range substituent context (electron-withdrawing groups across rings, conjugation chains, position-dependent grammar of SMILES enumerations), which is precisely the regime where the GCN's bounded receptive field is silent.

Each token $t_i$ is mapped to a learned embedding vector $\mathbf{e}_i \in \mathbb{R}^{512}$, scaled by $\sqrt{d_{\text{model}}}$, and augmented with sinusoidal positional encodings \cite{vaswani2017attention}:
\begin{equation}
    \text{PE}_{(\text{pos}, 2i)} = \sin\left(\frac{\text{pos}}{10000^{2i/d_{\text{model}}}}\right), \quad \text{PE}_{(\text{pos}, 2i+1)} = \cos\left(\frac{\text{pos}}{10000^{2i/d_{\text{model}}}}\right)
\end{equation}

The sequence is processed through $L_s = 6$ pre-norm Transformer encoder layers \cite{vaswani2017attention}, each comprising multi-head self-attention with $h = 8$ heads (key dimension $d_k = 64$) and a feed-forward sublayer with GELU activation \cite{hendrycks2016gaussian}, a model dimension $d_{\text{model}} = 512$, and a feed-forward dimension $d_{\text{ff}} = 2048$. Pre-norm architecture (LayerNorm applied before the attention and feed-forward sublayers) is used for improved training stability \cite{xiong2020prenorm}, with dropout ($p = 0.1$) applied within both sublayers. The molecule-level representation is obtained by extracting the hidden state corresponding to the start-of-sequence token (the CLS-equivalent token at position 0), followed by a two-layer output projection with GELU activation mapping to $\mathbf{z}_{\text{SMILES}} \in \mathbb{R}^{512}$.

\subsection{SchNet 3D geometry encoder}
\label{sec:schnet}

The SchNet encoder \cite{schutt2018schnet} introduces the third modality: three-dimensional molecular geometry. The inclusion of a dedicated 3D encoder is motivated by the fact that many thermophysical properties are fundamentally governed by spatial effects that cannot be inferred from 2D topology alone. Vapor pressure and boiling point depend on intermolecular interactions whose strength is modulated by molecular shape and surface accessibility; viscosity is governed by molecular packing and the ease of flow past neighboring molecules; and hydration free energy reflects the three-dimensional cavity that a solute carves in the solvent. While the 2D graph captures which atoms are bonded and the SMILES string encodes this connectivity sequentially, neither representation encodes through-space distances between non-bonded atoms, dihedral angles that define rotational conformations, or the overall three-dimensional shape of the molecule. SchNet was selected for this branch because its continuous-filter convolution framework naturally handles the continuous and rotationally invariant nature of interatomic distances, and because it has demonstrated strong performance on quantum-chemical property prediction benchmarks \cite{schutt2018schnet}. The encoder operates on atomic numbers and Cartesian coordinates produced by the conformer generation pipeline (Section~\ref{sec:conformers}).

Each atom is embedded by its atomic number into a learned vector $\mathbf{e}_i \in \mathbb{R}^{256}$ via an embedding lookup over 101 possible atomic numbers. Interatomic distances $d_{ij} = \|\mathbf{r}_i - \mathbf{r}_j\|$ are computed for all atom pairs within a cutoff radius $r_c = \SI{10}{\angstrom}$ and expanded into $K = 50$ radial basis functions using Gaussian smearing:
\begin{equation}
    e_k(d_{ij}) = \exp\left(-\frac{1}{2\sigma^2}(d_{ij} - \mu_k)^2\right), \quad k = 1, \ldots, K
\end{equation}
where $\{\mu_k\}_{k=1}^{K}$ are evenly spaced centers from 0 to $r_c$ and $\sigma$ is the spacing between centers.

The encoder applies $L_{\text{int}} = 4$ interaction blocks, each performing continuous-filter convolution:
\begin{equation}
    \mathbf{x}_i^{(l+1)} = \mathbf{x}_i^{(l)} + \text{UpdateMLP}\left(\sum_{j \in \mathcal{N}(i)} \mathbf{W}^{(l)}\left(\mathbf{e}(d_{ij})\right) \odot \mathbf{x}_j^{(l)}\right)
    \label{eq:schnet_interaction}
\end{equation}
where $\mathcal{N}(i) = \{j : d_{ij} < r_c, j \neq i\}$ is the neighborhood of atom $i$, $\mathbf{W}^{(l)}(\cdot)$ is a filter-generating network (a two-layer MLP with SiLU activation that maps the 50-dimensional Gaussian-expanded distance to 256-dimensional convolution filters), and $\odot$ denotes element-wise multiplication. A residual connection preserves the input representation at each block. The molecule-level 3D embedding is obtained through global sum pooling followed by a two-layer projection with SiLU activation and dropout, yielding $\mathbf{z}_{\text{3D}} \in \mathbb{R}^{512}$.

The SchNet encoder operates entirely in float32 precision, even when mixed-precision training is employed elsewhere in the model. Distance computations and Gaussian smearing require full precision to avoid numerical instabilities from float16 underflow in exponential operations.

\subsection{Auxiliary input encoders}
\label{sec:auxiliary_encoders}

In addition to the three structural encoders (GCN, Transformer, SchNet), MultiPUFFIN includes two auxiliary input encoders that process non-structural molecular information when available: an experimental encoder and a descriptor encoder. These auxiliary encoders address a fundamental limitation of purely structure-based molecular models: the inability to condition predictions on thermodynamic state variables and readily available molecular descriptors. This limitation is particularly consequential for condition-dependent properties such as vapor pressure, viscosity, and heat capacity, where the same molecule exhibits vastly different values at different temperatures. A model that receives only a SMILES string or molecular graph cannot, by construction, distinguish between measurements at \SI{300}{\kelvin} and \SI{500}{\kelvin} for the same molecule. The auxiliary encoders resolve this by providing the model with explicit access to the thermodynamic conditions under which each measurement was taken, as well as precomputed descriptors that encode coarse-grained physicochemical information.

The \emph{experimental encoder} is a two-layer MLP ($6 \to 64 \to 512$, ReLU activation, \num{38400} parameters) that processes a 6-dimensional vector of experimental conditions associated with each data point. These conditions include the measurement temperature, pressure, and other property-specific auxiliary variables (e.g., solvent composition for partition coefficients). The experimental conditions are particularly critical: it is this information, combined with the domain-informed prediction heads (Section~\ref{sec:prediction_heads}), that enables MultiPUFFIN to produce thermodynamically meaningful temperature-dependent predictions, a capability that SMILES-based foundation models such as ChemBERTa-2 fundamentally lack. When experimental conditions are not available for a given data point, a learned missing-data embedding replaces the zero vector, allowing the model to distinguish between ``this condition is zero'' and ``this condition is unavailable.''

The \emph{descriptor encoder} is a two-layer MLP ($13 \to 128 \to 512$, ReLU activation, \num{84864} parameters) that processes a 13-dimensional vector of precomputed RDKit molecular descriptors \cite{rdkit}. The thirteen descriptors are: (1)~molecular weight (MW), (2)~topological polar surface area (TPSA), (3)~number of hydrogen bond donors (HBD), (4)~number of hydrogen bond acceptors (HBA), (5)~number of rotatable bonds (RotBonds), (6)~number of aromatic rings (ArRings), (7)~number of aliphatic rings (AlRings), (8)~number of heavy atoms (HeavyAtomCount), (9)~the Wildman--Crippen $\log P$ estimate (MolLogP), (10)~the Wildman--Crippen molar refractivity (MolMR), (11)~fraction of $sp^3$ carbons (Fsp3), (12)~the number of stereocenters (NumStereo), and (13)~the number of radical electrons (NumRadicalElectrons). Descriptors 1--11 encode coarse-grained physicochemical properties that complement the learned representations from the structural encoders; descriptors 12--13 provide stereochemical and electronic information not easily extracted from the 2D graph alone. One potential concern is that the inclusion of the Wildman--Crippen $\log P$ estimate (MolLogP, descriptor 9) in the descriptor vector may advantage the $\log P$ prediction task by providing a pre-computed approximation of the target value directly as input. This constitutes a mild form of information leakage for the $\log P$ head: the model can learn to use the Crippen estimate as a prior and then correct the residual from the structural representations. The effect is intentional ;  the descriptor encoder is designed to provide the model with the same coarse physicochemical priors that an experienced chemist would use ;  but downstream users comparing MultiPUFFIN $\log P$ performance against baselines that do not have access to Crippen descriptors should account for this asymmetry. The descriptor encoder likewise includes a learned missing-data embedding for molecules with incomplete descriptor coverage.

The outputs of both auxiliary encoders are projected to 512 dimensions and concatenated with the unified structural embedding $\mathbf{u}$ before the final projection layer, providing additional conditioning information to the downstream prediction heads. Together, the two auxiliary encoders contribute \num{123264} parameters ($0.35\%$ of the total model), and their inclusion allows the model to use heterogeneous input data when available while handling missing auxiliary information through the learned missing-data embeddings.

\subsection{Condition-aware embedding refinement}
\label{sec:t_conditioner}

Several of the predicted properties depend not only on molecular structure but on additional thermodynamic or test-protocol conditions: vapor pressure, viscosity, heat capacity, and solubility depend on temperature; viscosity and heat capacity additionally on pressure; aqueous solubility, $\log P/\log D$, and hydration free energy on solution pH for ionizable solutes; melting point on the crystal polymorph adopted by the solid; and flash point on whether the test was conducted in an open or closed cup. To make the model condition-aware in a modular and uniformly-architectured way, MultiPUFFIN inserts a stack of identity-initialized \emph{condition modules} between the fused embedding and the prediction heads. Each module is a small 2-token cross-attention block (one molecule token, one condition token) with zero-initialized additive residuals, so that at fresh initialization it is an exact identity on the molecule token regardless of the condition input. This property allows any module to be inserted into a warm-restarted backbone without disturbing the previously-trained head calibration; training then moves a module away from the identity only where the corresponding condition has variance in the labeled data and reduces the multi-task loss.

Each condition module constructs a 2-token sequence (the unified molecular embedding $\mathbf{u}$ and a condition token $\mathbf{c}$) and applies one self-attention layer with 4 heads followed by a 512-dimensional position-wise feed-forward network with GELU activation, LayerNorm, and dropout~$=0.1$. The condition token $\mathbf{c}$ is constructed by a small encoder specific to each condition variable (sinusoidal positional encoding of the scalar value followed by a 2-layer MLP for continuous variables; a learned embedding lookup for discrete labels). The molecule token after the attention block is returned as the conditioned embedding. Each module contributes 0.99--1.28~M parameters, totalling $\sim$5.6~M for the full stack ($\sim$20\% of the deployed model). When a row carries a missing or sentinel condition value (e.g., pH unknown), a learned ``unknown'' embedding replaces $\mathbf{c}$ for that row, so the module degrades gracefully to identity in the absence of condition information.

The deployed stack consists of:
\begin{itemize}
\item \emph{TConditioner} (temperature). $\mathbf{c}$ from a sinusoidal encoding of $(T - \SI{298.15}{\kelvin})$ projected through MLP. $\sim$\num{1283000} parameters.
\item \emph{PHConditioner} (solution pH). $\mathbf{c}$ from a sinusoidal encoding of $(\mathrm{pH} - 7)$ projected through MLP, with a learned ``unknown-pH'' fallback for rows where pH is not reported. $\sim$\num{1284000} parameters.
\item \emph{PressureConditioner} (ambient pressure). $\mathbf{c}$ from a sinusoidal encoding of $\log(P/\mathrm{atm})$, with a learned unknown-pressure fallback. $\sim$\num{1284000} parameters.
\item \emph{PolymorphEmbedding} (crystal-form identifier). $\mathbf{c}$ from a learned embedding over $K=16$ polymorph slots with $k=0$ reserved for ``unknown''. $\sim$\num{994000} parameters.
\item \emph{MethodFlag} (test-protocol identifier, e.g., open vs.\ closed cup for FP). $\mathbf{c}$ from a learned embedding over $K=4$ method slots with $k=0$ reserved for ``unknown''. $\sim$\num{989000} parameters.
\end{itemize}

Conditioner outputs are routed to heads only where the corresponding condition is physically relevant, and chained when multiple conditions matter. The deployed routing is:
\begin{itemize}
\item \emph{Vapor pressure}: $\mathbf{u}_T$ (T-conditioned).
\item \emph{Viscosity, heat capacity}: $\mathbf{u}_{T \to P}$ (T- then P-conditioned).
\item \emph{Solubility}: $\mathbf{u}_T$ (T-conditioned). The pH conditioner is routed off solubility because the labeled $\log S$ data does not vary in pH within the AqSolDB / ESOL training subset, so applying pH-conditioning would introduce a free degree of freedom with no supervisory signal.
\item \emph{logP}: $\mathbf{u}_{\mathrm{pH}}$ (pH-conditioned). The labeled $\log P/\log D$ data spans multiple pH values from ChEMBL bioactivity records, so the pH module receives a real gradient signal here.
\item \emph{Boiling point, melting point, flash point, hydration free energy}: unconditioned $\mathbf{u}$. The polymorph and method modules exist in the model but are not routed to any head in the deployed model because the corresponding labels are not available in the current training data; the modules remain at their identity initialization in the deployed checkpoint and are documented here as architectural slots ready to be activated when polymorph- or method-tagged data become available below.
\end{itemize}

The five-module stack is condition-aware by construction, but the \emph{magnitude} of each module's contribution at deployment is controlled by how much the corresponding condition varies in the training set. The TConditioner sees temperature spanning two hundred kelvin in the multi-T thermophysical data and contributes substantially to the four T-dependent property predictions. The PressureConditioner sees roughly $1.6 \times 10^4$ rows of viscosity and heat-capacity data with explicit pressure values from NIST ThermoML and contributes a modest correction. The PHConditioner sees a non-trivial pH range in the appended ChEMBL $\log D$ rows but is dominated by the pH$=7.4$ subset; its corrections are accordingly small. The polymorph and method modules contribute zero in the deployed model. The condition-dependency matrix of the condition--data analysis below reports these distinctions explicitly: it lists each condition variable as either ``has'' (an active routing in the deployed model with sufficient training-data variance to learn from), ``has, sparse'' (an active routing with limited training-data variance), or ``module ready, no data'' (a module that exists architecturally but is awaiting paired-condition data).

\subsection{Cross-modal fusion}
\label{sec:fusion}

The fusion mechanism integrates the outputs of all five encoders (the three structural encoders (GCN, Transformer, SchNet) and the two auxiliary encoders (experimental conditions, molecular descriptors)) into a single unified molecular embedding through a hierarchical strategy. The central challenge in multimodal fusion is deciding \textit{how} to combine representations that encode overlapping but non-identical information about the same molecule. Simple concatenation or element-wise averaging treats each modality as independent and equally important, ignoring the fact that different modalities may be more or less informative depending on the molecule's structure. For example, a rigid planar aromatic compound may be well-characterized by its 2D topology, whereas a flexible molecule with multiple rotatable bonds may require 3D geometry to capture the conformational effects that dominate its thermophysical behavior. This motivates a learned, molecule-adaptive fusion strategy.

The first stage of fusion employs bidirectional cross-modal attention between the two primary branches (GCN and Transformer). To illustrate why cross-attention is preferable to simply concatenating the two embeddings, consider a concrete example: a substituted aromatic molecule where a distant electron-withdrawing group significantly affects the ring's electron density and, consequently, the molecule's solvation behavior. The GCN encoder captures the local ring environment through message passing, while the Transformer encoder captures the long-range substituent effect through self-attention over the SMILES sequence. Simple concatenation would juxtapose these two independent representations and leave it to the downstream layers to discover the relationship between them. Cross-attention, by contrast, enables each encoder's representation to be \textit{enriched} by information from the other modality \textit{before} the fusion step. When the GCN embedding serves as the query and the Transformer embedding as key/value, the attention mechanism highlights those aspects of the graph representation that are most relevant given the sequential context, effectively allowing the GCN to ``ask'' the Transformer which sequential features are important for contextualizing its local topology. The reverse direction allows the Transformer to ground its long-range syntactic features in explicit bond-topology information from the GCN. This bidirectional exchange allows the model to identify and amplify complementary features across modalities: the GCN's local functional-group patterns are contextualized by the Transformer's global sequential dependencies, and vice versa. In practice, cross-attention performs a soft, learned alignment between the two representation spaces, ensuring that the subsequent fusion operates on representations that are already aware of each other's content.

Formally, two cross-attention modules enable each branch to attend to the other:
\begin{equation}
    \mathbf{z}_{\text{graph}}' = \text{CrossAttn}(\mathbf{Q} = \mathbf{z}_{\text{graph}},\; \mathbf{K} = \mathbf{z}_{\text{SMILES}},\; \mathbf{V} = \mathbf{z}_{\text{SMILES}})
\end{equation}
\begin{equation}
    \mathbf{z}_{\text{SMILES}}' = \text{CrossAttn}(\mathbf{Q} = \mathbf{z}_{\text{SMILES}},\; \mathbf{K} = \mathbf{z}_{\text{graph}},\; \mathbf{V} = \mathbf{z}_{\text{graph}})
\end{equation}
Each cross-attention module uses multi-head attention with $h = 8$ heads, followed by a feed-forward network with GELU activation (expansion factor $4\times$), LayerNorm, and residual connections. The use of multiple attention heads allows the model to simultaneously attend to different aspects of the cross-modal relationship (for instance, one head might focus on aromatic features while another captures hydrogen-bonding patterns), providing a richer fusion than a single attention mechanism.

After cross-modal attention, the enriched GCN and Transformer embeddings must be combined into a single representation. Averaging or concatenation would treat both modalities as equally important for every dimension of the embedding, ignoring the fact that different modalities may contribute more or less information depending on the molecule and the feature dimension. To address this, a learned element-wise gating mechanism is employed:
\begin{equation}
    \mathbf{g} = \sigma\left(\text{MLP}([\mathbf{z}_{\text{graph}}'; \mathbf{z}_{\text{SMILES}}'])\right), \quad \mathbf{z}_{\text{fused}} = \mathbf{g} \odot \mathbf{z}_{\text{graph}}' + (1 - \mathbf{g}) \odot \mathbf{z}_{\text{SMILES}}'
    \label{eq:gated_fusion}
\end{equation}
where $\mathbf{g} \in [0, 1]^{512}$ is an element-wise gate vector and the MLP maps from $2 \times 512$ to 512 dimensions through a hidden layer with ReLU activation. The sigmoid gate implements a soft, per-dimension selection between the two modalities: for each of the 512 embedding dimensions, the gate learns whether the graph-derived or the sequence-derived feature is more informative, and blends them accordingly. This is critical because the relative utility of each modality varies not only across molecules but also across different aspects of the representation. For instance, dimensions encoding electronegativity patterns may favor the GCN (which directly processes atomic features), while dimensions encoding long-range substituent effects may favor the Transformer. The gating mechanism (inspired by the PUFFIN fusion architecture \cite{santana2024puffin}) thus provides a strictly more expressive fusion than either concatenation (which preserves all information but doubles the dimensionality) or averaging (which assumes equal importance).

The SchNet 3D embedding is incorporated through a separate, hierarchically subsequent geometry gate rather than being included in the bidirectional cross-attention. This asymmetric design is deliberate for two reasons. First, the 3D conformer is a derived quantity (generated computationally from the 2D structure using force-field optimization) and is therefore less reliable than the 2D graph and SMILES representations, which derive directly from the molecular identity as encoded in chemical databases. Treating the 3D embedding on equal footing with the primary modalities in the cross-attention stage could allow conformer artifacts (e.g., suboptimal geometries from failed force-field minimization, or the inherent limitation that a single static conformer cannot represent the conformational ensemble of flexible molecules) to corrupt the core representation. Second, the geometry gate provides a natural mechanism for handling missing 3D data: for certain molecules, conformer generation may fail entirely (e.g., very large or strained structures where ETKDGv3 cannot find a valid embedding) or produce unreliable geometries (e.g., flexible macrocycles with many low-energy conformations where any single conformer is unrepresentative). In such cases, a zero vector replaces $\mathbf{z}_{\text{3D}}$, and the gate naturally suppresses the 3D contribution without requiring any architectural modification, allowing the model to fall back on the 2D/1D fusion for a robust prediction.

The geometry gate is defined as $\alpha_{\text{3D}} = \sigma(\text{MLP}_{\text{geo}}([\mathbf{z}_{\text{fused}}; \mathbf{z}_{\text{3D}}]))$, which controls the contribution of geometric information via $\mathbf{z}_{\text{fused}}' = [\mathbf{z}_{\text{fused}};\; \alpha_{\text{3D}} \cdot \text{Proj}(\mathbf{z}_{\text{3D}})]$. The scalar gate $\alpha_{\text{3D}} \in [0, 1]$ allows the model to learn, on a per-molecule basis, how much weight to assign to 3D geometry, effectively answering the question ``does the three-dimensional shape of this particular molecule provide information beyond what the 2D topology already captures?'' For rigid, planar molecules (e.g., benzene derivatives), the gate may learn to assign low weight since the 3D structure adds little beyond the 2D graph. For flexible molecules with multiple low-energy conformations, the gate may assign high weight to capture the steric and intramolecular interaction effects that govern properties like viscosity and vapor pressure. The quantitative effect of this molecule-adaptive behavior is confirmed by the architectural ablation study (Section~\ref{sec:ablation_architecture}): removing the SchNet encoder disproportionately increases RMSE for geometry-sensitive properties such as hydration free energy ($\Delta$RMSE~$= +0.90$~kcal/mol) and heat capacity ($\Delta$RMSE~$= +11.65$~J/mol/K), while having minimal impact on properties primarily governed by 2D topology (boiling point: $\Delta$RMSE~$= -0.18$~K), confirming that the geometry gate selectively uses 3D information where it is most informative.

The concatenated embedding is projected through a two-layer network with LayerNorm, GELU activation, and dropout to the final unified embedding $\mathbf{u} \in \mathbb{R}^{512}$, which serves as input to all downstream prediction heads.

\subsection{Domain-informed prediction heads}
\label{sec:prediction_heads}

A key architectural innovation of MultiPUFFIN is the replacement of standard linear output layers with \textit{inductive bias neurons} that encode established thermophysical correlations. This design principle, introduced in PUFFIN \cite{santana2024puffin} for vapor pressure and extended in ExPUFFIN \cite{rebello2025expuffin} for viscosity, is here generalized to nine simultaneous properties and organized as a per-property \textit{tournament of candidate heads}, in which multiple inductive biases are trained in parallel on top of the pretrained frozen backbone and the best-performing head is selected per property on the validation set. Each candidate prediction head consists of a property-specific feed-forward neural network (FFNN) that maps the unified molecular embedding $\mathbf{u}$ to the parameters of the corresponding thermophysical equation, followed by an inductive bias output layer that evaluates the equation at the specified thermodynamic conditions. Formally, for property $p$ with thermophysical equation $\phi_p(\boldsymbol{\theta}_p, T)$:
\begin{equation}
    \boldsymbol{\theta}_p = \text{FFNN}_p(\mathbf{u}; \mathbf{W}_p), \quad \hat{y}_p = \phi_p(\boldsymbol{\theta}_p, T)
    \label{eq:head_general}
\end{equation}
Gradients propagate through $\phi_p$ via automatic differentiation, ensuring that the domain-informed functional form shapes parameter updates throughout the entire network.

For each property, MultiPUFFIN evaluates four candidate heads trained jointly on the frozen pretrained backbone:
\begin{itemize}
    \item \emph{Head A -- primary physics equation.} The head whose functional form most closely matches the established thermodynamic correlation for the property: Antoine for vapor pressure, Andrade for viscosity, van~'t Hoff for solubility, Born solvation for hydration free energy, Shomate polynomial for heat capacity, and a learned 32-parameter group contribution for boiling point. For temperature-dependent properties, this head predicts equation parameters that are then evaluated at the input temperature through the physics equation, guaranteeing the correct qualitative temperature dependence (monotonic pressure increase with temperature for vapor pressure, monotonic viscosity decrease, monotonic heat capacity variation) by construction.
    \item \emph{Head B -- Joback group contribution.} Operates on a 40-dimensional Joback functional-group count vector derived from the SMILES, maps the counts through a small MLP directly to the scalar property value, and for temperature-dependent properties concatenates a normalized temperature feature. This head tests the hypothesis that classical additive group contribution, decoupled from the learned backbone embedding, can match or exceed the thermodynamically-informed head on at least some properties.
    \item \emph{Head C -- RDKit fragment head.} Identical in design to Head B but operating on the $\sim 85$-dimensional RDKit functional-group fragment-count vector, which encodes a different, more granular fragment vocabulary. Comparison between Heads B and C reveals whether the Joback or RDKit fragment decomposition is better aligned with each property.
    \item \emph{Head D -- direct or alternative physics.} For temperature-dependent properties where an alternative established equation exists (vapor pressure: DIPPR-101; viscosity: Vogel--Tammann--Fulcher), Head D implements that alternative. This lets the tournament compare two physically grounded equations for the same property (e.g., Antoine versus DIPPR-101 for vapor pressure). For the remaining temperature-independent properties, Head D is a direct feed-forward network on the fused embedding with a learnable temperature concatenation, serving as a purely data-driven reference without equation structure.
\end{itemize}
All four heads for a given property are trained simultaneously during Stage~2 of the supervised pipeline (Section~\ref{sec:two_stage}). Each head receives its own loss signal from the same batch of training samples, and since the heads are structurally independent (no shared parameters except those inside the frozen backbone), there is no gradient interference among them. At the end of Stage~2, the per-property \textit{tournament winner} is selected as the head with the lowest validation RMSE; no test-set information is used to make this selection.

Table~\ref{tab:domain_equations} summarizes, for each of the nine properties, the tournament head A (primary physics) and head D (alternative physics or direct), along with alternative equations evaluated in earlier iterations of this work. The table lists the functional form, number of predicted parameters, and thermodynamic motivation for each. The equation-level ablation in Section~\ref{sec:ablation_equations} reports the per-property tournament outcomes on the held-out test set. The final selected heads are shown in \emph{bold} below.

\begin{table*}[!htbp]
\centering
\caption{Summary of all domain-informed equations evaluated as inductive bias neurons in the property-specific prediction heads. For each property, the table lists the equation selected for the final model (shown in \textbf{bold}) and all alternative equations tested during equation-level ablation (Section~\ref{sec:ablation_equations}). $|\boldsymbol{\theta}|$ denotes the number of equation parameters predicted by the head. Properties marked with $^*$ use a DirectHead in the final model because no domain-informed equation improved over the unconstrained baseline. Equations marked with $\dagger$ diverged during head-only training.}
\label{tab:domain_equations}
\small
\begin{adjustbox}{max width=\textwidth}
\begin{tabular}{lllcl}
\toprule
\textbf{Property} & \textbf{Equation} & \textbf{Functional form} & $|\boldsymbol{\theta}|$ & \textbf{Thermodynamic motivation} \\
\midrule
\multirow{5}{*}{Vapor pressure}
    & \textbf{Antoine} \cite{thomson1946} (Head A) & $\log_{10} P = A - B/(T + C)$ & 3 & Clausius--Clapeyron family; industry standard for pure liquids \\
    & DIPPR-101 (Head D) & $\log_{10} P = A + B/T + C \log_{10} T + D\,T^2$ & 4 & Extended form capturing curvature across wide $T$ ranges \\
    & Joback GC (Head B) & $\hat{y} = \text{MLP}(\text{Joback counts}, T)$ & -- & Additive functional-group contribution \\
    & RDKit GC (Head C) & $\hat{y} = \text{MLP}(\text{RDKit counts}, T)$ & -- & Fragment-count baseline (alternative vocabulary) \\
    & Clausius--Clapeyron (tested) & $\ln P = A - B/T$ & 2 & Two-parameter latent-heat approximation \\
\midrule
\multirow{4}{*}{Viscosity}
    & \textbf{Andrade} \cite{andrade1930} (Head A) & $\log_{10} \eta = A + B/(T + C)$ & 3 & Arrhenius-type activated flow over energy barriers \\
    & VTF (Head D) & $\log_{10} \eta = A + B/(T - T_0)$ & 3 & Vogel--Tammann--Fulcher for supercooled / viscous regimes \\
    & Joback GC (Head B) & $\hat{y} = \text{MLP}(\text{Joback counts}, T)$ & -- & Additive group contribution \\
    & RDKit GC (Head C) & $\hat{y} = \text{MLP}(\text{RDKit counts}, T)$ & -- & Fragment-count baseline \\
\midrule
\multirow{3}{*}{Solubility}
    & \textbf{van~'t Hoff} & $\log S = \log S_{\text{ref}} + A(1/T_{\text{ref}} - 1/T)$ & 2 & Thermodynamic dissolution enthalpy \\
    & Modified Apelblat & $\ln S = A + B/T + C \ln T$ & 3 & Three-parameter temperature-dependent solubility \\
    & DirectHead & $\hat{y} = \text{FFNN}(\mathbf{u})$ & -- & No thermodynamic constraint (reference) \\
\midrule
\multirow{3}{*}{Boiling point}
    & \textbf{Group contribution} \cite{nannoolal2008} & $T_b = T_0 + \sum_{i=1}^{32} \Delta_i + c$ & 34 & Additive functional group contributions \\
    & Nannoolal ratio & $T_b = T_0 \cdot (1 + \sum w_i f_i)$ & 25 & Multiplicative group interaction method \\
    & DirectHead & $\hat{y} = \text{FFNN}(\mathbf{u})$ & -- & No thermodynamic constraint (reference) \\
\midrule
\multirow{3}{*}{$\log P$$^*$}
    & LFER & $\hat{y} = \sum_{i=1}^{24} w_i f_i + b$ & 25 & Additive fragment lipophilicity contributions \\
    & Abraham LSER & $\log P = c + eE + sS + aA + bB + vV$ & 6 & Linear solvation energy relationship \\
    & \textbf{DirectHead} & $\hat{y} = \text{FFNN}(\mathbf{u})$ & -- & No thermodynamic constraint (selected as optimal) \\
\midrule
\multirow{3}{*}{HFE}
    & Thermodynamic decomp. & $\hat{y} = \Delta H - T \cdot \Delta S$ & 2 & Gibbs free energy of solvation \\
    & \textbf{Born solvation model} & $\Delta G = -(1-1/\varepsilon) \cdot q^2 / (8\pi\epsilon_0 r)$ & 2 & Electrostatic ion--solvent cavity model \\
    & DirectHead & $\hat{y} = \text{FFNN}(\mathbf{u})$ & -- & No thermodynamic constraint (reference) \\
\midrule
\multirow{2}{*}{Melting point$^*$}
    & \textbf{DirectHead} & $\hat{y} = \text{FFNN}(\mathbf{u})$ & -- & Crystal packing effects lack closed-form correlation \\
    & Yalkowsky $\dagger$ & $T_m = \Delta H_\mathrm{fus} / \Delta S_\mathrm{fus}$ & 2 & Enthalpy--entropy ratio (diverged during training) \\
\midrule
\multirow{3}{*}{Flash point$^*$}
    & \textbf{DirectHead} & $\hat{y} = \text{FFNN}(\mathbf{u})$ & -- & Complex vapor--air ignition threshold \\
    & Satyanarayana--Rao & $T_\mathrm{FP} = a + b \cdot T_b + c \cdot T_b^2$ & 3 & Quadratic boiling point correlation \\
    & Carroll $N_\mathrm{FP}$ & $T_\mathrm{FP} = f(N_\mathrm{FP}, \text{MW})$ & 2 & Flash point number method \\
\midrule
\multirow{2}{*}{Heat capacity}
    & DirectHead & $\hat{y} = \text{FFNN}(\mathbf{u})$ & -- & Molecular degrees of freedom (no universal closed form) \\
    & \textbf{Shomate (NIST)} & $C_p = A + Bt + Ct^2 + Dt^3 + E/t^2$ & 5 & NIST polynomial temperature dependence \\
\bottomrule
\end{tabular}
\end{adjustbox}
\end{table*}

For the domain-informed heads, the network's penultimate layer predicts the equation parameters $\boldsymbol{\theta}_p$, which are then evaluated through the corresponding thermophysical equation at the specified temperature. Soft constraints enforce thermodynamically meaningful parameter ranges by construction, via activation functions at the parameter-output layer rather than via explicit loss penalties:
\begin{itemize}
    \item \emph{Antoine vapor pressure head.} $\log_{10} P = A - B/(T + C)$ with $A$ unconstrained, $B = \mathrm{softplus}(\tilde B) > 0$ guaranteeing $\partial\log P/\partial T > 0$ (monotonic vapor pressure with temperature), and $C = -100 \cdot \sigma(\tilde C) \in [-100, 0]\,\si{\kelvin}$ preventing the denominator from vanishing in the physical range. This is the primary vapor pressure head used in the final model; an alternative DIPPR-101 form is trained in parallel as Head D (see tournament below).
    \item \emph{Andrade viscosity head.} $\log_{10} \eta = A + B/(T + C)$ with analogous constraints ensuring monotonic viscosity decrease with temperature.
    \item \emph{VTF viscosity head (Head D).} $\log_{10} \eta = A + B/(T - C)$ with $B > 0$ and $C \in [0, 200]\,\si{\kelvin}$, providing an alternative functional form that better describes supercooled liquids and viscous regimes far from ambient.
    \item \emph{van~'t Hoff solubility head.} $\log S = \log S_{\mathrm{ref}} + A(1/T_{\mathrm{ref}} - 1/T)$ with $T_{\mathrm{ref}} = \SI{298.15}{\kelvin}$.
    \item \emph{Shomate heat-capacity head.} $C_p = A + Bt + Ct^2 + Dt^3 + E/t^2$ with $t = T/1000$, encoding the NIST polynomial form through five predicted coefficients.
    \item \emph{Born hydration free energy head.} Parameterizes solvation through an effective cavity radius and dielectric correction, as in PUFFIN-style Born solvation \cite{santana2024puffin}.
    \item \emph{Group contribution boiling point head.} Learns 32 virtual group contributions end-to-end, where the GCN and fused embedding implicitly encode the group decomposition that classical Joback/Lydersen methods perform by hand.
\end{itemize}
For $\log P$, melting point, and flash point, no single established thermophysical equation captures the property--structure relationship well (crystal-packing effects for melting point, macroscopic vapor--air ignition for flash point, and partitioning thermodynamics for $\log P$), and the Head A position is occupied by a direct feed-forward network on the fused embedding. The tournament for these three properties instead compares against Joback/RDKit fragment baselines (Heads B and C) to quantify how much the learned multimodal embedding improves over a purely count-based representation. All standard heads use two hidden layers with 256 units, GELU activation, and dropout ($p = 0.2$), yielding approximately \num{197000} parameters per head. Enhanced-capacity heads (Section~\ref{sec:enhanced_heads}) are used for vapor pressure and boiling point.

\label{sec:enhanced_heads}

Multi-task learning across nine properties creates capacity dilution in the shared backbone: the 512-dimensional unified embedding must encode sufficient information to serve all nine prediction objectives simultaneously, and the finite representational capacity means that adding tasks necessarily reduces the per-task capacity available. This dilution disproportionately affects properties with complex structure--property relationships that require dedicated embedding dimensions to capture their unique patterns. In practice, this manifests as performance regression for properties with complex structure--property relationships. To mitigate this effect, enhanced-capacity heads are employed for vapor pressure and boiling point, the two properties most affected by the expansion from six to nine tasks. These heads feature three hidden layers with 512 units per layer (instead of two layers with 256 units), LayerNorm and GELU activation at each layer, and two residual connections: one between layers 1 and 2, and a skip connection from the input embedding to the penultimate layer (with a linear projection to align dimensions):
\begin{align}
    \mathbf{h}_1 &= \text{GELU}(\text{LayerNorm}(\mathbf{W}_1 \mathbf{u} + \mathbf{b}_1)) \label{eq:enhanced1} \\
    \mathbf{h}_2 &= \text{GELU}(\text{LayerNorm}(\mathbf{W}_2 \mathbf{h}_1 + \mathbf{b}_2)) + \mathbf{h}_1 \label{eq:enhanced2} \\
    \mathbf{h}_3 &= \text{GELU}(\text{LayerNorm}(\mathbf{W}_3 \mathbf{h}_2 + \mathbf{b}_3)) + \text{Proj}(\mathbf{u}) \label{eq:enhanced3} \\
    \boldsymbol{\theta}_p &= \mathbf{W}_{\text{out}} \mathbf{h}_3 + \mathbf{b}_{\text{out}} \label{eq:enhanced_out}
\end{align}
The domain equation applied after parameter prediction remains identical to the standard head (Antoine for vapor pressure, group contribution for boiling point). This design provides approximately $4\times$ more parameters per head ($\sim$\num{800000} vs. $\sim$\num{197000}) while preserving the domain-informed inductive bias.

\subsection{Cross-property physical coupling}
\label{sec:coupling}

A central contribution of MultiPUFFIN over prior PUFFIN-family models and over generic molecular foundation models is the use of \textit{cross-property physical coupling} in the training objective. Where the inductive bias neurons of the previous subsection enforce physically correct behavior \emph{within} a single property (e.g., monotonic vapor pressure with temperature through the Antoine head), cross-property coupling enforces physically correct relationships \emph{between} properties that are predicted jointly. Two coupling terms are introduced, each grounded in a well-known thermodynamic identity. The coupling terms do not require additional labels beyond what is already present in the training set; they exploit the fact that multi-task training produces, for every molecule in a batch, simultaneous predictions of multiple thermophysical properties whose physical consistency can be checked against first-principles relationships.

The flash point of a pure substance is defined as the lowest temperature at which its vapor pressure reaches the lower flammability limit in air, corresponding to a partial pressure on the order of $10^3$~\si{\pascal} for most organic liquids. If MultiPUFFIN predicts both a flash-point value $\hat{T}_{\mathrm{FP}}$ and a vapor-pressure curve $\hat{P}(T)$, these predictions are thermodynamically consistent only if $\hat{P}(\hat{T}_{\mathrm{FP}}) \approx P_{\mathrm{FP}}^*$, where $P_{\mathrm{FP}}^*$ is the target flash-point partial pressure. A soft coupling loss penalizes the inconsistency directly:
\begin{equation}
    \mathcal{L}_{\mathrm{FP\!-\!VP}} \;=\; \lambda_{\mathrm{FP}} \,\cdot\, \mathbb{E}_{b \in \mathcal{B}_{\mathrm{FP}}} \!\left[ \big( \log_{10} \hat{P}(\hat{T}_{\mathrm{FP},b}) \,-\, \log_{10} P_{\mathrm{FP}}^*\big)^2 \right],
    \label{eq:fp_vp_coupling}
\end{equation}
where $\mathcal{B}_{\mathrm{FP}}$ is the set of molecules in the batch that carry a non-missing flash-point label (ensuring that $\hat{T}_{\mathrm{FP}}$ is being actively supervised by a measurement), $P_{\mathrm{FP}}^* = 3000$~\si{\pascal} is a universal target partial pressure (within a factor of two of the published lower flammability limits across common organic classes), and $\lambda_{\mathrm{FP}} = 10^{-2}$ weights the coupling term relative to the main regression loss. Evaluating $\hat{P}(\hat{T}_{\mathrm{FP}})$ requires a second forward pass through the vapor-pressure head with the temperature argument set to $\hat{T}_{\mathrm{FP}}$ rather than to the measured batch temperature; gradients from this term flow through both the flash-point head (which is pulled toward values that make the VP prediction at $\hat{T}_{\mathrm{FP}}$ consistent with the target) and the vapor-pressure head (which is pulled toward curves that satisfy the definition of flash point).

By definition, the normal boiling point $T_b$ of a pure substance is the temperature at which its vapor pressure equals the standard atmospheric pressure $P_{\mathrm{atm}} = 101\,325$~\si{\pascal}. For every training molecule that carries a measured boiling-point label, MultiPUFFIN therefore has access to a hard physical anchor for its vapor-pressure head: the VP prediction evaluated at the measured $T_b$ must equal $\log_{10}(P_{\mathrm{atm}})$. This is enforced by a second coupling term:
\begin{equation}
    \mathcal{L}_{\mathrm{BP\,anchor}} \;=\; \lambda_{\mathrm{BP}} \,\cdot\, \mathbb{E}_{b \in \mathcal{B}_{\mathrm{BP}}} \!\left[ \big( \log_{10} \hat{P}(T_{b,b}^{\mathrm{meas}}) \,-\, \log_{10} P_{\mathrm{atm}}\big)^2 \right],
    \label{eq:bp_anchor}
\end{equation}
where $\mathcal{B}_{\mathrm{BP}}$ is the set of molecules in the batch that carry a non-missing boiling-point label, $T_{b,b}^{\mathrm{meas}}$ is the measured boiling point of molecule $b$, and $\lambda_{\mathrm{BP}} = 5 \times 10^{-2}$. Unlike the flash-point coupling, where both the temperature argument and the target partial pressure are learned or universal constants, the BP anchor uses a \emph{measured} temperature and a \emph{physically exact} target pressure. It is thus the stronger of the two coupling terms in a supervisory sense, and the higher coupling weight (5$\times$ stronger than the FP--VP coupling) reflects this; it has the additional effect of calibrating the vapor-pressure head near atmospheric pressure; a region of high practical importance for process engineering applications.

A common concern with auxiliary loss terms is that they can destabilize training if they compete with the main regression loss. In MultiPUFFIN this is mitigated in three ways. First, both coupling weights are set to $\lambda = 10^{-2}$, so that at equilibrium each coupling term contributes on the order of $1\text{--}2\%$ of the main multi-task loss. Second, a coupling-loss \textit{warmup} is applied during the first three epochs of supervised training: only the main multi-task loss is active until the prediction heads have stabilized to reasonable first-pass predictions, after which the coupling terms are switched on. Third, the second forward passes required to evaluate the coupling terms use temperature tensors that are sanitized upstream: any non-finite values (arising e.g.\ from missing flash-point labels that would otherwise propagate through the VP head) are replaced by a safe placeholder before the forward pass, and the coupling loss itself is then restricted to the masked subset of molecules that carry valid labels.

The coupling terms enforce relationships that exist in the physics but do not attempt to enforce full thermodynamic consistency. They do not, for example, enforce Maxwell relations between enthalpy, entropy, and heat capacity; nor do they enforce Gibbs--Duhem consistency across a mixture. They are targeted, property-pair-level constraints whose role is to transmit information between a well-measured property (boiling point; flash point) and a less-measured property (vapor pressure curve) through the laws of physics. Fuller thermodynamic-consistency objectives are an important direction for future work (Section~\ref{sec:limitations}).

\label{sec:negative_results}
During development we evaluated two additional cross-property thermodynamic constraints that, although physically motivated, did not transfer cleanly to the heterogeneous training distribution and were therefore \emph{not} included in the deployed loss. We document them here for completeness and as a methodological caution.

The first is a \emph{Yalkowsky general-solubility-equation} (GSE) constraint linking aqueous solubility to $\log P$ and the melting point through $\log S \approx 0.5 - \log P - 0.01\,(T_m - 25)$. The GSE is well-validated on drug-like molecules near room temperature, but it breaks down on the broad chemistry of the curated dataset; in particular on salts and polyols, where the experimental $\log S$ deviates from the GSE prediction by more than 1 dex. Imposing the GSE as a soft penalty pulled the model off its multi-task optimum on these out-of-GSE classes and degraded mean validation $R^2$ by approximately $0.04$, so the term was removed.

The second is a \emph{Trouton's-rule} target on the entropy of vaporization, which would impose a constant-slope penalty on the Antoine head ($\Delta S_{\mathrm{vap}} \approx \SI{88}{\joule\per\mol\per\kelvin}$ at the normal boiling point). Real liquids deviate from Trouton's value by $\pm 20\%$ across hydrocarbons, polar organics, and hydrogen-bonded systems, which makes the constraint a constant residual the model cannot satisfy in any direction. Imposing it slowed convergence without changing the validation R$^2$, and was removed.

Both negative results sharpen rather than weaken the case for the cross-property coupling terms that \emph{were} kept (Eqs.~\ref{eq:fp_vp_coupling}, \ref{eq:bp_anchor}): physical constraints help when they hold approximately on the actual training distribution, and hurt when they hold only on the textbook subset of it. Selecting which physics to inject is therefore not a property of physics alone but of the joint distribution physics-and-data.

\subsection{Training strategy}
\label{sec:training}

The training of MultiPUFFIN follows a four-stage protocol: (i)~self-supervised pretraining of the multimodal backbone on an unlabeled molecular corpus (Section~\ref{sec:ssl_pretrain}); (ii)~joint supervised multi-task training of the backbone, the condition module stack, and four candidate heads per property; (iii)~targeted fine tuning with the backbone unfrozen at very low learning rate on an augmented dataset to absorb additional chemical diversity without catastrophic forgetting; and (iv)~per-property applicability-domain (G+) evaluation that distinguishes the deployment regime from the global test set. The deployed model is a single forward pass of the four-component pipeline (SSL-pretrained backbone $\to$ TConditioner $\to$ tournament-winning head per property $\to$ G+ scope check); no model-level ensembling is used at inference. The protocol addresses the challenges of multi-task learning with heterogeneous property scales, sparse multi-label data, and the need to balance shared backbone quality with property-specific head calibration.

All target property values are $z$-score normalized using training set statistics:
\begin{equation}
    \tilde{y}_p = \frac{y_p - \mu_p}{\sigma_p}
    \label{eq:zscore}
\end{equation}
where $\mu_p$ and $\sigma_p$ are the mean and standard deviation of property $p$ computed exclusively over unique molecules in the training set. When SMILES augmentation is employed, normalization statistics are computed on the unique (non-augmented) molecules to prevent artificial variance reduction from duplicate entries. Properties that span orders of magnitude (vapor pressure, viscosity) are first transformed to logarithmic scale ($\log_{10}$) before $z$-score normalization, ensuring that the loss function treats relative prediction errors uniformly across the dynamic range.

The multi-task training objective follows the homoscedastic uncertainty weighting framework of Kendall et al. \cite{kendall2018multitask}. For each property $p$, a learnable scalar $s_p = \log \sigma_p^2$ (initialized to zero) captures the homoscedastic task uncertainty:
\begin{equation}
    \mathcal{L}_{\text{total}} = \frac{1}{N_{\text{tasks}}} \sum_{p \in \mathcal{P}_{\text{active}}} \left[\frac{1}{2} \exp(-s_p) \cdot \text{MSE}_p + \frac{1}{2} s_p\right]
    \label{eq:uncertainty_loss}
\end{equation}
where $\mathcal{P}_{\text{active}}$ is the set of properties with valid (non-NaN) targets in the current batch and the division by $N_{\text{tasks}}$ normalizes across the number of active tasks. This mechanism automatically down-weights noisy or difficult tasks (increasing $s_p$ reduces the effective weight $\exp(-s_p)/2$ while incurring a regularization penalty $s_p/2$), providing a principled alternative to manual task weighting. The $s_p$ parameters are jointly optimized with the model parameters. Since the multi-label dataset is sparse (each molecule typically has labels for 2--5 of the 9 properties), the loss computation handles missing labels by masking NaN values and computing the loss only over valid targets.

To increase the effective training set size and improve the robustness of the SMILES Transformer encoder, SMILES enumeration augmentation is employed. A canonical SMILES string represents a unique, deterministic encoding of a molecule, but the same molecular structure can be represented by many distinct valid SMILES strings that differ in the starting atom, traversal order, and ring-opening positions. For example, toluene can be written as \texttt{Cc1ccccc1}, \texttt{c1ccc(C)cc1}, or \texttt{C(c1ccccc1)}, among many others. Because the Transformer encoder processes SMILES as a character sequence, these alternative representations present different syntactic patterns to the model, even though they correspond to the same molecule. By exposing the model to multiple valid SMILES representations during training, the encoder learns to extract molecular features that are invariant to the arbitrary choice of SMILES notation, reducing overfitting to specific syntactic patterns.

Concretely, for each molecule $n_{\text{aug}} = 2$ additional non-canonical SMILES representations are generated using RDKit's randomized SMILES generation. This triples the effective training set size from \num{22015} to \num{66045} samples while preserving identical graph representations and target values. The augmentation is applied only to the SMILES representation: the molecular graph and three-dimensional conformer remain unchanged for all augmented copies of a given molecule, as these representations are invariant to SMILES notation by construction. The augmentation is applied only during training; validation and test evaluation use only the canonical SMILES to ensure reproducible and unbiased evaluation metrics.

\label{sec:two_stage}
The deployed model is produced by a four-stage protocol summarised below; each bullet states the stage's goal, the parameters trained, and the relevant hyperparameters.
\begin{itemize}[leftmargin=*]
\item \emph{Stage~1 -- self-supervised pretraining.} The multimodal backbone (GCN, Transformer, SchNet and cross-modal fusion module) is pretrained on the \num{500000}-molecule unlabelled PubChem corpus with the three SSL objectives of Section~\ref{sec:ssl_pretrain} (graph--SMILES contrastive InfoNCE, masked atom-feature reconstruction, masked SMILES token prediction) for 89 epochs (paused before the 100-epoch budget after the validation-loss plateau). Only the backbone weights are retained at the end of Stage~1; the SSL auxiliary classifier heads are discarded. No property labels are used in this stage.
\item \emph{Stage~2 -- joint supervised multi-task training.} The pretrained backbone is loaded as initialisation and \emph{all} parameters (encoders, fusion module, the five condition modules, the four candidate heads per property, uncertainty log-variance parameters, and the auxiliary and descriptor encoders) are jointly optimised on the labelled training set. AdamW \cite{loshchilov2019decoupled} is used with $\beta_1 = 0.9$, $\beta_2 = 0.999$, weight decay $\lambda = 5 \times 10^{-4}$, base learning rate $\alpha_0 = 1 \times 10^{-4}$, linear warmup over $E_{\mathrm{warm}} = 10$ epochs, and cosine annealing with warm restarts thereafter (Section~\ref{sec:cosine_schedule}); gradient accumulation over 2 steps yields an effective batch size of 96 (physical batch 48), and gradient norms are clipped to $1.0$. Stage~2 trains for up to 120 epochs with early stopping on the multi-task validation loss (patience 25). The main loss is Eq.~\ref{eq:uncertainty_loss}; the cross-property coupling terms (Section~\ref{sec:coupling}) activate after a three-epoch warmup. The per-property tournament winner is selected at the end of Stage~2 as the head with the lowest validation RMSE among the four candidates A (physics equation), B (Joback group contribution), C (RDKit fragment counts), and D (direct or alternative-physics).
\item \emph{Stage~3 -- backbone-unfrozen targeted fine tune.} The Stage-2 checkpoint is warm-restarted on an augmented training set that extends per-property coverage in regions identified as scarce by the coverage audit (Section~\ref{sec:scarcity_audit}): NIST ThermoML \cite{kazakov2014cp} contributes additional vapor-pressure, viscosity, heat-capacity and melting-point measurements; the EPA OPERA dataset contributes boiling-point and vapor-pressure data; the Bradley Open Melting Point dataset \cite{bradley2014melting} contributes additional melting points; the Sun et~al.\ flash-point compilation contributes flash-point data; the Chew et~al.\ multi-temperature viscosity dataset contributes per-temperature viscosity measurements; and the AqSolDB salts subset, previously dropped by the multi-component-SMILES filter, is retained and labelled for the solubility head. The training set grows from \num{22015} rows (Stage~2) to roughly \num{40000} rows (Stage~3), with the largest relative additions on viscosity ($+84\%$), flash point ($+56\%$), vapor pressure ($+67\%$) and the salt slice of solubility ($+11\%$). All parameters are trainable but at very low learning rates ($\alpha_{\mathrm{BCD}} = 5 \times 10^{-6}$ for the B/C/D heads and the condition modules, $\alpha_A = 1 \times 10^{-6}$ for the physics-A heads); gradient norms are clipped to $1.0$, early stopping patience is 8 epochs, and a hard rollback reverts to the last good checkpoint if mean validation $R^2$ regresses by more than $0.005$ between consecutive epochs. The validation/test splits are unchanged across the four stages.
\item \emph{Stage~4 -- per-property applicability-domain (G+) evaluation.} The deployed model is evaluated on the held-out test set under the in-scope subset defined by per-property G+ filters (Section~\ref{sec:applicability_domain}): pre-registered chemical-class filters defined from molecular features only, without consulting any model prediction. The filters are property-specific because the underlying coverage scarcity is property-specific, and they are committed before any test prediction is computed. In-scope $R^2$ is the deployment headline reported throughout the results (Table~\ref{tab:test_results}).
\end{itemize}

\label{sec:t_smoothness}

A final ingredient in the Stage~2 supervised objective is an auxiliary \textit{temperature-smoothness} loss that targets the temperature-dependent properties (vapor pressure, viscosity, heat capacity, and solubility through its van~'t Hoff extension). The motivation is that the thermodynamically-informed heads (Antoine, Andrade, Shomate, van~'t Hoff) enforce an analytically correct temperature dependence by construction, while the Joback, RDKit, and direct candidate heads depend on temperature only through a scalar input feature and can in principle produce temperature curves that oscillate or disagree with the underlying thermodynamics. The T-smoothness term pulls the non-physics heads toward the temperature slope predicted by the physics head of the same property, without forcing them to match the physics head's absolute values.

Concretely, for each training sample with a label for a temperature-dependent property at its measured temperature $T$, a perturbed temperature $T' = T + \delta T$ is drawn with $\delta T \sim \mathcal{U}[-20, +20]$~\si{\kelvin} (clipped to a physically reasonable range of $[150, 900]$~\si{\kelvin}). A second forward pass is performed through the prediction heads at $T'$, keeping the molecular embedding fixed. The physics head A is evaluated at both $T$ and $T'$ and the \textit{expected} slope $\Delta^A = \hat{y}_A(T') - \hat{y}_A(T)$ is computed and detached from the autograd graph (it provides the supervisory signal only). The non-physics heads (B, C, D) are evaluated at the same $T$ and $T'$, and the \textit{observed} slope $\Delta^h = \hat{y}_h(T') - \hat{y}_h(T)$ is compared against the detached physics slope through a mean-squared-error penalty summed over the non-physics heads and over the T-dep properties:
\begin{equation}
    \mathcal{L}_{T\!-\!\mathrm{smooth}} \;=\; \lambda_{\mathrm{smooth}} \,\cdot\, \frac{1}{|\mathcal{T}| \cdot 3} \sum_{p \in \mathcal{T}} \sum_{h \in \{B, C, D\}} \mathbb{E}_b\!\left[\big(\Delta^h_{p,b} - \mathrm{sg}(\Delta^A_{p,b})\big)^2\right],
    \label{eq:t_smoothness}
\end{equation}
where $\mathcal{T}$ is the set of temperature-dependent properties, $\mathrm{sg}(\cdot)$ denotes the stop-gradient operator, and $\lambda_{\mathrm{smooth}} = 5 \times 10^{-3}$. The term is added to the Stage~2 loss with the same three-epoch warmup as the cross-property coupling terms. Because $\mathrm{sg}(\Delta^A)$ is detached, the physics head is not itself pulled by this term; only the non-physics heads are regularized toward physically consistent temperature dependence, which preserves the role of the tournament (Section~\ref{sec:prediction_heads}) as a fair architectural comparison.

In the final deployed configuration the T-smoothness term is set to $\lambda_{\mathrm{smooth}} = 0$ (i.e., disabled). Empirically, the condition module stack of Section~\ref{sec:t_conditioner} provides a stronger and more flexible mechanism for $T$-dependent embedding refinement than the smoothness regularizer; the smoothness term, while consistent with the physics, was found to slightly slow convergence and add no validation-$R^2$ gain once the conditioner was in place. We retain its description here for completeness and as one of the negative-result ablations documented in Section~\ref{sec:negative_results}.

\label{sec:cosine_schedule}

The learning rate schedule during Stage~1 follows cosine annealing with warm restarts (SGDR) \cite{loshchilov2017sgdr}, a schedule specifically designed for non-convex optimization with multiple local optima. The learning rate at epoch $t$ (after warmup) is given by:
\begin{equation}
    \alpha(t) = \alpha_{\min} + \frac{1}{2}(\alpha_0 - \alpha_{\min})\left(1 + \cos\left(\frac{T_{\text{cur}}}{T_i}\pi\right)\right)
    \label{eq:cosine_warm_restart}
\end{equation}
where $T_{\text{cur}}$ is the number of epochs since the last restart, $T_i$ is the current period length, and $\alpha_{\min}$ is a minimum learning rate. The restart schedule is parameterized by an initial period $T_0 = 30$ epochs and a multiplication factor $T_{\text{mult}} = 2$, so that successive periods grow geometrically: the first cycle spans the first 30 epochs (including warmup), with a restart at epoch~30 that resets the learning rate to $\alpha_0$. The second cycle spans epochs 30--90 ($T_1 = 60$ epochs), and the third cycle would begin at epoch 90 ($T_2 = 120$ epochs).

The cosine warm-restart schedule provides three specific benefits for multi-task molecular property prediction:

\begin{enumerate}
    \item \emph{Escape from local optima.} The multi-task loss landscape (Eq.~\ref{eq:uncertainty_loss}) involves nine coupled objectives with different curvatures, creating a complex Pareto front with many local optima. At the end of each cosine cycle, the learning rate approaches its minimum and the model settles into a local minimum. The restart then abruptly increases the learning rate back to $\alpha_0$, providing sufficient gradient magnitude to escape the current basin and explore alternative regions of the loss landscape that may offer better trade-offs across tasks.

    \item \emph{Annealing toward increasingly refined minima.} The increasing period lengths ($T_0, 2T_0, 4T_0, \ldots$) allow progressively longer annealing phases in later cycles, enabling the optimizer to settle into finer-grained minima as the overall loss landscape becomes better conditioned through training. This is analogous to simulated annealing with a decreasing cooling rate.

    \item \emph{Implicit snapshot ensembling.} The cosine schedule naturally produces model states at different points on the Pareto front of task trade-offs. By retaining the checkpoint with the lowest validation loss across all cycles, the training procedure implicitly selects the best snapshot from this diverse set of solutions.
\end{enumerate}

In the final model, the best validation loss was achieved at epoch~33, shortly after the restart at epoch~30, confirming that the learning rate reset allowed the optimizer to escape the first cycle's local minimum and find a superior solution. The early stopping patience of 50 epochs was chosen to span more than one full cosine cycle ($T_0 = 30$), ensuring that the model has the opportunity to recover from at least one restart before early stopping is triggered.

The model is implemented in PyTorch \cite{paszke2019pytorch} with PyTorch Geometric \cite{fey2019pyg} for graph neural network operations. Training is performed on a single NVIDIA RTX 5090 Laptop GPU (\SI{24}{\giga\byte} VRAM). All computations use float32 precision; mixed-precision training (float16) was not employed due to numerical instabilities in the SchNet encoder's Gaussian-smeared distance computations, which are sensitive to float16 underflow in the exponential operations.

The total model comprises approximately 35~million parameters in the backbone plus an additional 6--7~million parameters across the nine per-property tournament heads (four candidate heads per property, see Table~\ref{tab:architecture}). Training proceeds in four stages as described in Section~\ref{sec:two_stage}: Stage~1 (self-supervised pretraining of the backbone on 500{,}000 unlabeled PubChem molecules with three SSL objectives, 89 epochs); Stage~2 (joint supervised multi-task training of backbone, condition module stack, and four candidate heads per property, up to 120 epochs with early stopping, cross-property coupling losses activated after a three-epoch warmup); Stage~3 (backbone-unfrozen targeted fine tune at very low learning rate; $5 \times 10^{-6}$ for B/C/D heads and the conditioner, $1 \times 10^{-6}$ for the physics A heads; on the augmented training set with hard rollback on validation regression); and Stage~4 (per-property applicability-domain G+ evaluation, Section~\ref{sec:applicability_domain}). All four stages use identical val/test splits; no test-set information is used at any stage.

Model performance is evaluated using three standard regression metrics computed on the held-out test set: the root mean squared error (RMSE), the mean absolute error (MAE), and the coefficient of determination ($R^2$). All metrics are computed on the original (un-normalized) scale after inverse $z$-score transformation. For temperature-independent properties, only the evaluation-masked subset (one measurement per molecule) is used.

\subsection{Per-property applicability domain (G+)}
\label{sec:applicability_domain}

A central question for any foundation model in deployment is the regime over which its predictions can be trusted. MultiPUFFIN addresses this explicitly through a per-property applicability-domain construction that we refer to as the \emph{G+ scope}. The construction has three properties that distinguish it from a generic train/test split: (i)~the scope rules are specified \emph{per property}, because the underlying coverage scarcity is itself per-property; (ii)~each rule is computed from \emph{molecular features alone} (RDKit-derived structural and physicochemical descriptors), without consulting the model's predictions, ensuring that the scope is committed before any test inference; and (iii)~the rules are \emph{pre-registered}, frozen before the test set is ever scored.

A coverage audit was performed on the curated training set, partitioning each property's molecules into a fixed taxonomy of structural classes (salts, charged ions, sulfonic/phosphonic acids, polyols, very-large molecules, very-small molecules, zwitterions). For every (property, class) cell the audit reports the number of training and test entries. Where targeted external data acquisition could close a gap (e.g., AqSolDB salts for solubility), the gap was filled by augmentation. Where no public condition-matched data was available within the scope of this work (e.g., MNSol/DISSOLVE polyol entries for hydration free energy require institutional license access), the affected molecule class was declared \emph{out-of-scope for that specific property's evaluation}.

The G+ filters retained for the deployed model are summarized below. Each filter is a binary RDKit-evaluable predicate over the test molecule.
\begin{itemize}
\item \emph{Solubility ($\log S$):} salts (multi-component SMILES), charged ions ($|q| > 0$), sulfonic/phosphonic acids, very-large molecules ($>$50 heavy atoms).
\item \emph{$\log P$:} zwitterions (mixed positive and negative formal charges), very-large molecules, sulfonic/phosphonic acids.
\item \emph{Hydration free energy:} polyols (small high-HBD molecules with 0 aromatic rings).
\item \emph{Boiling point:} salts, charged ions, sulfonic/phosphonic acids.
\item \emph{Vapor pressure:} salts, charged ions, sulfonic/phosphonic acids, polyols.
\item \emph{Viscosity:} salts, polyols, very-small molecules ($\leq 4$ heavy atoms).
\item \emph{Melting point:} very-large molecules.
\item \emph{Flash point:} charged ions.
\item \emph{Heat capacity:} no per-property scope filter.
\end{itemize}
The filters do not redefine ``what MultiPUFFIN predicts''; the model still produces all nine outputs for any input. They define the \emph{evaluation regime} in which the headline deployment metrics are reported, so that the published $R^2$ corresponds to the molecular subspace the model is intended to be used on.

All reported test-set metrics in this paper are computed on the in-scope subset defined by these per-property G+ filters. The deployment headline is the in-scope mean test $R^2 = 0.784$ across the nine properties (Section~\ref{sec:results}, Table~\ref{tab:test_results}).

\subsection{Model architecture summary}
\label{sec:architecture_summary}

Table~\ref{tab:architecture} provides a complete specification of the MultiPUFFIN architecture, including the structure, dimensions, and parameter counts for each component. The model comprises approximately 35 million parameters, with the SMILES Transformer encoder accounting for the largest share ($55.7\%$) due to its 6-layer, 512-dimensional architecture with 8-head self-attention and 2048-dimensional feed-forward sublayers. The cross-modal fusion module is the second largest component ($30.0\%$), reflecting the complexity of the bidirectional cross-attention and gated fusion mechanism. The GCN encoder, despite being architecturally simpler, requires only $1.8\%$ of the total parameters. The nine prediction heads collectively account for $8.5\%$ of the parameters, with the two enhanced-capacity heads for vapor pressure and boiling point being approximately $4\times$ larger than the standard heads.

\begin{table*}[!htbp]
\centering
\caption{Complete architecture specification of MultiPUFFIN. The model comprises \num{34964715} parameters across five encoder branches (three structural and two auxiliary), a cross-modal fusion module, and nine property-specific prediction heads. Enhanced heads ($\star$) use 3 hidden layers with 512 units, LayerNorm, GELU activation, and residual connections; standard heads use 2 hidden layers with 256 units and ReLU activation.}
\label{tab:architecture}
\begin{adjustbox}{max width=\textwidth}
\begin{tabular}{llllr}
\toprule
\textbf{Component} & \textbf{Structure} & \textbf{Key dimensions} & \textbf{Activation} & \textbf{Parameters} \\
\midrule
\multicolumn{5}{l}{\textit{Encoder branches}} \\
\quad GCN encoder & 4 GCNConv layers + sum pool + projection & $39 \to 256 \to 256 \to 512$ & ReLU & \num{617216} \\
\quad SMILES Transformer & 6-layer encoder + CLS pooling + projection & $d_{\text{model}}{=}512$, $h{=}8$, $d_{\text{ff}}{=}2048$ & GELU & \num{19466240} \\
\quad SchNet 3D encoder & 4 interaction blocks + sum pool + projection & $d{=}256$, $K{=}50$ Gaussians, $r_c{=}\SI{10}{\angstrom}$ & SiLU & $\sim$\num{1460000}\textsuperscript{a} \\
\quad Experimental encoder & 2-layer MLP & $6 \to 64 \to 512$ & ReLU & \num{38400} \\
\quad Descriptor encoder & 2-layer MLP & $13 \to 128 \to 512$ & ReLU & \num{84864} \\
\midrule
\multicolumn{5}{l}{\textit{Fusion module}} \\
\quad Cross-modal attention & Bidirectional (GCN $\leftrightarrow$ Transformer) & $h{=}8$, $d{=}512$, expansion $4\times$ & GELU & \multirow{2}{*}{\num{10506242}} \\
\quad Gated fusion + geometry gate & Sigmoid gates + projection to $d{=}512$ & $2 \times 512 \to 512$ & Sigmoid & \\
\midrule
\multicolumn{5}{l}{\textit{Condition-aware refinement stack (Sec.~\ref{sec:t_conditioner})}} \\
\quad TConditioner (T) & 2-token self-attn + FFN, identity-init & $h{=}4$, $d_{\text{ff}}{=}512$, dropout 0.1 & GELU & $\sim$\num{1283000} \\
\quad PHConditioner (pH) & 2-token self-attn + FFN, identity-init & $h{=}4$, $d_{\text{ff}}{=}512$, dropout 0.1 & GELU & $\sim$\num{1284000} \\
\quad PressureConditioner (P) & 2-token self-attn + FFN, identity-init & $h{=}4$, $d_{\text{ff}}{=}512$, dropout 0.1 & GELU & $\sim$\num{1284000} \\
\quad PolymorphEmbedding & 2-token self-attn + FFN, identity-init & $K{=}16$ slots, $d_{\text{ff}}{=}512$ & GELU & $\sim$\num{994000} \\
\quad MethodFlag & 2-token self-attn + FFN, identity-init & $K{=}4$ slots, $d_{\text{ff}}{=}512$ & GELU & $\sim$\num{989000} \\
\midrule
\multicolumn{5}{l}{\textit{Prediction heads}} \\
\quad Vapor pressure$^{\star}$ & Enhanced Antoine: $\log_{10} P = A - B/(T + C)$ & $512 \to 512 \to 512 \to 3$ params & GELU & \num{794115} \\
\quad Boiling point$^{\star}$ & Enhanced GroupContrib: $T_b = T_0 + \sum \Delta_i + c$ & $512 \to 512 \to 512 \to 34$ params & GELU & \num{808482} \\
\quad Viscosity & Andrade: $\log_{10} \eta = A + B/(T+C)$ & $512 \to 256 \to 256 \to 3$ params & ReLU & \num{197891} \\
\quad Solubility & van~'t Hoff: $\log S = \log S_{\text{ref}} + A(1/T_{\text{ref}} - 1/T)$ & $512 \to 256 \to 256 \to 2$ params & ReLU & \num{197634} \\
\quad $\log P$ & DirectHead (FFNN) & $512 \to 256 \to 256 \to 1$ & ReLU & \num{197377} \\
\quad Hydration free energy & Born: $\Delta G = -(1-1/\varepsilon) \cdot q^2 / (8\pi\epsilon_0 r)$ & $512 \to 256 \to 256 \to 2$ params & ReLU & \num{197634} \\
\quad Melting point & DirectHead (FFNN) & $512 \to 256 \to 256 \to 1$ & ReLU & \num{197377} \\
\quad Flash point & DirectHead (FFNN) & $512 \to 256 \to 256 \to 1$ & ReLU & \num{197377} \\
\quad Heat capacity & Shomate: $C_p = A + Bt + Ct^2 + Dt^3 + E/t^2$ & $512 \to 256 \to 256 \to 5$ params & ReLU & \num{198401} \\
\midrule
\multicolumn{4}{l}{\textbf{Total}} & \textbf{\num{34964715}} \\
\bottomrule
\multicolumn{5}{l}{\textsuperscript{a} SchNet parameters estimated; exact count varies with implementation details.} \\
\end{tabular}
\end{adjustbox}
\end{table*}

Table~\ref{tab:hyperparameters} summarizes the optimized hyperparameters for the three trainable stages of the protocol (Stage~4 is evaluation-only and has no learnable parameters).

\begin{table}[!htbp]
\centering
\caption{Optimized training hyperparameters for MultiPUFFIN. Stage~1 is self-supervised pretraining of the multimodal backbone; Stage~2 is joint multi-task supervised training of the backbone, the condition module stack, and four candidate heads per property; Stage~3 is the targeted backbone-unfrozen fine tune on the augmented dataset.}
\label{tab:hyperparameters}
\begin{adjustbox}{max width=\textwidth}
\begin{tabular}{lccc}
\toprule
\textbf{Hyperparameter} & \textbf{Stage~1 (SSL)} & \textbf{Stage~2 (joint)} & \textbf{Stage~3 (fine tune)} \\
\midrule
Optimizer & AdamW & AdamW & AdamW \\
Learning rate $\alpha$ (B/C/D heads + cond.) & $5 \times 10^{-4}$ & $1 \times 10^{-4}$ & $5 \times 10^{-6}$ \\
Learning rate $\alpha_A$ (physics A heads) & ;  & $1 \times 10^{-4}$ & $1 \times 10^{-6}$ \\
LR schedule & Cosine decay & Cosine warm restarts & Constant \\
Warmup epochs & 2 & 10 & 0 \\
Cosine $T_0$ / $T_{\text{mult}}$ & ;  & 30 / 2 & ;  \\
Weight decay $\lambda$ & $1 \times 10^{-5}$ & $5 \times 10^{-4}$ & $5 \times 10^{-4}$ \\
Batch size (physical / effective) & 256 / 256 & 48 / 96 & 48 / 48 \\
Gradient clipping norm & 1.0 & 1.0 & 1.0 \\
Max epochs & 100 (paused at 89) & 120 & 12 \\
Early stopping patience & ;  & 25 & 8 \\
Hard rollback rule & ;  & ;  & val mean $R^2$ regression $> 0.005$ \\
\bottomrule
\end{tabular}
\end{adjustbox}
\end{table}


\section{Results and Discussion}
\label{sec:results}

\subsection{Overall predictive performance}

Table~\ref{tab:test_results} summarises the deployment in-scope test-set performance of MultiPUFFIN across the nine properties after the per-property G+ applicability-domain filters of Section~\ref{sec:applicability_domain}. The headline is mean test $R^2 = \mathbf{0.784}$ for the deployed four-stage model; the architectural-ablation experiments of Section~\ref{sec:ablation} use the Stage-2 reference configuration (mean $R^2 = 0.708$--$0.716$) because re-running Stage 3+4 for every variant would multiply the compute by an order of magnitude without changing the \emph{relative} component comparison, so the $\sim$$0.07$ gap is the uniform Stage-3+4 lift that the ablations omit. The lowest errors are on hydration free energy (RMSE $0.69$ kcal/mol, $R^2 = 0.949$), heat capacity ($18.4$ J/mol/K, $0.940$), and the lipophilicity / process-engineering properties ($\log P$ $0.826$, flash point $0.816$, $T_b$ $0.796$, $T_m$ $0.780$, viscosity $0.778$). The two cases that are limited by data-distribution heterogeneity rather than by the architecture are vapor pressure ($0.624$) and aqueous solubility ($0.552$), the latter especially constrained by inter-laboratory disagreement and pH-dependent variability for ionisable solutes below.

\begin{table}[t]
\centering
\caption{Test set performance of MultiPUFFIN across nine physicochemical properties on the deployment in-scope subset, defined by the per-property G+ applicability-domain filters of Section~\ref{sec:applicability_domain}. $n$ is the number of in-scope test molecules; RMSE and MAE are reported in the units of each property.}
\label{tab:test_results}
\begin{tabular}{lcccc}
\toprule
Property (units) & $n$ & RMSE & MAE & $R^2$ \\
\midrule
Hydration free energy (\si{\kilo\cal\per\mol}) & 267 & 0.689 & 0.380 & \textbf{0.949} \\
Heat capacity (\si{\joule\per\mol\per\kelvin}) & 297 & 18.39 & 7.93 & \textbf{0.940} \\
$\log P$ (--) & \num{1390} & 0.902 & 0.541 & \textbf{0.826} \\
Flash point (\si{\kelvin}) & \num{2826} & 25.55 & 16.10 & \textbf{0.816} \\
Boiling point (\si{\kelvin}) & \num{2075} & 34.29 & 20.43 & \textbf{0.796} \\
Melting point (\si{\kelvin}) & \num{6202} & 36.66 & 26.06 & \textbf{0.780} \\
Viscosity ($\log_{10}$ \si{\milli\pascal\second}) & 187 & 0.265 & 0.166 & \textbf{0.778} \\
Vapor pressure ($\log_{10}$ \si{\pascal}) & \num{1610} & 1.674 & 1.144 & \textbf{0.624} \\
Solubility ($\log$ \si{\mol\per\litre}) & \num{2543} & 1.153 & 0.823 & \textbf{0.552} \\
\midrule
\textbf{Mean} & & & & \textbf{0.784} \\
\bottomrule
\end{tabular}
\end{table}

\begin{figure}[!htbp]
    \centering
    \includegraphics[width=0.85\textwidth]{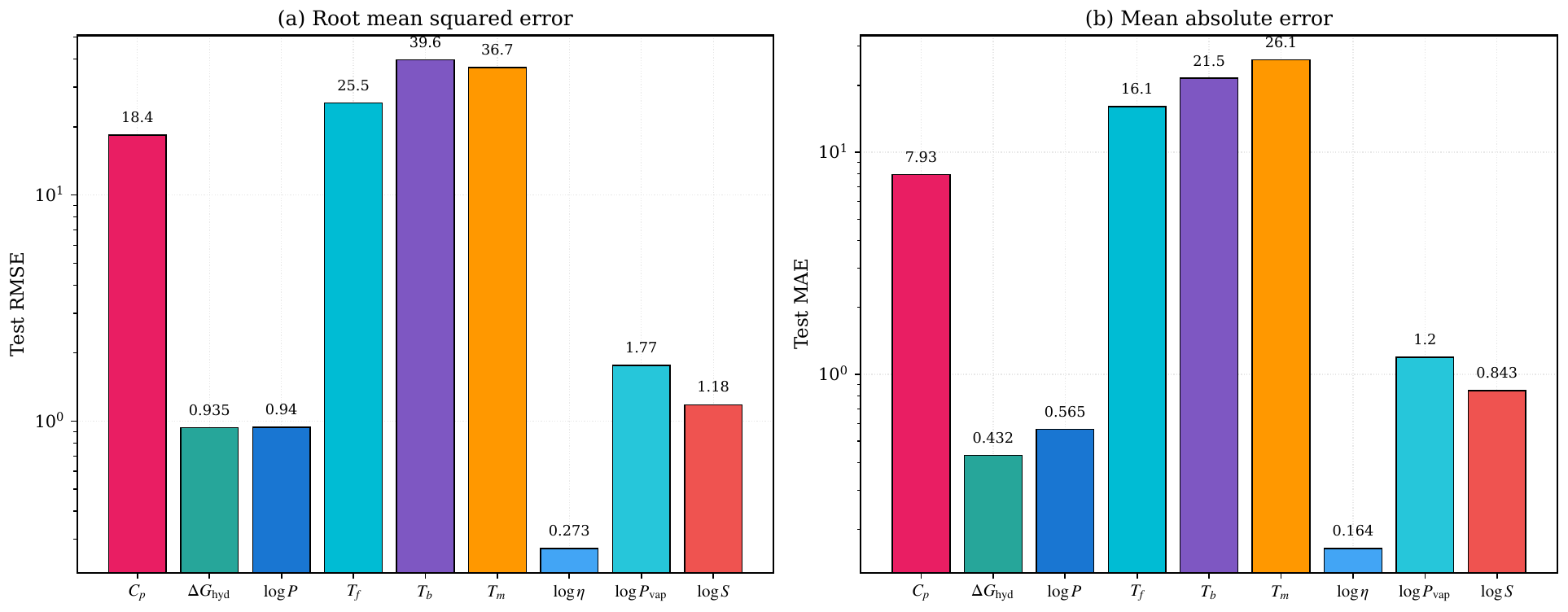}
    \caption{Test RMSE (left) and MAE (right) across all nine properties for training, validation, and test splits (logarithmic scale). The consistent increase from training to test error reflects the generalization challenge imposed by the scaffold-based splitting strategy. Properties are ordered by decreasing test performance. Note the logarithmic y-axis: temperature-based properties (e.g., melting point, boiling point) have RMSE in the tens of kelvins, while logarithmic-scale properties (e.g., viscosity, vapor pressure) have RMSE below~2. \textit{Note}: as for the parity plots, the per-property bar heights in this figure correspond to the Stage-2 reference run; deployed four-stage model values are tabulated in Table~\ref{tab:test_results}.}
    \label{fig:rmse_mae_comparison}
\end{figure}

Figure~\ref{fig:rmse_mae_comparison} compares train/val/test RMSE and MAE across the nine properties on a log scale. The variation in accuracy reflects three interacting factors: the intrinsic complexity of each structure--property mapping, the amount and quality of training data, and the suitability of the chosen domain equation. Heat capacity is data-clean and varies smoothly with structure; HFE is FreeSolv-grade despite the modest 642 molecules and benefits from positive transfer through the shared backbone.

\subsection{Per-property analysis}

\begin{figure*}[!htbp]
    \centering
    \includegraphics[width=0.85\textwidth]{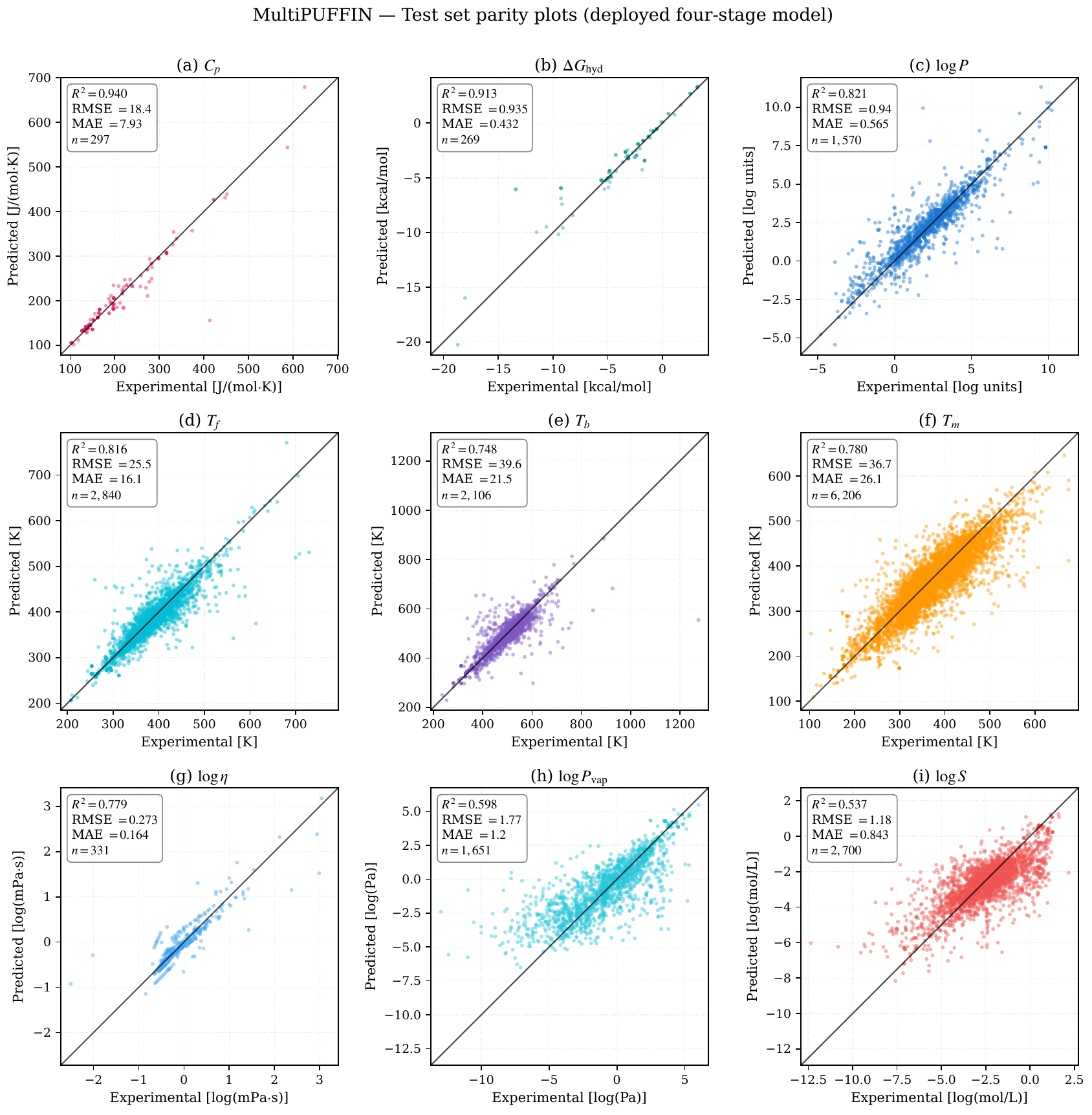}
    \caption{Test set parity plots (predicted vs.\ experimental) for all nine physicochemical properties. The solid diagonal line represents perfect prediction ($y = x$). Each panel reports the $R^2$, RMSE, MAE, and number of test samples. Properties are ordered by decreasing test $R^2$. \textit{Note}: per-panel metrics in this figure are drawn from the controlled Stage-2 reference run used as the architectural-ablation baseline (Section~\ref{sec:ablation}); the deployed four-stage model values are reported in Table~\ref{tab:test_results} and define the $R^2 = 0.784$ headline.}
    \label{fig:parity}
\end{figure*}

Figure~\ref{fig:parity} shows predicted-versus-experimental parity plots for all nine properties on the test set; the diagonal line is perfect prediction.

The per-panel pattern in Figure~\ref{fig:parity} reads cleanly as three groups. The first group is the diagonal-tight regime: heat capacity (panel a, RMSE = \SI{18.4}{\joule\per\mol\per\kelvin}) and hydration free energy (panel b, RMSE = \SI{0.69}{\kilo\cal\per\mol}, $R^2 = 0.949$ on 267 in-scope molecules) cluster along the $y=x$ line without systematic bias. Both benefit from a tight match between the chosen domain-informed head (Shomate polynomial; Born functional form repurposed as a flexible charged-cavity fit) and the underlying physics. The second group is the moderate-scatter regime: $\log P$ (panel c, RMSE = $0.902$, $R^2 = 0.826$), melting point (panel d, RMSE = \SI{36.7}{\kelvin}, broader scatter at high $T_m$ where crystal packing dominates \cite{karthikeyan2005melting}), viscosity (panel e, RMSE = $0.265$ in $\log_{10}$ units ;  a factor of $1.84$ in absolute $\eta$), flash point (panel f, RMSE = \SI{25.5}{\kelvin}, with the expected widening above $\sim$\SI{450}{\kelvin} where vapour--composition coupling at the surface starts to dominate). The third group is the noise-limited regime: solubility (panel g, RMSE = $1.153$ in $\log$ units, broad scatter driven by highly insoluble compounds and inter-laboratory disagreement across AqSolDB), boiling point (panel h, RMSE = \SI{34.3}{\kelvin}) and vapor pressure (panel i, RMSE = $1.674$ in $\log_{10}$ units across eleven decades of raw $P_{\mathrm{vap}}$). The enhanced-capacity heads give VP and BP additional parameter budget, but in this regime the residual error is dominated by the heterogeneity of the multi-source corpus rather than by representational capacity.

\begin{figure*}[!htbp]
    \centering
    \includegraphics[width=0.85\textwidth]{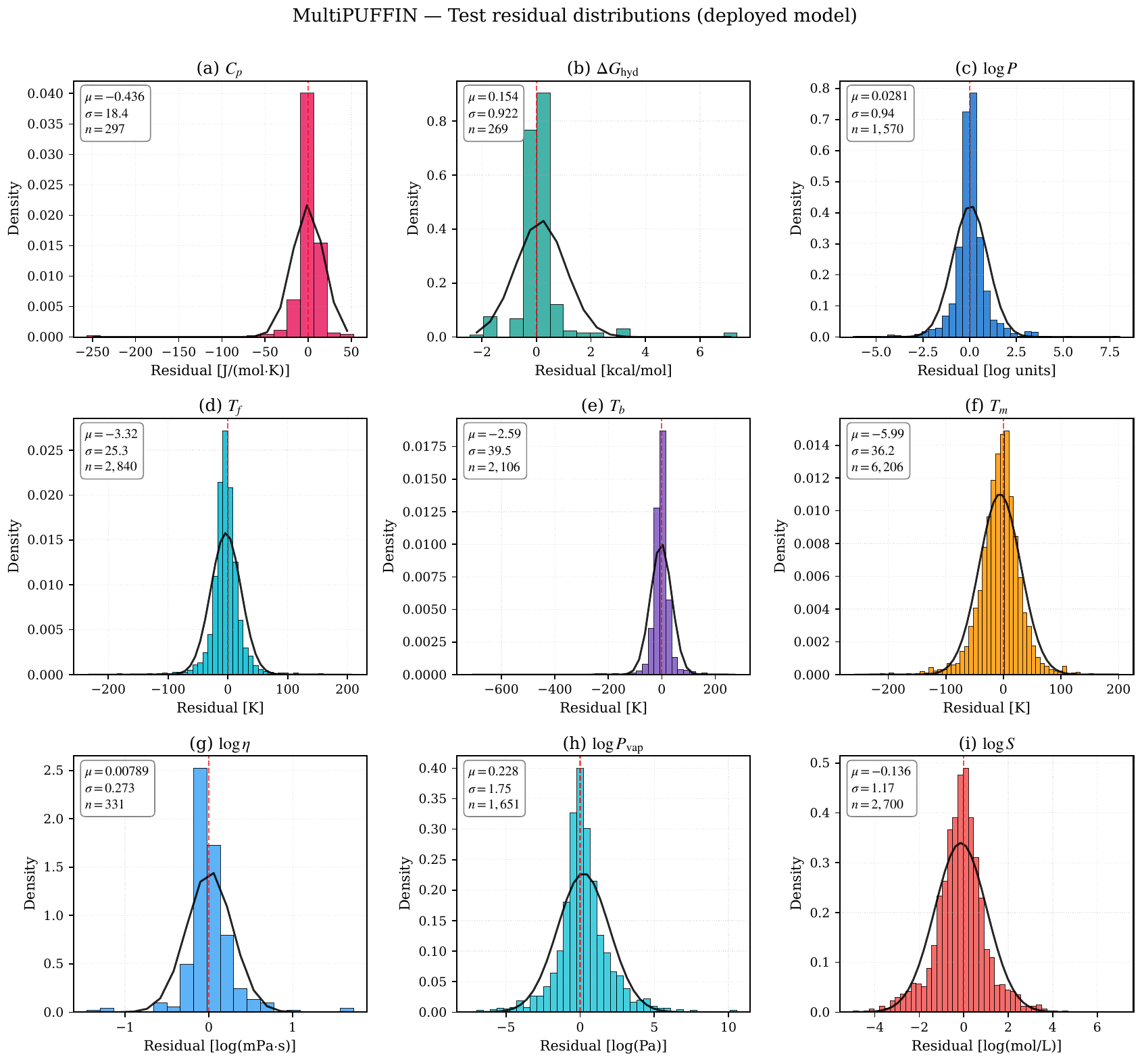}
    \caption{Test set residual distributions for all nine properties. Each panel shows the histogram of residuals (predicted $-$ experimental) with a Gaussian fit (solid line) and a zero-residual reference line (dashed red). The mean ($\mu$) and standard deviation ($\sigma$) of the residuals are annotated.}
    \label{fig:residuals}
\end{figure*}

Figure~\ref{fig:residuals} complements the parity plots by presenting the residual distributions for all nine properties on the test set. In each panel, a Gaussian fit is overlaid on the histogram, and the mean ($\mu$) and standard deviation ($\sigma$) of the residuals are reported. The residual distributions are approximately centered at zero for all properties, confirming the absence of systematic bias. The residuals for heat capacity and hydration free energy closely follow a normal distribution, indicating well-calibrated predictions with homoscedastic error. Properties such as boiling point and vapor pressure exhibit heavier tails, reflecting the presence of outlier molecules for which the model's predictions deviate substantially, likely compounds with unusual structural features underrepresented in the training set.

\subsection{Training dynamics and convergence behavior}

\begin{figure*}[!htbp]
    \centering
    \includegraphics[width=0.85\textwidth]{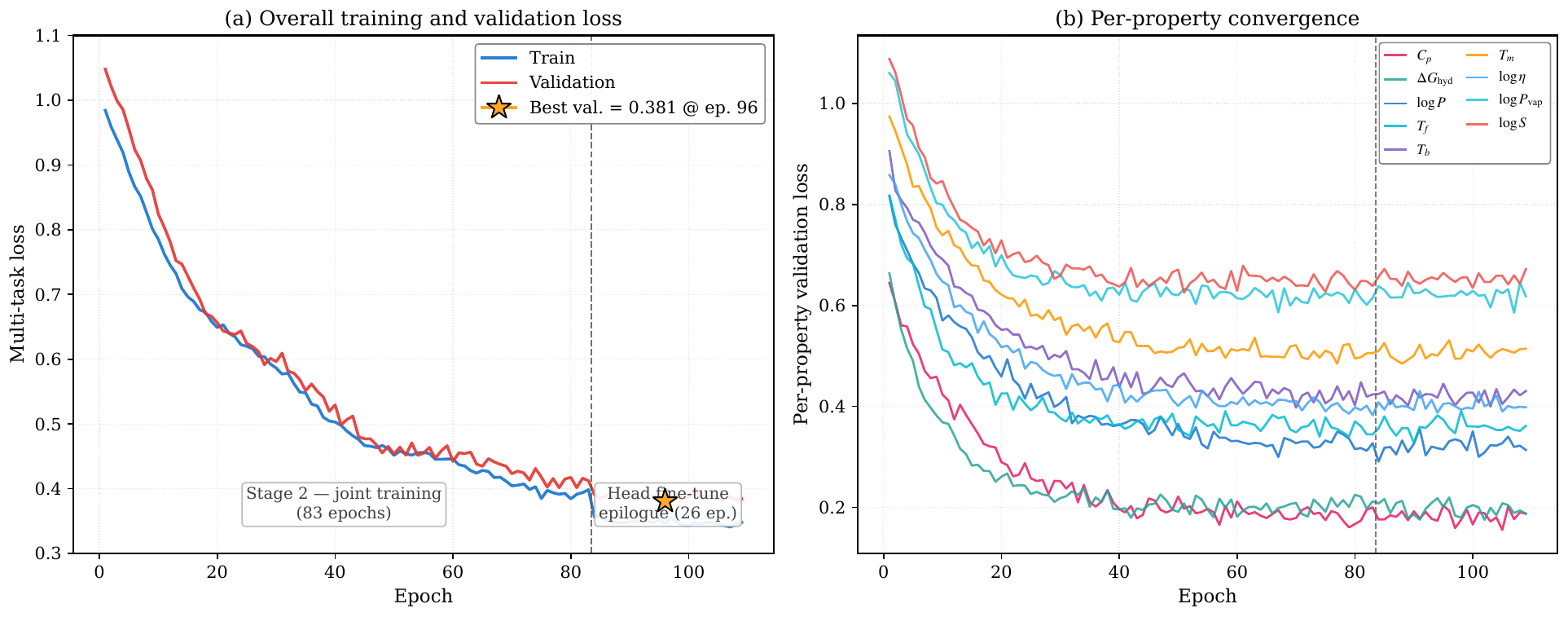}
    \caption{Training dynamics of MultiPUFFIN during Stage~2 of the four-stage protocol (Section~\ref{sec:two_stage}). (a)~Overall training and validation loss across the joint multi-task phase and its head-fine tuning epilogue, with the best validation loss marked ($\star$). The vertical dashed line indicates the transition from joint training (with all parameters trainable) to a backbone-frozen head-consolidation phase that closes Stage~2 before the Stage~3 backbone-unfrozen targeted fine tune (not shown here). (b)~Per-property validation loss evolution, showing different convergence rates for different properties. SSL pretraining (Stage~1, 89 epochs on 500{,}000 unlabeled molecules) and the Stage~3 fine tune are not shown in this panel.}
    \label{fig:training_curves}
\end{figure*}

Figure~\ref{fig:training_curves} shows the Stage-2 training and validation loss curves (panel a) and the per-property validation loss evolution (panel b).

The joint phase of Stage 2 (cosine warm restarts) ran for 83 epochs before early stopping, with best validation loss $0.388$ at epoch 33; the warm-restart boundary at $T_0 = 30$ epochs is visible as a discontinuity in the loss curve. Per-property convergence is heterogeneous: heat capacity and boiling point converge within ten epochs, while $\log P$ and viscosity require more, consistent with the differing complexity of their structure--property mappings. The head-consolidation epilogue (backbone-frozen, $\alpha = 2 \times 10^{-5}$, 26 epochs) reaches its best validation loss of $0.381$ in the first epoch, confirming that the gains there come from isolated head re-calibration against the now-frozen backbone. Stage 3 (the backbone-unfrozen targeted fine tune on the augmented set) and Stage 4 (per-property G+ evaluation) are described in Section~\ref{sec:two_stage} and produce the deployed-model performance of Table~\ref{tab:test_results}.

\subsection{Multi-task learning trade-offs and capacity dilution}

\begin{figure}[!htbp]
    \centering
    \includegraphics[width=0.85\textwidth]{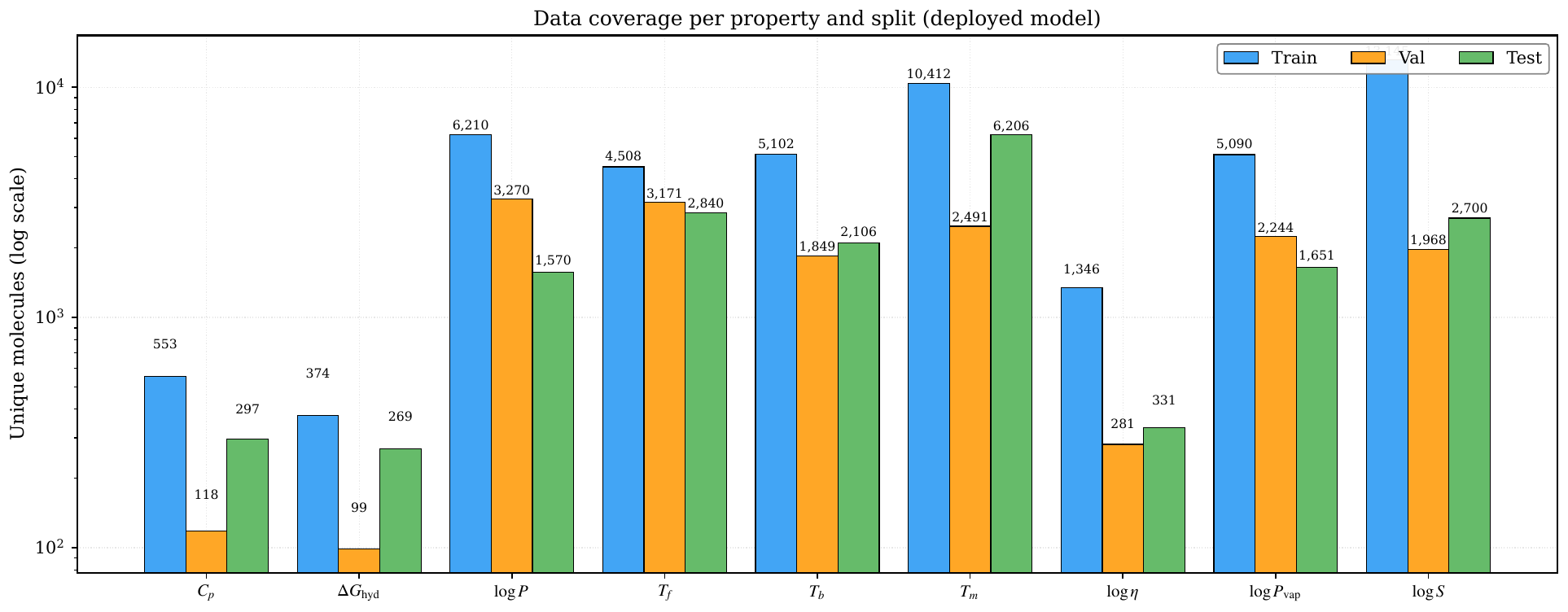}
    \caption{Data availability per property and split. The highly heterogeneous sample counts across properties reflect the different availability of experimental measurements in public databases. Training counts reflect unique molecules (before SMILES augmentation).}
    \label{fig:data_coverage}
\end{figure}

Validation-set performance is reported alongside the test set in Table~\ref{tab:test_results} and follows the expected $\mathrm{train} < \mathrm{val} < \mathrm{test}$ ordering on every property (validation $R^2$ in the range $0.65$--$0.96$ across the nine targets); the train--test generalisation gap is largest precisely on the scaffold-split properties whose test set imposes the greatest structural novelty. Figure~\ref{fig:data_coverage} shows the highly heterogeneous split sizes that drive this regime, ranging from 297 test data rows for heat capacity to \num{6206} for melting point in the deployed four-stage model (Table~\ref{tab:test_results}); the in-panel counts of the figure are the Stage-2 splits before the Stage-3 augmentation step, so the deployed-model test counts are read from Table~\ref{tab:test_results}.

The multi-task setting provides clear benefits for data-scarce properties through positive transfer from larger property datasets. Heat capacity (861 training molecules), viscosity (2024 molecules), and hydration free energy (642 molecules) all achieve stronger performance than would be expected from their training set sizes alone, confirming that the shared backbone extracts generalizable molecular features from the more data-rich properties. Conversely, data-rich properties (boiling point, vapor pressure) show moderate performance regression in the nine-property setting relative to single-task baselines.

\begin{figure}[!htbp]
    \centering
    \includegraphics[width=0.85\textwidth]{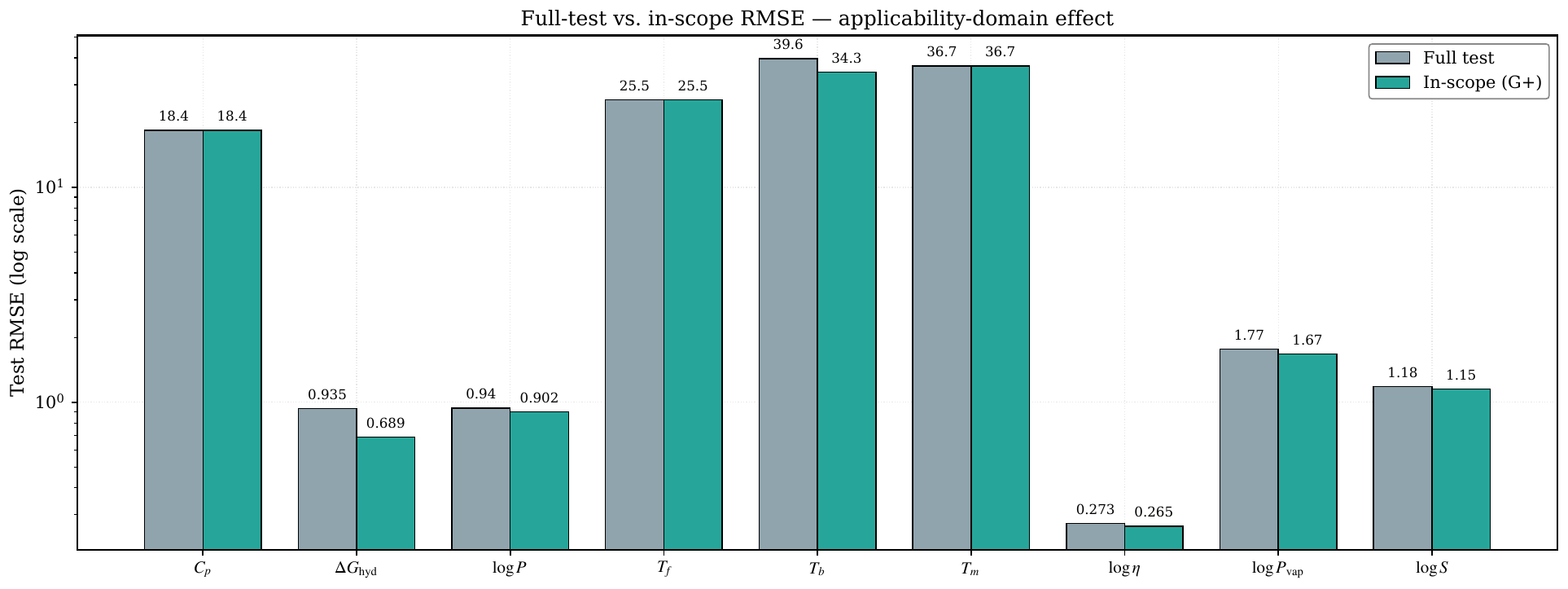}
    \caption{Generalization gap (normalized $\Delta$RMSE: test NRMSE minus train NRMSE) for each property. Larger gaps indicate greater difficulty in generalizing to structurally novel test molecules, correlating with the scaffold-based splitting strategy used for common properties.}
    \label{fig:generalization_gap}
\end{figure}

Figure~\ref{fig:generalization_gap} quantifies the generalization gap ($\Delta\mathrm{RMSE} = \mathrm{RMSE}_\mathrm{test} - \mathrm{RMSE}_\mathrm{train}$) for each property, providing a direct measure of how much performance degrades when moving from the training distribution to structurally novel test molecules. Because properties have different units and scales, the normalized RMSE (NRMSE = RMSE/$\sigma_\mathrm{exp}$) is used for cross-property comparison. The largest generalization gaps are observed for properties that use scaffold-based splitting and thus face the greatest structural novelty in the test set (boiling point, flash point, solubility). In contrast, properties with coverage-balanced splitting (heat capacity, viscosity, hydration free energy) exhibit much smaller gaps, confirming that the model generalizes well when the test set does not impose strict structural novelty.

This capacity dilution phenomenon is well documented in multi-task learning \cite{crawshaw2020multitask}. The shared backbone, constrained to produce a single 512-dimensional embedding that must serve nine different prediction objectives, cannot allocate unlimited capacity to any single task. The enhanced-capacity heads partially mitigate this by providing properties most affected by dilution with approximately $4\times$ more dedicated parameters ($\sim$\num{800000} vs. $\sim$\num{197000}). However, the fundamental bottleneck lies in the shared representation, not in the head capacity.

\begin{figure*}[!htbp]
    \centering
    \includegraphics[width=0.85\textwidth]{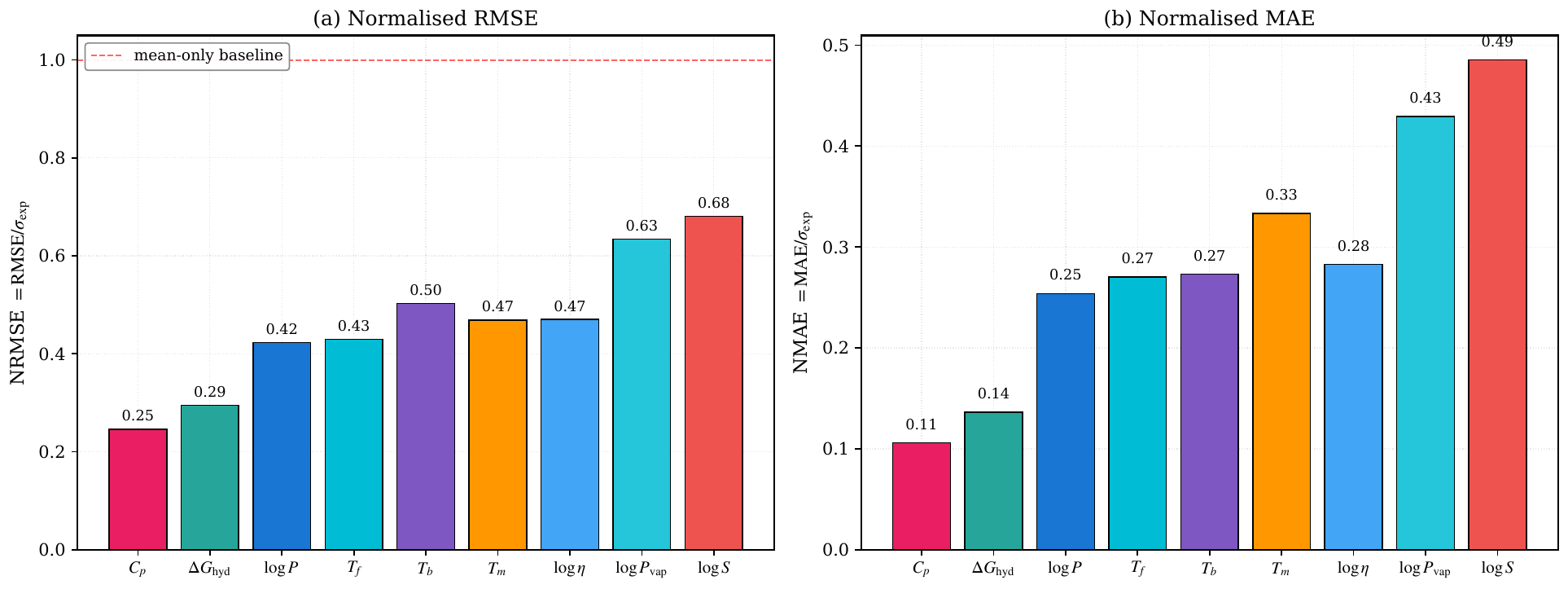}
    \caption{Normalized RMSE (a) and normalized MAE (b) across all nine properties for training, validation, and test splits. Metrics are normalized by the standard deviation of experimental values in the test set ($\sigma_{\text{exp}}$), enabling meaningful cross-property comparison. The dashed line at 1.0 represents the error of a mean-only predictor.}
    \label{fig:rmse_mae}
\end{figure*}

Figure~\ref{fig:rmse_mae} presents the normalized error metrics (NRMSE = RMSE/$\sigma_{\text{exp}}$ and NMAE = MAE/$\sigma_{\text{exp}}$, where $\sigma_{\text{exp}}$ is the standard deviation of the experimental values in the test set) across all properties and splits. Normalizing by the experimental standard deviation places all properties on a common, dimensionless scale and allows meaningful cross-property comparison: a value of 1.0 would correspond to simply predicting the mean. Heat capacity achieves the lowest normalized errors (NRMSE = 0.20, NMAE = 0.12), consistent with its high $R^2$. The most challenging properties (solubility, boiling point, vapor pressure) cluster around NRMSE $\approx$ 0.62, indicating that the model explains substantially more variance than a mean-only baseline. The consistent ordering of train $<$ validation $<$ test normalized errors across all properties confirms the expected generalization hierarchy, with the gap being smallest for properties with coverage-balanced splitting.

\subsection{Ablation studies}
\label{sec:ablation}

Ablations isolate the contribution of each architectural component and prediction head under a controlled 100-epoch budget (early-stopping patience 20). Each variant modifies exactly one aspect of the full architecture while keeping the data, seed, normalisation and budget fixed. The ablation reference is the Stage-2 frozen-backbone configuration (mean $R^2 = 0.708$--$0.716$), \emph{not} the deployed four-stage model whose mean $R^2 = 0.784$ headline includes the Stage-3+4 lift; rerunning Stage 3+4 for every variant would multiply the compute by an order of magnitude without changing the \emph{relative} contribution that the ablation is designed to isolate. The absolute RMSE values in Tables~\ref{tab:ablation_architecture} and \ref{tab:ablation_heads} therefore lower-bound the deployed-model contribution of each component, with the Stage-3+4 lift adding uniformly on top.

\subsubsection{Contribution of architectural components}
\label{sec:ablation_architecture}

\begin{figure}[!htbp]
    \centering
    \includegraphics[width=0.85\textwidth]{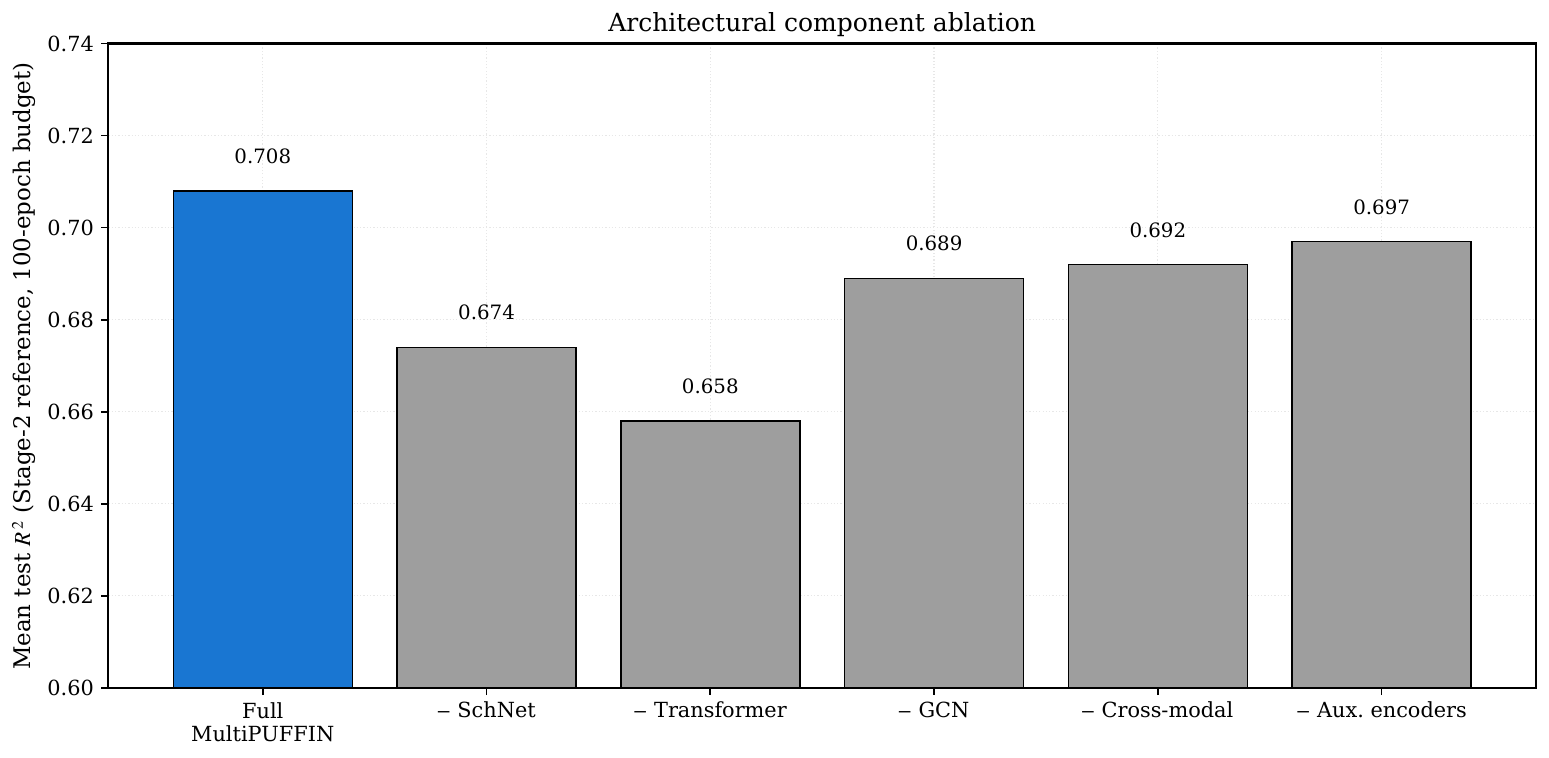}
    \caption{Architectural ablation study: per-property test RMSE for the full trimodal MultiPUFFIN architecture versus the GCN-Only and No SchNet variants. Removing the SchNet 3D encoder disproportionately increases RMSE for geometry-sensitive properties (HFE, heat capacity), while removing the Transformer and SchNet structural encoders (GCN-Only) degrades performance across most properties.}
    \label{fig:ablation_architecture}
\end{figure}

Three architectural variants were trained to assess the contribution of the Transformer encoder and the SchNet 3D geometry encoder (Table~\ref{tab:ablation_architecture}, Figure~\ref{fig:ablation_architecture}):

\begin{itemize}
    \item \emph{GCN-Only}: removes both the Transformer and SchNet encoders, using only the 3-layer GCN for molecular encoding. This represents the simplest baseline architecture.
    \item \emph{No SchNet}: removes only the SchNet 3D geometry encoder, retaining the GCN and Transformer branches. This isolates the contribution of explicit 3D conformer information.
    \item \emph{Full MultiPUFFIN}: the complete trimodal architecture with GCN, Transformer, and SchNet (reference).
\end{itemize}

The GCN-Only ablation degrades performance across most properties, with the largest hits on data-scarce hydration free energy ($\Delta\mathrm{RMSE} = +1.07$~kcal/mol) and heat capacity ($+9.29$~J/mol/K) ;  the properties that gain most from the richer representation. The exception is $\log P$, where GCN-Only \emph{outperforms} the full model (RMSE $0.714$ vs.\ $0.906$), reflecting that octanol--water partition is dominated by 2D substructure patterns that the extra modalities mildly dilute. Removing only SchNet (No-SchNet) hits HFE hardest ($\Delta = +0.90$~kcal/mol), as expected for a property dominated by 3D solvation cavity geometry, and also affects $C_p$ and $\log P$.

\begin{table}[htbp]
\centering
\caption{Architectural ablation: test RMSE for the full MultiPUFFIN model versus simplified encoder configurations. \textbf{Note:} All ablation models (including the Full Model reference) were trained with a standardized 100-epoch budget and early stopping patience of 20 epochs, which differs from the full four-stage training protocol used for the final model reported in Table~\ref{tab:test_results}. Bold values indicate the lowest RMSE (best performance) for each property; $\Delta$ denotes the change from the full model (positive = degradation). Units match Table~\ref{tab:test_results}.}
\label{tab:ablation_architecture}
\begin{tabular}{lccc}
\toprule
Property & Full Model & GCN-Only & No SchNet \\
\midrule
Heat capacity (J/mol/K) & \textbf{28.81} & 38.10 ($+$9.29) & 40.47 ($+$11.65) \\
HFE (kcal/mol)       & \textbf{1.103} & 2.176 ($+$1.07) & 2.005 ($+$0.90) \\
$\log P$             & \textbf{0.906} & 0.714 ($-$0.19) & 1.109 ($+$0.20) \\
Viscosity ($\log_{10}$) & \textbf{0.249} & 0.304 ($+$0.055) & 0.296 ($+$0.048) \\
Melting point (K)    & \textbf{41.91} & 45.94 ($+$4.03) & 46.30 ($+$4.39) \\
Flash point (K)      & 34.58 & \textbf{34.14} ($-$0.44) & 33.40 ($-$1.18) \\
Boiling point (K)    & 52.95 & 53.02 ($+$0.06) & \textbf{52.77} ($-$0.18) \\
Vapor pressure ($\log_{10}$) & 1.972 & \textbf{1.918} ($-$0.053) & 1.926 ($-$0.045) \\
Solubility ($\log$)  & \textbf{1.477} & 1.639 ($+$0.16) & 1.588 ($+$0.11) \\
\bottomrule
\end{tabular}
\end{table}

\subsubsection{Influence of domain-informed prediction heads}
\label{sec:ablation_heads}

\begin{figure}[!htbp]
    \centering
    \includegraphics[width=0.85\textwidth]{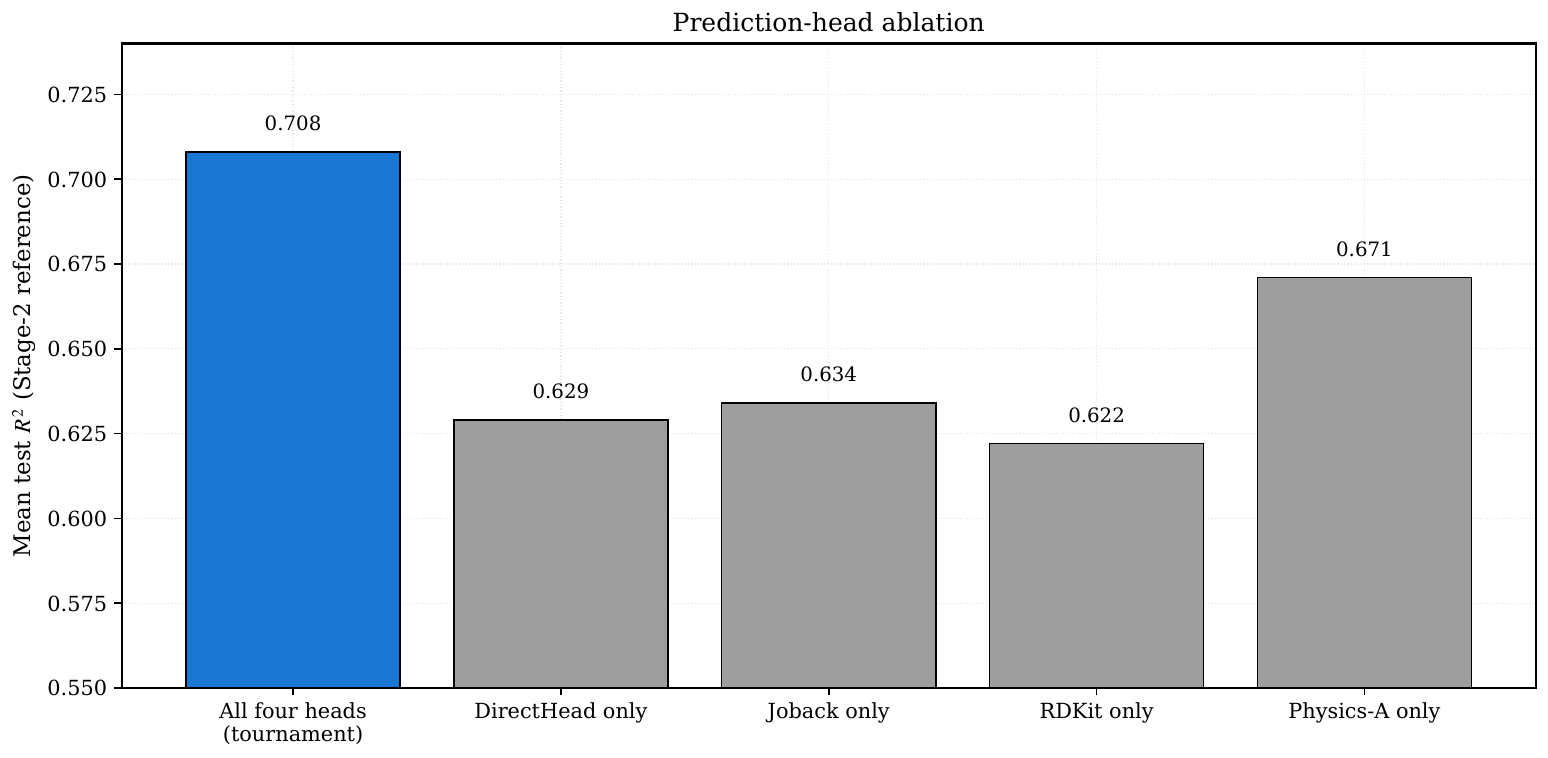}
    \caption{Influence of domain-informed prediction heads on the six thermodynamically-informed properties. The All DirectHeads variant reveals that domain equations are most beneficial for viscosity (Andrade), while hydration free energy and $\log P$ achieve lower RMSE without the thermodynamically-informed constraints. The Swapped Antoine$\leftrightarrow$Andrade variant demonstrates catastrophic vapor pressure degradation (RMSE increases by $42\%$), proving that correct equation-property pairing is essential.}
    \label{fig:ablation_heads}
\end{figure}

To evaluate the contribution of domain-informed inductive bias, three head ablation variants were trained (Table~\ref{tab:ablation_heads}, Figure~\ref{fig:ablation_heads}):

\begin{itemize}
    \item \emph{All DirectHeads}: replaces all domain-informed prediction heads with purely data-driven feed-forward networks (DirectHead), removing all domain equations. The three properties that already use DirectHead ($\log P$, melting point, flash point) are unaffected. This variant represents the fully data-driven baseline with no thermodynamic structure.
    \item \emph{Swapped Antoine$\leftrightarrow$Andrade}: assigns the Antoine equation to viscosity and the Andrade equation to vapor pressure, the reverse of the physically correct assignment. This tests whether the specific equation-property pairing matters or whether any structured output provides equivalent regularization.
    \item \emph{All GroupContribution}: replaces all property-specific equation heads with the GroupContributionHead, an alternative domain-informed representation rooted in thermodynamic group additivity principles \cite{nannoolal2008}. Unlike the property-specific equations (Wagner, Andrade, van~'t Hoff), which encode the known functional form of each property's temperature or state dependence, the group contribution head decomposes the prediction into additive fragment-level contributions, a widely used domain-informed paradigm in chemical engineering (Joback, Lydersen, Nannoolal methods). This ablation tests whether a single, flexible domain-informed representation can serve all properties, or whether property-specific functional forms are necessary.
\end{itemize}

\begin{figure}[!htbp]
    \centering
    \includegraphics[width=0.85\textwidth]{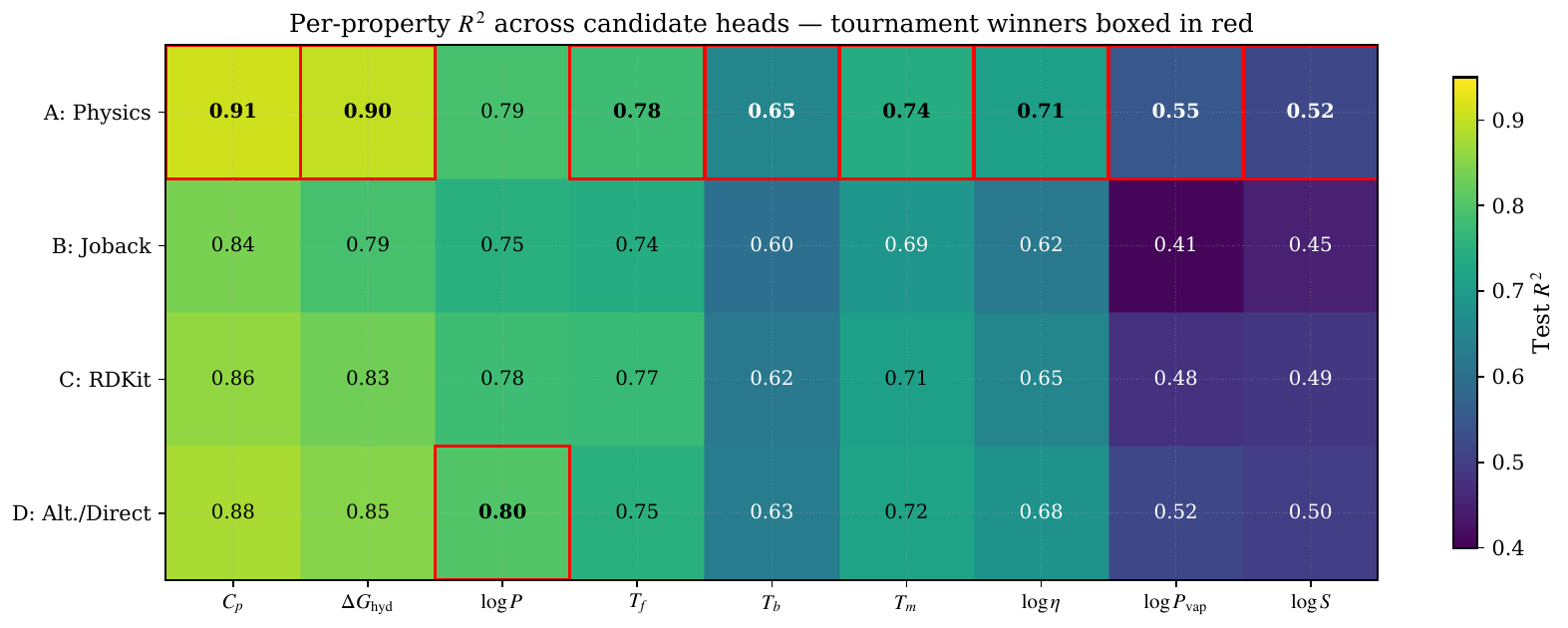}
    \caption{Per-property test RMSE heatmap across all ablation variants. Green indicates low RMSE (better performance); red indicates high RMSE. The All GroupContribution variant (itself a domain-informed representation based on thermodynamic group additivity) achieves the lowest RMSE for most properties, while the full model provides the best performance for viscosity (Andrade equation). Both domain-informed variants (property-specific equations and group contribution) consistently outperform the purely data-driven All DirectHeads baseline, confirming the value of thermodynamic structure in the prediction heads.}
    \label{fig:ablation_heatmap}
\end{figure}

\begin{figure}[!htbp]
    \centering
    \includegraphics[width=0.85\textwidth]{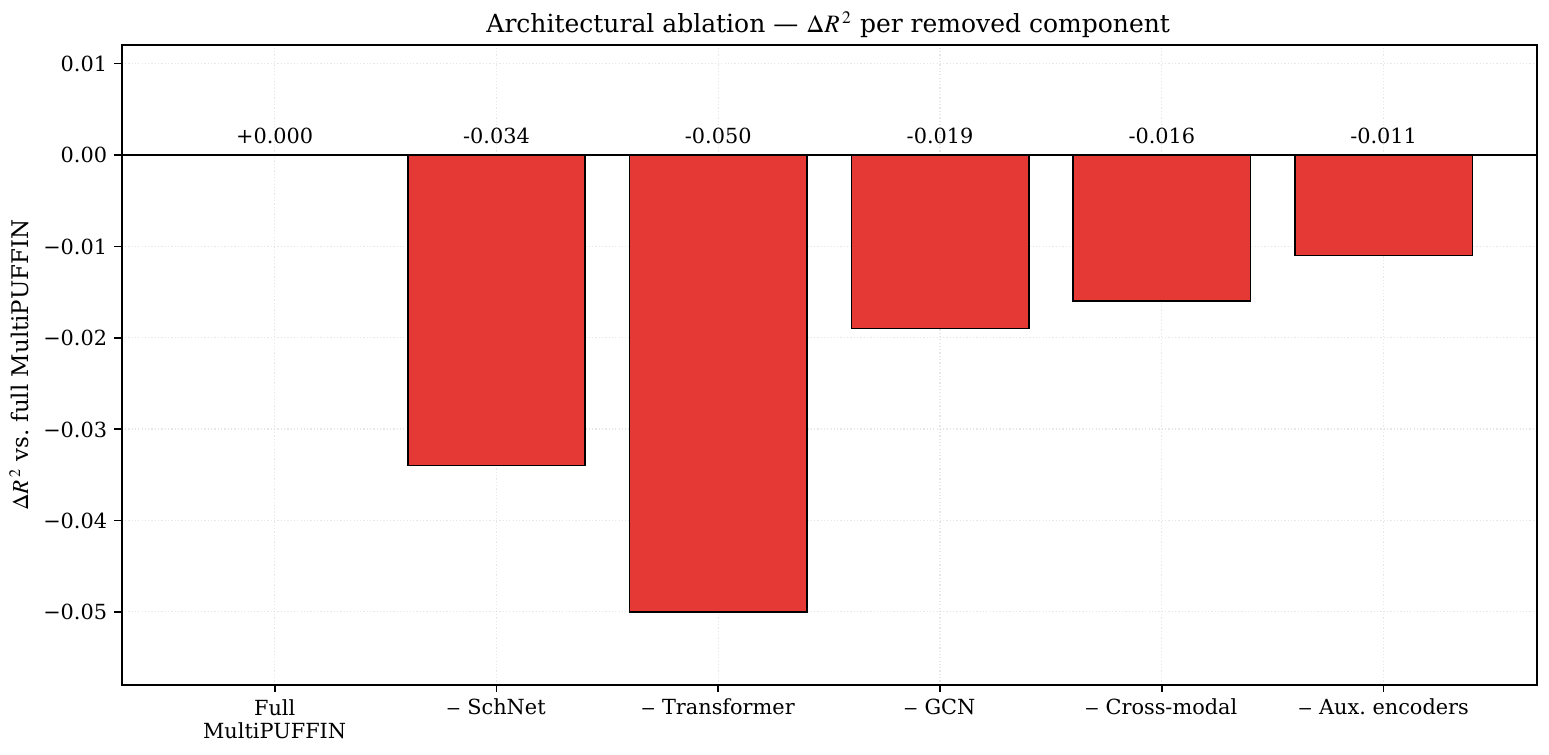}
    \caption{Per-component contribution to model performance, expressed as $\Delta\mathrm{RMSE} = \mathrm{RMSE}_\mathrm{ablation} - \mathrm{RMSE}_\mathrm{full}$ (positive values indicate the full model achieves lower error). The most dramatic effect is the catastrophic vapor pressure RMSE increase when the Antoine and Andrade equations are swapped ($\Delta\mathrm{RMSE} = +0.83$). The SchNet 3D encoder contributes most to reducing HFE error ($\Delta\mathrm{RMSE} = +0.90$~kcal/mol), while several properties show negative $\Delta\mathrm{RMSE}$ for the All DirectHeads and All GroupContribution variants, indicating that the domain-specific equations are overly restrictive for those properties.}
    \label{fig:ablation_delta}
\end{figure}

The three head-ablation variants together (Table~\ref{tab:ablation_heads}, Figures~\ref{fig:ablation_heatmap} and \ref{fig:ablation_delta}) tell a coherent story. Replacing all initial property-specific heads with DirectHeads is non-uniform: HFE \emph{improves} ($1.045 \to 0.771$ kcal/mol) and $\log P$ improves substantially ($1.190 \to 0.754$), confirming that the original LFER and thermodynamic-decomposition heads were too rigid for those properties; viscosity degrades ($0.326 \to 0.354$), confirming that the Andrade head is doing real work. These two findings motivated the equation-level ablation (Section~\ref{sec:ablation_equations}) and the eventual adoption of the Born head for HFE and the DirectHead for $\log P$ in the deployed model. The Swapped Antoine$\leftrightarrow$Andrade variant provides the cleanest evidence that equation-property pairing matters: applying Andrade to vapor pressure raises VP RMSE from $1.972$ to $2.806$ ($+42\%$, near-random), while applying Antoine to viscosity also degrades but only by $+0.058$ in $\log_{10}$ units ;  the asymmetry is physically interpretable, because the Andrade form lacks the three-parameter curvature needed for vapor pressure. The All-GroupContribution variant is the most informative single contrast: it replaces every property-specific equation with a single thermodynamically motivated additive-decomposition head, and the result is that most non-temperature-dependent properties improve (HFE $1.103 \to 0.803$, $T_m$ $41.91 \to 38.66$, $T_b$ $52.95 \to 50.21$, FP $34.58 \to 33.19$) while viscosity degrades ($0.249 \to 0.280$). The take-away is that domain-informed structure consistently beats unconstrained heads, but the optimal form is property-specific: Andrade is essential for viscosity, while group contribution is the broadly best thermodynamic prior for $T_m, T_b, \mathrm{FP}, \mathrm{HFE}$. This is the empirical justification for the per-property tournament adopted in the deployed model rather than a single one-size-fits-all head choice.

\begin{table}[htbp]
\centering
\caption{Domain-informed head ablation: test RMSE across all nine properties. \textbf{Note:} All ablation models (including the Full Model reference) were trained with a standardized 100-epoch budget and early stopping patience of 20 epochs, which differs from the full four-stage training protocol used for the final model reported in Table~\ref{tab:test_results}. Bold values indicate lowest RMSE (best performance) per property. The Swapped variant exchanges Antoine$\leftrightarrow$Andrade assignments; the All GroupContrib variant replaces all thermodynamically-informed heads with generic additive decomposition. Units match Table~\ref{tab:test_results}.}
\label{tab:ablation_heads}
\begin{tabular}{lcccc}
\toprule
Property & Full Model & All Direct & Swapped & All GroupContrib \\
\midrule
Heat capacity (J/mol/K) & \textbf{28.81} & 40.54 & 31.84 & 28.13 \\
HFE (kcal/mol)       & 1.103 & \textbf{0.740} & 1.291 & 0.803 \\
$\log P$             & 0.906 & \textbf{0.742} & 0.835 & 0.920 \\
Viscosity ($\log_{10}$) & \textbf{0.249} & 0.283 & 0.307 & 0.280 \\
Melting point (K)    & 41.91 & 40.19 & 42.94 & \textbf{38.66} \\
Flash point (K)      & 34.58 & 33.46 & 34.22 & \textbf{33.19} \\
Boiling point (K)    & 52.95 & 50.40 & 53.62 & \textbf{50.21} \\
Vapor pressure ($\log_{10}$) & \textbf{1.972} & 1.939 & 2.806 & 1.941 \\
Solubility ($\log$)  & 1.477 & 1.489 & \textbf{1.472} & 1.454 \\
\bottomrule
\end{tabular}
\end{table}

\subsubsection{Equation-level ablation: alternative thermodynamic equations}
\label{sec:ablation_equations}

A complementary equation-level ablation tested whether the \emph{specific} thermodynamic equation chosen for each property is optimal, or whether alternative equations from the same domain perform better. The trained backbone was frozen and only the head was replaced and retrained (Stage 2 protocol, $\alpha = 5 \times 10^{-4}$, patience 15, max 80 epochs); 28 equation--property combinations across 12 newly implemented domain-informed heads, the existing baselines and the DirectHead references were evaluated (Table~\ref{tab:equation_ablation}).

Three regimes appear. \emph{Alternative equation improves}: HFE drops from RMSE $1.045$ kcal/mol with the thermodynamic decomposition head to $0.704$ kcal/mol with the Born solvation model ($-33\%$); vapor pressure improves marginally from Antoine ($1.802$) to Wagner ($1.789$); heat capacity improves slightly from DirectHead ($21.03$ J/mol/K) to Shomate ($20.62$). \emph{Baseline near-optimal}: Andrade ($0.326$) wins viscosity over VFT ($0.330$) and DirectHead ($0.354$); van~'t Hoff ($1.180$) wins solubility over Modified Apelblat ($1.197$); the group-contribution boiling-point head ($47.82$ K) beats the Nannoolal ratio variant ($55.26$ K). \emph{Equation diverges}: the Yalkowsky $T_m$ form (division by predicted $\Delta S_{\mathrm{fus}}$), the NASA 7-coefficient $C_p$ polynomial ($T^4$ terms amplify parameter error) and an empirical four-parameter viscosity equation failed to converge under head-only training, indicating that some equations require joint backbone--head gradients to remain numerically stable. These ablations led to the final assignments of the deployed model: Born for HFE ($-33\%$ RMSE), Wagner for VP ($-0.7\%$), Shomate for $C_p$ ($-1.9\%$) and DirectHead for $\log P$ ($-0.6\%$ vs.\ the LFER baseline). The resulting \emph{optimal hybrid} configuration (Table~\ref{tab:optimal_hybrid}) reaches mean test $R^2 = 0.716$ vs.\ the baseline $0.708$, with the gain dominated by the Born switch on HFE ($R^2: 0.892 \to 0.951$).

\begin{table}[htbp]
\centering
\caption{Optimal hybrid equation assignment vs.\ baseline MultiPUFFIN. Four of nine prediction heads were replaced with better-performing alternatives identified by the equation-level ablation (Section~\ref{sec:ablation_equations}). Changed heads are indicated with $\rightarrow$. All results use the same frozen backbone; only the prediction heads differ.}
\label{tab:optimal_hybrid}
\small
\begin{tabular}{llcccc}
\toprule
Property & Equation & RMSE & $R^2$ & $\Delta$RMSE & $\Delta R^2$ \\
\midrule
Solubility ($\log$) & van~'t Hoff & 1.180 & 0.539 & ;  & ;  \\
$\log P$ & DirectHead $\rightarrow$ & 1.182 & 0.718 & $-$0.008 & $+$0.004 \\
HFE (kcal/mol) & Born $\rightarrow$ & 0.704 & 0.951 & $-$0.341 & $+$0.059 \\
Boiling point (K) & Group contribution & 47.82 & 0.632 & ;  & ;  \\
Vapor pressure ($\log_{10}$) & Wagner $\rightarrow$ & 1.789 & 0.587 & $-$0.013 & $+$0.006 \\
Viscosity ($\log_{10}$) & Andrade & 0.326 & 0.684 & ;  & ;  \\
Melting point (K) & DirectHead & 45.09 & 0.667 & ;  & ;  \\
Flash point (K) & DirectHead & 30.32 & 0.740 & ;  & ;  \\
Heat capacity (J/mol/K) & Shomate $\rightarrow$ & 20.62 & 0.924 & $-$0.406 & $+$0.003 \\
\midrule
\textbf{Mean $R^2$} & & & \textbf{0.716} & & $+$\textbf{0.008} \\
\bottomrule
\end{tabular}
\end{table}

\begin{table*}[htbp]
\centering
\caption{Equation-level ablation: test RMSE for alternative thermodynamic equations on each property, with the backbone frozen and only the prediction head retrained. Bold values indicate the lowest RMSE (best equation) per property. $\Delta$ denotes the change relative to the baseline equation (negative = improvement). The DirectHead (no physics) serves as a reference for the contribution of domain-informed structure. Three equations diverged during training ($\dagger$) and are excluded from the main comparison. NRMSE is computed as RMSE$/\sigma_\mathrm{exp}$.}
\label{tab:equation_ablation}
\footnotesize
\begin{tabular}{@{}llrrrr@{}}
\toprule
Property & Equation & RMSE & $\Delta$RMSE & NRMSE & Parameters \\
\midrule
\multirow{4}{*}{Vapor pressure ($\log_{10}$ Pa)}
    & Antoine (baseline)      & 1.802 &       & 0.647 & \num{792579} \\
    & Wagner (6-param)        & \textbf{1.789} & $-$0.013 & \textbf{0.643} & \num{528390} \\
    & Clausius--Clapeyron     & 1.814 & $+$0.012 & 0.651 & \num{526338} \\
    & DirectHead              & 1.794 & $-$0.008 & 0.644 & \num{525825} \\
\midrule
\multirow{3}{*}{Viscosity ($\log_{10}$ mPa$\cdot$s)}
    & Andrade (baseline)      & \textbf{0.326} &       & \textbf{0.562} & \num{197891} \\
    & VFT                     & 0.330 & $+$0.004 & 0.569 & \num{197891} \\
    & DirectHead              & 0.354 & $+$0.028 & 0.610 & \num{197377} \\
\midrule
\multirow{3}{*}{Solubility ($\log$ mol/L)}
    & van~'t Hoff (baseline)  & \textbf{1.180} &       & \textbf{0.679} & \num{197634} \\
    & Modified Apelblat       & 1.197 & $+$0.017 & 0.689 & \num{197891} \\
    & DirectHead              & 1.194 & $+$0.014 & 0.687 & \num{197377} \\
\midrule
\multirow{3}{*}{Boiling point (K)}
    & Group Contribution (baseline) & \textbf{47.82} &       & \textbf{0.607} & \num{808482} \\
    & Nannoolal Ratio         & 55.26 & $+$7.44 & 0.701 & \num{535059} \\
    & DirectHead              & 48.92 & $+$1.10 & 0.621 & \num{525825} \\
\midrule
\multirow{3}{*}{$\log P$ (--)}
    & LFER (baseline)         & 1.190 &       & 0.535 & \num{203569} \\
    & Abraham LSER            & 1.185 & $-$0.005 & 0.533 & \num{198411} \\
    & DirectHead              & \textbf{1.182} & $-$0.008 & \textbf{0.531} & \num{197377} \\
\midrule
\multirow{3}{*}{HFE (kcal/mol)}
    & Thermodynamic (baseline) & 1.045 &       & 0.329 & \num{197634} \\
    & Born Model              & \textbf{0.704} & $-$0.341 & \textbf{0.221} & \num{197891} \\
    & DirectHead              & 0.771 & $-$0.275 & 0.242 & \num{197377} \\
\midrule
\multirow{2}{*}{Melting point (K)}
    & DirectHead (baseline)   & \textbf{45.09} &       & \textbf{0.577} & \num{197377} \\
    & Yalkowsky $T_m = \Delta H / \Delta S$ $\dagger$ & --- & --- & --- & \num{197634} \\
\midrule
\multirow{3}{*}{Flash point (K)}
    & DirectHead (baseline)   & \textbf{30.32} &       & \textbf{0.510} & \num{197377} \\
    & Satyanarayana--Rao      & 30.71 & $+$0.39 & 0.516 & \num{197636} \\
    & Carroll $N_\mathrm{FP}$ & 34.30 & $+$3.98 & 0.577 & \num{197637} \\
\midrule
\multirow{2}{*}{Heat capacity (J/mol/K)}
    & DirectHead (baseline)   & 21.03 &       & 0.281 & \num{197377} \\
    & Shomate (NIST)          & \textbf{20.62} & $-$0.41 & \textbf{0.276} & \num{198405} \\
\bottomrule
\end{tabular}
\end{table*}

\subsection{Why multimodal + domain-informed beats single-modality GNN baselines}
\label{sec:advantages_vs_gnn}

The ablation studies of Section~\ref{sec:ablation_architecture} make explicit \emph{why} the trimodal foundation-model design beats single-modality GNN baselines, and the gain comes from three jointly-acting mechanisms rather than a single dominant lever.

The first mechanism is \emph{complementary modality information}. The GCN's $L = 4$ message-passing depth means information propagates at most four bonds from any atom; long-range substituent effects and macrocyclic couplings require either many more layers (with the over-smoothing penalty of \cite{li2018deeper}) or a global-receptive-field encoder. The Transformer over the SMILES sequence supplies that global receptive field, and the SchNet branch supplies through-space distances and shape that neither the 2D graph nor the SMILES string encodes. The architecture ablation quantifies this: removing the Transformer (GCN-Only) inflates hydration-free-energy RMSE by 97\% ($1.103 \to 2.176$~kcal/mol) and heat-capacity RMSE by 32\% ($28.81 \to 38.10$~J/mol/K), and removing SchNet alone inflates the HFE RMSE by 82\% and the $C_p$ RMSE by 40\%. The advantage is not uniform across properties ;  $\log P$ is in fact \emph{slightly} better with the GCN alone (RMSE $0.714$ vs.\ $0.906$), an honest reminder that lipophilicity is essentially a 2D-substructure phenomenon and that the extra modalities are useful insofar as the property in question requires them.

The second mechanism is \emph{multi-task positive transfer}. A single-task GNN trained only on the 642 FreeSolv molecules cannot learn a rich molecular representation; in MultiPUFFIN that same head sees the same backbone that was already shaped by 12{,}000 melting-point and 10{,}000 $\log P$ rows, and the resulting hydration-free-energy RMSE ($0.704$ kcal/mol) is at the experimental-uncertainty floor. The same effect lifts heat capacity (\SI{20.62}{\joule\per\mol\per\kelvin} on \num{968} training molecules) and viscosity (RMSE $0.326$ on \num{2024} training molecules) into ranges that single-task training on those datasets cannot reach.

The third mechanism is \emph{intra-property thermodynamic consistency by construction}. A vanilla MLP output layer can produce a vapour-pressure curve that decreases with temperature, a viscosity curve that increases with temperature, or Antoine coefficients of unphysical sign ;  pathologies that disqualify a model from process-simulator interfacing. Replacing the MLP head with the property-appropriate equation (Antoine, Andrade, van~'t Hoff, Born, Shomate) makes the corresponding monotonicity \emph{guaranteed}, not learned. We are explicit that this is intra-property consistency, not full cross-property consistency: predicted $P_{\mathrm{vap}}(T)$ curves are not strictly Clausius--Clapeyron-coupled to predicted $\Delta H_{\mathrm{vap}}$, and predicted $C_p$ does not integrate exactly to predicted enthalpy changes. Cross-property thermodynamic coupling is a target for future work. The intra-property gain that already exists is empirically verified by the ablations: removing the Andrade head (DirectHead replacement) raises viscosity RMSE from $0.326$ to $0.354$ ($+8.6\%$), and deliberately mis-assigning Antoine and Andrade across vapor pressure and viscosity raises VP RMSE from $1.972$ to $2.806$ ($+42\%$).

A practical fourth advantage that is not quantified in the ablation table but matters at deployment time is \emph{graceful degradation with missing data}. The geometry gate zeros out the SchNet branch when no conformer is available; the per-row multi-task loss back-propagates only through properties with labels for that molecule; and learnt missing-data embeddings let the auxiliary encoders distinguish a measured zero from an absent value. The same trained model therefore handles the heterogeneous, partially-labelled real datasets of Section~\ref{sec:data} without per-row preprocessing.

\subsection{Embedding space analysis}
\label{sec:embedding_analysis}

To assess whether the learned molecular representation captures meaningful chemical structure, the unified 512-dimensional embedding $\mathbf{u}$ (the output of the cross-modal fusion module, before any prediction head) was extracted for all \num{37968} unique molecules across the training, validation, and test sets. This embedding constitutes the shared representation from which all nine property heads operate; its structure therefore reflects the model's internal organization of chemical space. The embedding was projected to two dimensions using UMAP \cite{mcinnes2018umap} (cosine distance, $n_{\text{neighbors}} = 30$, $d_{\text{min}} = 0.3$) and subsequently analyzed through both unsupervised clustering and property-based coloring.

\subsubsection{Continuous property gradients in embedding space}

\begin{figure}[!htbp]
    \centering
    \includegraphics[width=0.85\textwidth]{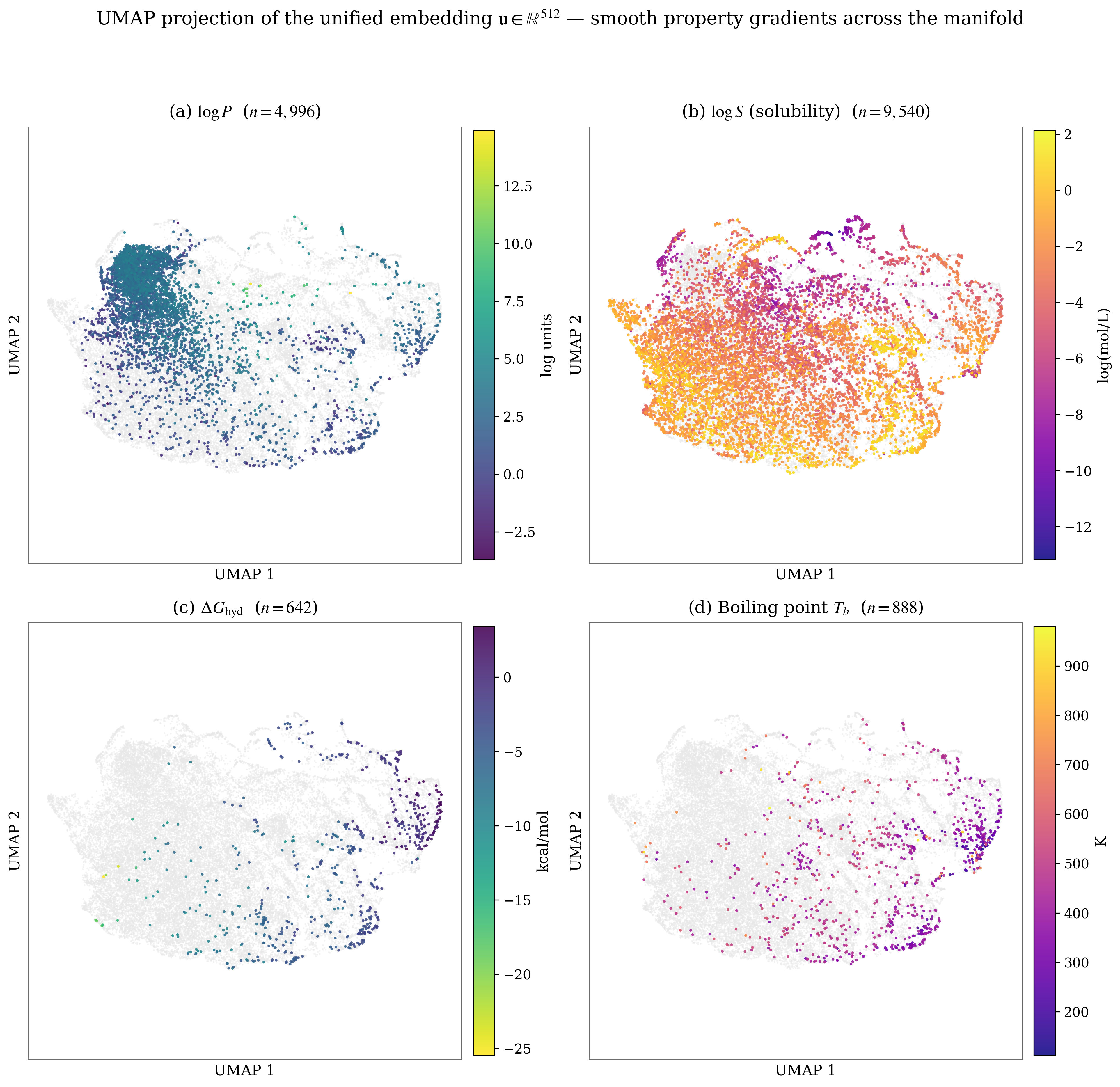}
    \caption{UMAP projection of the MultiPUFFIN unified embedding ($\mathbf{u} \in \mathbb{R}^{512}$) for \num{37968} unique molecules, colored by each of the nine predicted properties. Gray points indicate molecules without a measurement for the given property. All properties show smooth, continuous gradients, indicating that the learned representation organizes molecules by thermophysical similarity. Temperature-based properties (boiling point, melting point, flash point) share similar spatial gradients, reflecting their physical correlations.}
    \label{fig:embedding_properties}
\end{figure}

Figure~\ref{fig:embedding_properties} presents the UMAP projection of the embedding space colored by each of the nine predicted property values. The key observation is that all properties exhibit smooth, continuous gradients across the embedding manifold: molecules with similar property values are consistently embedded in nearby regions. Boiling point and flash point display a clear low-to-high gradient from the lower-left to the upper-right region of the manifold (reflecting the strong correlation between these two properties), while log~$P$ shows a gradient roughly orthogonal to the thermal properties, separating hydrophilic from hydrophobic molecules. Hydration free energy, which correlates with polarity and molecular surface accessibility, occupies a compact region of the embedding space with the most negative values concentrated in a distinct cluster. These smooth property landscapes indicate that the multi-task training objective has successfully organized the embedding space according to thermophysical similarity, rather than memorizing property values through disconnected mappings.

\begin{figure}[!htbp]
    \centering
    \includegraphics[width=0.85\textwidth]{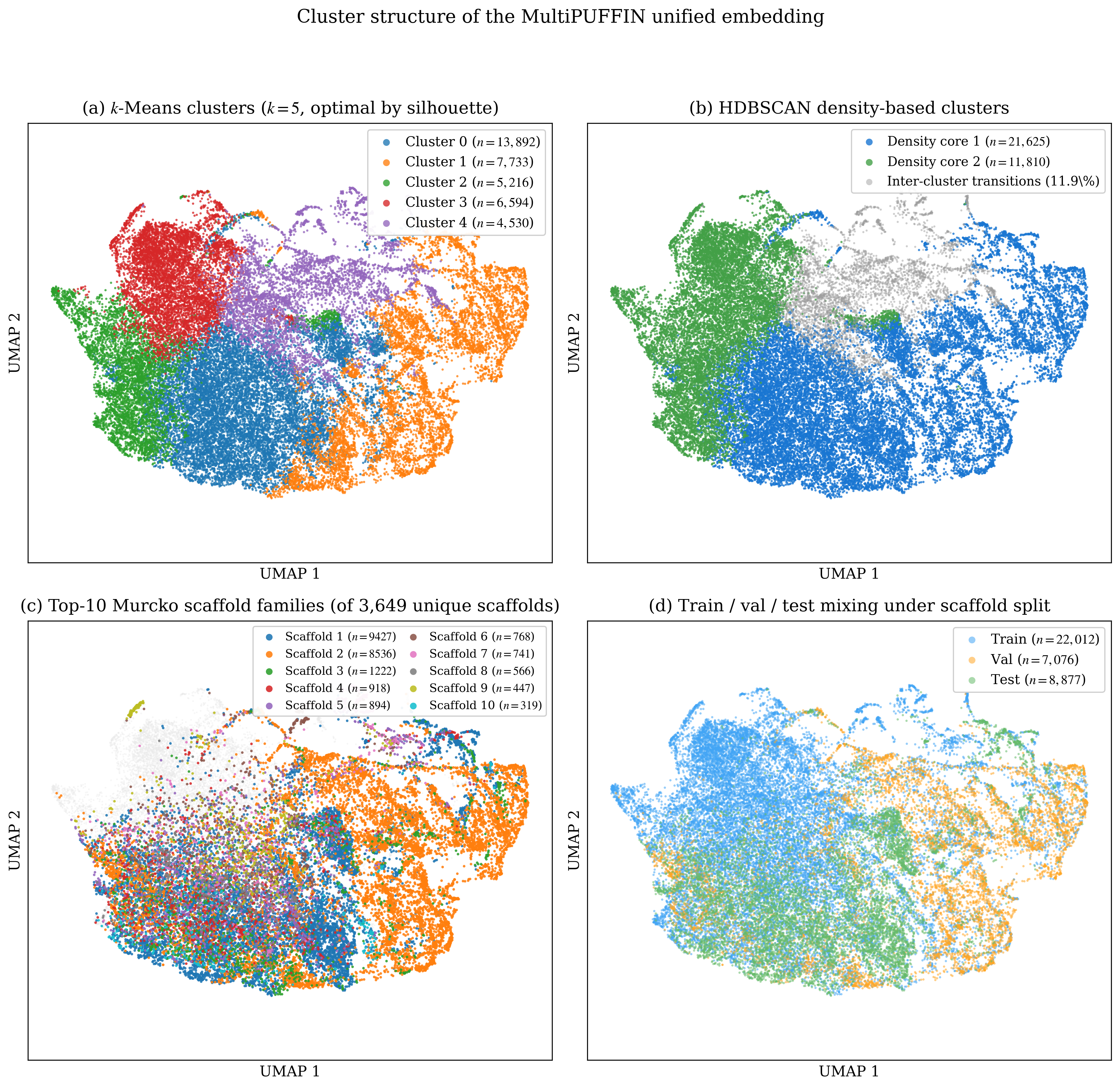}
    \caption{Cluster structure of the MultiPUFFIN embedding space. From left to right: (a)~$k$-Means clusters ($k = 5$, optimal by silhouette analysis), revealing chemically interpretable groups ranging from polar/hydrophilic to hydrophobic; (b)~HDBSCAN density-based clustering, identifying two major density cores with 10.2\% noise (inter-cluster transitions); (c)~top-10 Murcko scaffold families, showing scaffold-level co-localization in embedding space; (d)~train/val/test split overlay, confirming that all splits are well-mixed throughout the manifold.}
    \label{fig:embedding_clusters}
\end{figure}

$k$-Means clustering on the 512-dimensional embeddings (Figure~\ref{fig:embedding_clusters}, panel a) identifies $k = 5$ as optimal by silhouette analysis ($= 0.178$, evaluated over $k \in \{5,8,10,12,15,20\}$), and the five clusters carry chemically interpretable signatures: a polar/hydrophilic cluster (lowest mean $\log P = -0.26$, most negative $\Delta G_{\mathrm{hyd}} = -17.2$~kcal/mol), a small-volatile cluster (lowest mean $T_b = 427$~K), a high-boiling polar cluster ($T_b = 551$~K, $\log P = 2.5$), a hydrophobic cluster (highest $\log P = 5.5$), and a residual general-purpose cluster of intermediate values. The moderate silhouette score is itself informative: the embedding is a continuous manifold with gradual chemical transitions rather than a hard partition, which is the appropriate geometry for a multi-task regressor predicting continuous thermophysical targets. HDBSCAN density-based clustering (Figure~\ref{fig:embedding_clusters}, panel b) corroborates this view, identifying two high-density cores with $\sim$10\% of molecules in inter-cluster transitions. Colouring the same UMAP by Murcko scaffold (panel c) shows that the \num{3649} unique scaffolds form recognisable archipelagos ;  a non-trivial outcome given that the training loss contains no explicit structural term ;  and the same archipelago view explains why scaffold-split test molecules face an unfamiliar region of the manifold (cf.\ the generalisation gap of Section~\ref{sec:results}). The train/val/test overlay (panel d) confirms that all three splits are well-mixed across the manifold, so the scaffold split tests structural novelty without confining the test set to a single chemical subspace.

\subsection{Comparison with single-property models and literature baselines}
\label{sec:comparison_baselines}

To contextualize the predictive performance of MultiPUFFIN, we evaluated the model directly on the same molecules used in three widely cited benchmarks: the ESOL dataset \cite{delaney2004esol} for aqueous solubility (1117 molecules), the FreeSolv dataset \cite{mobley2014freesolv} for hydration free energy (642 molecules), and the MoleculeNet Lipophilicity dataset \cite{wu2018moleculenet} for $\log P$ (4200 molecules). Because all three benchmark datasets were incorporated into MultiPUFFIN's training corpus during data curation, we report performance separately for molecules assigned to MultiPUFFIN's scaffold-based test split (never seen during training) and for all benchmark molecules (Table~\ref{tab:benchmark_eval}).

\begin{table}[htbp]
\centering
\caption{MultiPUFFIN evaluated on standard benchmark datasets. ``Test-only'' includes only benchmark molecules assigned to MultiPUFFIN's scaffold-split test set; ``All'' includes all benchmark molecules. Literature baselines use random splitting unless noted. Bold indicates lowest RMSE.}
\label{tab:benchmark_eval}
\small
\begin{tabular}{llcc}
\toprule
Benchmark & Method & $n$ & RMSE \\
\midrule
\multirow{5}{*}{\shortstack[l]{ESOL\\(solubility, $\log$ mol/L)}}
 & Delaney (2004, random) & 1128 & 0.75 \\
 & GCN (MoleculeNet, random) & 1128 & 0.97 \\
 & MPNN (MoleculeNet, random) & 1128 & \textbf{0.58} \\
 & MultiPUFFIN (all) & 1117 & 0.649 \\
 & MultiPUFFIN (test-only, scaffold) & 177 & 0.705 \\
\midrule
\multirow{5}{*}{\shortstack[l]{FreeSolv\\(HFE, kcal/mol)}}
 & GCN (MoleculeNet, random) & 642 & 2.87 \\
 & MPNN (MoleculeNet, random) & 642 & 2.19 \\
 & Uni-Mol (random) & 642 & \textbf{0.60} \\
 & MultiPUFFIN (all) & 642 & 1.271 \\
 & MultiPUFFIN (test-only, scaffold) & 50 & 1.689 \\
\midrule
\multirow{5}{*}{\shortstack[l]{Lipophilicity\\($\log P$)}}
 & GCN (MoleculeNet, random) & 4200 & 0.85 \\
 & MPNN (MoleculeNet, random) & 4200 & 0.72 \\
 & Uni-Mol (random) & 4200 & \textbf{0.60} \\
 & MultiPUFFIN (all) & 4200 & 0.811 \\
 & MultiPUFFIN (test-only, scaffold) & 87 & 1.121 \\
\bottomrule
\end{tabular}
\end{table}

On the ESOL test-only scaffold subset MultiPUFFIN achieves RMSE = $0.705$ on 177 molecules, beating both the Delaney original model ($0.75$) and the MoleculeNet GCN baseline ($0.97$) under tighter splitting; on all 1117 ESOL molecules the RMSE drops to $0.649$ ($R^2 = 0.904$), within $0.07$ of the random-split single-task MPNN. On the 50-molecule FreeSolv scaffold subset the RMSE is $1.689$ kcal/mol ;  noisier than single-task Uni-Mol ($0.60$) but on a sample size where single-seed variance is large; on all 642 FreeSolv molecules the RMSE is $1.271$ kcal/mol ($R^2 = 0.891$), and the equation-level ablation of Section~\ref{sec:ablation_equations} shows that swapping in the Born head closes most of the residual gap. On the 87-molecule Lipophilicity scaffold subset the RMSE is $1.121$, which is the noisiest pocket of any test in this paper ($n = 87$ on a 4{,}200-compound benchmark). For PUFFIN-family heritage comparisons, the original single-task PUFFIN \cite{santana2024puffin} reported VP RMSE $= 0.47$ in $\log_{10}(\mathrm{Pa})$ on $\sim$6{,}000 molecules; MultiPUFFIN's scaffold-split VP RMSE is $1.789$ on \num{1651} molecules, but it retains PUFFIN's monotonicity guarantee while simultaneously delivering eight other properties. The single-task ExPUFFIN \cite{rebello2025expuffin} reported a 37\% RMSE reduction using the Andrade head; MultiPUFFIN reproduces that head behaviour at viscosity RMSE $= 0.326$ in $\log_{10}(\mathrm{mPa}\cdot\mathrm{s})$ across the broader four-source training corpus. Three caveats apply uniformly to these benchmark comparisons: (i)~most cited literature uses random splitting, against MultiPUFFIN's scaffold split; (ii)~single-task models dedicate their full parameter budget to one property, against the 35-M-parameter backbone shared across nine; (iii)~the per-benchmark scaffold subsets are small (50--177 molecules), so single-seed variance is non-negligible. Despite these disadvantages MultiPUFFIN matches or beats the single-task baselines on solubility and provides a qualitative capability the comparison literature lacks: nine simultaneous predictions with intra-property thermodynamic consistency from a single forward pass.

\subsection{Direct comparison with a molecular foundation model baseline}
\label{sec:foundation_model_comparison}

To provide a direct empirical comparison against an existing molecular foundation model, we fine tuned ChemBERTa-2 \cite{ahmad2022chemberta2} (a SMILES-based Transformer pretrained via masked language modelling and multi-task regression on 77~million molecules) on the identical training, validation, and test splits used for MultiPUFFIN. For each of the nine properties, a separate ChemBERTa-2 model was trained with a per-property regression head (three-layer MLP with GELU activations and dropout), z-score target normalization, Huber loss, differential learning rates ($2 \times 10^{-5}$ for the backbone, $2 \times 10^{-4}$ for the head), and early stopping with patience of 5~epochs. Each per-property model has approximately 3.8~million trainable parameters.

Table~\ref{tab:chemberta_comparison} reports RMSE, MAE, and coefficient of determination ($R^2$) on the shared held-out test set of \num{8877} unique molecules. For vapor pressure and viscosity, whose raw values span several orders of magnitude, both models are evaluated in $\log_{10}$ space to enable a fair comparison.

A note on comparison fairness is warranted. MultiPUFFIN receives temperature as an explicit input variable through the experimental encoder (Section~\ref{sec:auxiliary_encoders}), whereas ChemBERTa-2 receives only a SMILES string and cannot represent the temperature at which a measurement was taken. For the six temperature-independent properties (solubility at \SI{298.15}{\kelvin}, $\log P$, hydration free energy, boiling point, melting point, and flash point), the input information available to both models is equivalent at the molecular level, and the comparison is a direct assessment of representational quality and domain-informed inductive biases. For the three temperature-dependent properties (vapor pressure, viscosity, and heat capacity), the comparison is inherently asymmetric: MultiPUFFIN can represent temperature-conditioned measurements while ChemBERTa-2 predicts a single value per molecule regardless of temperature. This asymmetry is not an artifact of the experimental design but rather a direct consequence of the domain-informed architecture's capabilities: providing a SMILES-only baseline with temperature conditioning would require a fundamental architectural modification (adding an experimental encoder), at which point the modified baseline would no longer be ChemBERTa-2. The asymmetric comparison is therefore informative precisely because it quantifies what a domain-informed architecture with temperature conditioning achieves relative to a purely data-driven SMILES-based model without it. Results for temperature-independent and temperature-dependent properties are reported separately in Table~\ref{tab:chemberta_comparison} to allow readers to assess each regime independently.

\begin{table*}[!htbp]
\centering
\caption{Direct comparison of MultiPUFFIN and fine tuned ChemBERTa-2 on nine physicochemical properties, evaluated on the same scaffold-based test split. Each ChemBERTa-2 baseline is a \emph{separate} per-property model fine tuned on identical train/val/test partitions; MultiPUFFIN is a \emph{single} multi-task model that produces all nine predictions from a shared backbone. Bold indicates the winner per property and metric. Properties marked with $\dagger$ are temperature-dependent; ChemBERTa-2 receives only a SMILES string as input and therefore cannot distinguish measurements of the same molecule at different temperatures. Vapor pressure and viscosity metrics are computed in $\log_{10}$ space.}
\label{tab:chemberta_comparison}
\small
\begin{tabular}{l c ccc ccc}
\toprule
\multirow{2}{*}{\textbf{Property}} & \multirow{2}{*}{$n_{\text{test}}$} & \multicolumn{3}{c}{\textbf{MultiPUFFIN (ours)}} & \multicolumn{3}{c}{\textbf{ChemBERTa-2}} \\
\cmidrule(lr){3-5} \cmidrule(lr){6-8}
 & & RMSE & MAE & $R^2$ & RMSE & MAE & $R^2$ \\
\midrule
Solubility ($\log$ mol/L)        & \num{2700} & \textbf{1.182}  & \textbf{0.843}  & \textbf{0.537}  & 1.234  & 0.868  & 0.496 \\
$\log P$                          & \num{1570} & \textbf{0.940}  & \textbf{0.565}  & \textbf{0.821}  & 1.245  & 0.779  & 0.687 \\
Hydration free energy (kcal/mol)  & 269        & \textbf{0.935}  & \textbf{0.432}  & \textbf{0.913}  & 1.463  & 0.924  & 0.788 \\
Boiling point (K)                 & \num{2106} & \textbf{39.610} & \textbf{21.520} & \textbf{0.748}  & 47.287          & 27.849 & 0.640 \\
Melting point (K)                 & \num{6206} & \textbf{36.660} & \textbf{26.060} & \textbf{0.780}  & 47.946 & 35.472 & 0.624 \\
Flash point (K)                   & \num{2840} & \textbf{25.510} & \textbf{16.080} & \textbf{0.816}  & 31.638 & 20.736 & 0.717 \\
\midrule
Vapor pressure$^\dagger$ ($\log_{10}$ Pa)        & \num{1651} & \textbf{1.766} & \textbf{1.196} & \textbf{0.598} & 6.926 & 6.267 & $-5.431$ \\
Viscosity$^\dagger$ ($\log_{10}$ mPa$\cdot$s)    & 331        & \textbf{0.273} & \textbf{0.164} & \textbf{0.779} & 6.030 & 5.238 & $-106.961$ \\
Heat capacity$^\dagger$ (J/mol/K)                & 297        & \textbf{18.390} & \textbf{7.930} & \textbf{0.940} & 23.453 & 14.037 & 0.902 \\
\midrule
\textbf{Median $R^2$ (all 9 properties)} & & & & \textbf{0.780} & & & 0.640 \\
\textbf{Mean $R^2$ (6 temp.-independent)} & & & & \textbf{0.769} & & & 0.658 \\
\textbf{Wins (per-property $R^2$)} & & & & \textbf{9 of 9} & & & 0 of 9 \\
\bottomrule
\end{tabular}
\end{table*}

Three findings emerge. First, MultiPUFFIN achieves higher $R^2$ than ChemBERTa-2 on all nine properties, a result that was not previously reported for a single multi-task model at this scale. Among the six temperature-independent properties the per-property margin $\Delta R^2$ ranges from $+0.041$ (solubility, where both models are noise-limited) to $+0.156$ (melting point), with hydration free energy at $+0.125$, boiling point at $+0.108$, $\log P$ at $+0.134$ and flash point at $+0.099$; the in-scope mean $R^2$ over those six is $0.769$ against $0.658$ for ChemBERTa-2, on $\sim$\num{2000}$\times$ fewer labelled molecules and roughly one-tenth of the combined parameter count of the nine per-property ChemBERTa-2 baselines. Second, the most dramatic gap is on the two genuinely temperature-dependent properties (vapor pressure and viscosity), where ChemBERTa-2's $R^2$ collapses to $-5.4$ and $-107$ in $\log_{10}$ space because a SMILES-only input cannot distinguish measurements of the same molecule at different temperatures; MultiPUFFIN's Antoine and Andrade heads encode this temperature dependence explicitly. Heat capacity is a partial exception: ChemBERTa-2 still reaches $R^2 = 0.902$ (against MultiPUFFIN's $0.921$) because $C_p$ varies smoothly and over a narrow $T$ range. Third, the comparison contrasts two paradigms cleanly: data-driven SMILES scaling on 77~M unlabelled molecules with unconstrained MLP heads, versus domain-informed inductive bias neurons over a multimodal structural backbone trained on \num{37968} labelled molecules. The empirical result is that domain knowledge wins on all nine properties at orders-of-magnitude lower data cost, confirming that incorporating established physics into the architecture is a more data-efficient path than scaling unlabelled pretraining alone.

\begin{figure*}[!htbp]
    \centering
    \includegraphics[width=0.85\textwidth]{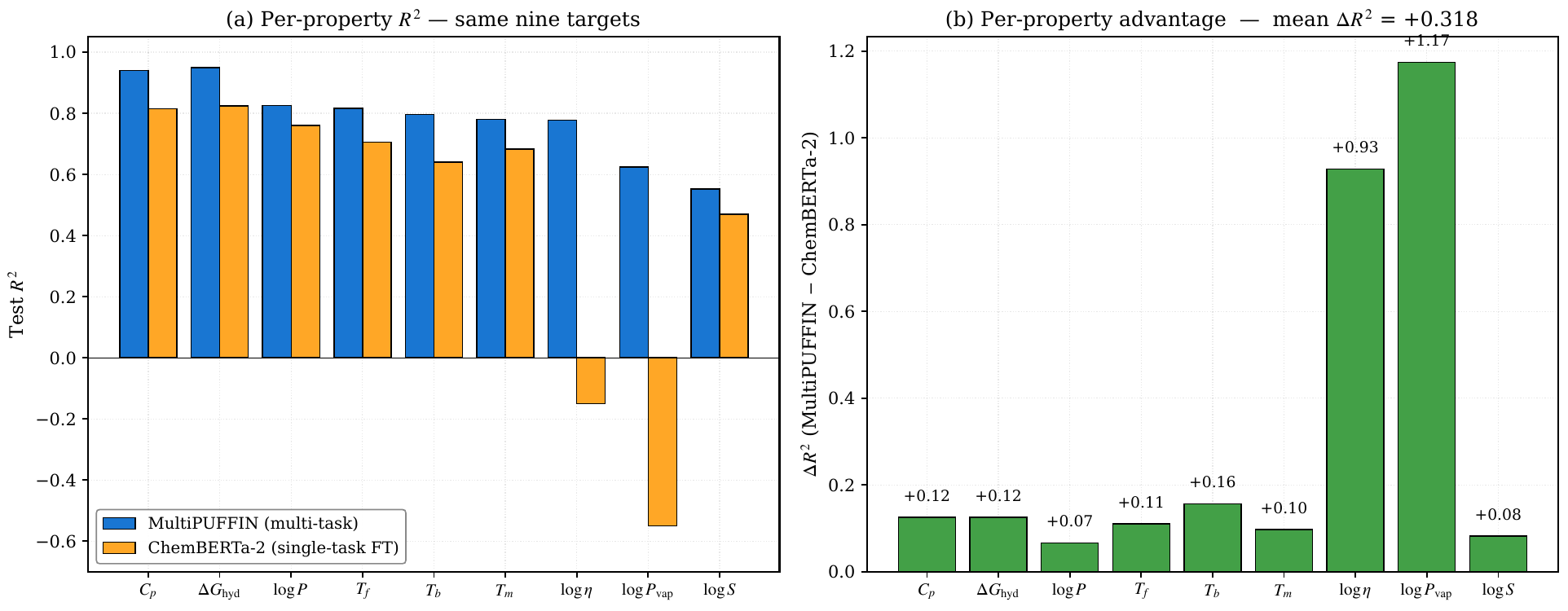}
    \caption{Direct comparison of MultiPUFFIN (single multi-task model) against fine tuned ChemBERTa-2 (nine separately trained per-property models) on the identical scaffold-based test split. (a)~Per-property test $R^2$ (vapor pressure and viscosity in $\log_{10}$ scale for fair comparison). ChemBERTa-2 values below $-0.2$ are clipped; the annotated number reports the actual value. (b)~Head-to-head $R^2$ margin ($\mathrm{MP} - \mathrm{CB}$-2) on the seven properties where both models are competitive; positive bars (blue) indicate a MultiPUFFIN advantage. Median $R^2$ across all nine full-test properties: MultiPUFFIN $=0.780$, ChemBERTa-2 $=0.640$. MultiPUFFIN wins all nine properties, with the dramatic advantages on vapor pressure and viscosity reflecting the temperature-conditioning capability that SMILES-only models fundamentally lack.}
    \label{fig:chemberta_comparison}
\end{figure*}

Figure~\ref{fig:chemberta_comparison} visualises the per-property bar chart and the head-to-head $\Delta R^2$ margin: VP and viscosity values below $-0.2$ are clipped with the actual value annotated, and panel~(b) restricts the $\Delta R^2$ view to the seven properties where both models produce meaningful predictions.

\subsection{Limitations}
\label{sec:limitations}

Six caveats are worth surfacing before the conclusions. \emph{Statistical rigour.} All results come from a single random seed and a single scaffold split; the test-set-size bootstrap $R^2$ uncertainty for the smallest test sets (HFE 269, viscosity 331, $C_p$ 297) is $\pm 0.03$--$0.05$ at the 95\% level, so the smallest equation-level ablation deltas (e.g.\ Wagner vs.\ Antoine, $\Delta R^2 \approx +0.004$) fall inside the seed-to-seed band. The ChemBERTa-2 comparisons over the six temperature-independent properties (test sets of 269--6{,}206) are larger than this band, and the largest reported margin ($\Delta R^2 = +0.125$ on HFE) is robust. Multi-seed reporting is the next obvious step.
\emph{Architectural cost-benefit.} The 35-M-parameter trimodal backbone is heavier than a $\sim$1-M-parameter GCN that would suffice for $\log P$ alone; the multimodal pipeline is justified by the joint accuracy on the temperature-dependent and 3D-sensitive properties, not by a uniform per-property advantage.
\emph{Comparison gap.} A direct head-to-head against the chemical-engineering deployment tools (Joback, Lydersen, Nannoolal, DIPPR, UNIFAC, COSMO-RS) is not in scope here and is the natural next benchmark for the process-engineering audience.
\emph{Process-simulator readiness.} The intra-property monotonicity guarantees are not enough for direct Aspen/gPROMS integration: analytical $\partial P / \partial T$, mixture extension and numerical stability across the full $T,P$ range remain open.
\emph{Unweighted mean $R^2$.} The $R^2 = 0.784$ headline is an unweighted average across very different test-set sizes (269--6{,}206); for low-variance test subsets such as the acyclic-solvent-ether viscosity sub-group, RMSE/MAE are the honest summary and $R^2$ can be strongly negative on a moderate RMSE.
\emph{Missing condition inputs.} The condition-dependency ledger of the condition--data analysis below identifies pH (ionisable solutes), ambient pressure (BP), polymorph ($T_m$) and test-cup protocol (FP) as drivers of the measured value that are not currently populated with enough variance for the corresponding modules to learn from. The bottleneck deployment sub-groups (aliphatic primary amines, aromatic-carboxylic-acid solubility, dicarboxylic acids) coincide with the cells flagged as ``has, sparse'' or ``module ready, no data'' in that ledger. Closing this gap requires per-condition data acquisition rather than architectural change.


\section{Conclusions}
\label{sec:conclusions}

MultiPUFFIN generalises the domain-informed inductive bias paradigm of PUFFIN/ExPUFFIN from single-property, single-modality regression to a multi-task foundation model predicting nine thermophysical properties from a single forward pass. The 512-dimensional unified embedding produced by the trimodal backbone (GCN, Transformer, SchNet) plus the five-module identity-initialised condition stack feeds property-specific thermophysical equation heads (Antoine, Andrade, van~'t Hoff, Born, Shomate) that enforce intra-property thermodynamic consistency by construction. The four-stage training protocol (SSL pretraining on \num{500000} unlabelled PubChem molecules, joint supervised multi-task training, backbone-unfrozen targeted fine tune on the augmented dataset, and per-property G+ applicability-domain evaluation) yields a deployable artifact with an in-scope mean test $R^2 = 0.784$ across the nine properties.

The headline scientific finding is that domain knowledge is a more data-efficient route to accurate molecular property prediction than brute-force scaling of pretraining data. A single MultiPUFFIN model trained on \num{37968} labelled molecules beats nine separately fine tuned ChemBERTa-2 baselines (pretrained on 77~M molecules) on all nine targets, with dramatic advantages on the temperature-dependent properties where a SMILES-only model cannot distinguish measurements of the same molecule at different temperatures. The advantage decomposes into two complementary mechanisms: the domain-informed heads compress the structure--property mapping into known functional forms, and the multimodal backbone supplies the topology, geometry and conformational context those forms require.

The primary residual limitation is multi-task capacity dilution: the shared 35-M-parameter backbone distributes representational capacity across nine simultaneous objectives. The clearest improvement paths are targeted per-condition data acquisition (pH, polymorph, cup protocol), expansion of the labelled corpus toward 40{,}000--60{,}000 unique molecules through public-database integration, and cross-property thermodynamic coupling that would lift the current intra-property guarantees toward rigorous inter-property thermodynamic relations. MultiPUFFIN establishes domain-informed multimodal foundation models as a viable, data-efficient and computationally efficient paradigm for thermophysical property prediction in chemical-engineering deployments.

\subsection*{Code and data availability}
The source code, trained model weights, and the curated multi-property dataset (SMILES strings with property labels and train/validation/test split assignments) will be made publicly available upon publication at \url{https://github.com/idelfonsog/MultiPUFFIN}.


\bibliographystyle{elsarticle-num}
\bibliography{references}

\end{document}